%% file: manuscript.tex
\RequirePackage{fix-cm}
\documentclass[twocolumn]{svjour3}           
\usepackage{adjustbox}
\usepackage{amsfonts}
\usepackage{amsmath}
\usepackage{booktabs}
\usepackage{colortbl}
\usepackage{forest}
\usepackage[breaklinks=true,colorlinks,citecolor=blue,linkcolor=blue]{hyperref}
\usepackage[misc]{ifsym}
\usepackage{gensymb}
\usepackage{graphicx}
\usepackage{fontawesome}
\usepackage{makecell}
\usepackage{multirow}
\usepackage{natbib}
\usepackage{tabularx}
\usepackage{tikz} 
\usepackage{pgf, pgfsys, pgffor} 
\usepackage{pgfplotstable} 
\usepackage{pifont}
\usepackage{url}

\pgfplotsset{compat=1.15} 
\usetikzlibrary{trees, positioning, shapes, shadows, arrows.meta} 

\def\makeheadbox{{%
\hbox to0pt{\vbox{\baselineskip=10dd\hrule\hbox
to\hsize{\vrule\kern3pt\vbox{\kern3pt
\hbox{\bfseries International Journal of Computer Vision Manuscript No.}
\hbox{VISI-D-26-00303}
\kern3pt}\hfil\kern3pt\vrule}\hrule}%
\hss}}}

\newcommand{\dwar}{$\downarrow$}
\newcommand{\upar}{$\uparrow$}
\newcommand{\eg}{\textit{e.g.}}
\newcommand{\ie}{\textit{i.e.}}
\newcommand{\ts}[2]{#1\textsuperscript{#2}}
\newcommand{\tf}[2]{#1\textsubscript{#2}}
\newcommand{\crbx}[2]{\scalebox{0.75}{\fcolorbox{#1}{#1}{\makebox[1.5ex][c]{\rule{0pt}{1.5ex}#2}}}}
\newcommand{\lnk}[1]{\href{#1}{\faExternalLink}}

\newcommand{\rev}[1]{\textcolor{black}{#1}}

\author{
Haozhe Xie\textsuperscript{1, *}\and
Beichen Wen\textsuperscript{1, *}\and
Zhaoxi Chen\textsuperscript{1}\and
Fangzhou Hong\textsuperscript{1}\and
Ziwei Liu~\textsuperscript{1, \Letter}}

\authorrunning{Haozhe Xie, Beichen Wen \textit{et al.}}

\institute{Haozhe Xie \at
S-Lab, Nanyang Technological University\\
\email{haozhe.xie@ntu.edu.sg}
\and
Beichen Wen \at
S-Lab, Nanyang Technological University\\
\email{beichen002@ntu.edu.sg}
\and
Zhaoxi Chen \at
S-Lab, Nanyang Technological University\\
\email{zhaoxi001@e.ntu.edu.sg}
\and
Fangzhou Hong \at
S-Lab, Nanyang Technological University\\
\email{fangzhou001@ntu.edu.sg}
\and
Ziwei Liu \at
S-Lab, Nanyang Technological University\\
\email{ziwei.liu@ntu.edu.sg}
\and
\textsuperscript{1} S-Lab, Nanyang Technological University \\
$^{*}$ Equal contribution\\
\textsuperscript{\Letter} Corresponding author
}
\date{Received: January 30, 2026 / Accepted: August 11, 2026}

\title{3D Scene Generation: A Survey}

\begin{document}

\maketitle

\newcolumntype{Y}{>{\centering\arraybackslash}X}

\input{sections/abstract}
\input{sections/introduction}

\input{sections/preliminaries}

\input{sections/methods}
\input{sections/datasets}
\input{sections/analysis}
\input{sections/applications}
\input{sections/future-work}
\input{sections/acknowledgement}

\bibliographystyle{spbasic}
\bibliography{references}

\end{document}

%% file: sections/abstract.tex
\begin{abstract}
3D scene generation seeks to synthesize spatially structured, semantically meaningful, and photorealistic environments for applications such as immersive media, robotics, autonomous driving, and embodied AI.
Early methods based on procedural rules offered scalability but limited diversity. 
Recent advances in deep generative models (\eg, GANs, diffusion models) and 3D representations (\eg, NeRF, 3D Gaussians) have enabled the learning of real-world scene distributions, improving fidelity, diversity, and view consistency. 
More recently, image- and video-diffusion models have enabled scene generation to leverage large-scale 2D visual priors, substantially improving photorealism.
This survey provides a systematic overview of state-of-the-art approaches, organizing them into four paradigms: procedural generation, neural 3D-based generation, image-based generation, and video-based generation. 
We analyze their technical foundations, trade-offs, and representative results, and review commonly used datasets, evaluation protocols, and downstream applications.
We conclude by discussing key challenges in generation capacity, 3D representation, data and annotations, and evaluation, and outline promising directions including higher fidelity, physics-aware and interactive generation, and unified perception-generation models.
This review organizes recent advances in 3D scene generation and highlights promising directions at the intersection of generative AI, 3D vision, and embodied intelligence.
To track ongoing developments, we maintain an up-to-date project page: \url{https://github.com/hzxie/Awesome-3D-Scene-Generation}.
\keywords{3D Scene Generation \and
Generative Models \and
AI Generated Content \and
3D Vision}
\end{abstract}

%% file: sections/introduction.tex
\section{Introduction}
\label{sec:intro}

The goal of generating 3D scenes is to create a spatially structured, semantically meaningful, and visually realistic 3D environment. 
As a cornerstone of computer vision, it supports a wide range of applications, from immersive filmmaking~\citep{mendiburu20123d,DBLP:journals/air/Anantrasirichai22} and expansive game worlds~\citep{DBLP:conf/siggraph/ParishM01, yannakakis2018artificial, short2017procedural} to architectural visualization~\citep{DBLP:journals/tog/MullerWHUG06,DBLP:conf/iccv/ChangCLMNT21}. 
It also plays a crucial role in AR/VR~\citep{lavalle2023virtual,DBLP:journals/fthci/LeeBZWXLKBH24,DBLP:journals/air/SolimanADH24}, robotics simulation~\citep{DBLP:journals/corr/abs-2403-09227,DBLP:conf/rss/abs-2410-24164,DBLP:journals/corr/abs-2601-22153}, and autonomous driving~\citep{DBLP:conf/corl/DosovitskiyRCLK17,DBLP:journals/pami/ChenWCJGL24} by providing high-fidelity environments for training and testing.
Beyond these applications, 3D scene generation is vital for advancing embodied AI~\citep{DBLP:conf/nips/DeitkeVHWESHKKM22,DBLP:journals/corr/abs-2407-10943,DBLP:conf/iclr/abs-2407-08725} and world models~\citep{DBLP:journals/corr/abs-2405-03520,DBLP:journals/corr/abs-2501-03575,DBLP:journals/corr/abs-2501-10928}, which depend on diverse, high-quality scenes for learning and evaluation. 
Realistic scene synthesis enhances AI agents' ability to navigate, interact, and adapt, driving progress in autonomous systems and virtual simulations.

\begin{figure}[!t]
  \resizebox{\linewidth}{!}{
    \input{figures/annual-stat}
  }
  \caption{\textbf{Annual statistics of 3D scene generation papers in computer vision conferences, journals, and preprints.} The notable rise in publications and the evolving trends in recent years highlight the need for a comprehensive survey.}
  \label{fig:annual-stat}
\end{figure}

\begin{figure}
  \centering
  \resizebox{\linewidth}{!}{
    \input{figures/organization}
  }
  \caption{\textbf{The overall structure of our comprehensive survey.} Our survey presents three core contributions: 1) a summary of key \textit{representations} and \textit{generative models} in 3D scene generation, 2) a \textit{hierarchical taxonomy} systematically organizing intertwined papers with in-depth analysis, and 3) an exploration of \textit{datasets}, \textit{evaluation metrics}, \rev{\textit{comparative analysis}}, \textit{applications}, along with an outlook on \textit{challenges and future directions}.}
  \label{fig:organization}
\end{figure}

As shown in Figure~\ref{fig:annual-stat}, 3D scene generation has gained significant attention in recent years.
Early scene generation methods relied on procedural generation using rule-based algorithms~\citep{DBLP:journals/tvcg/JiangYZW20} and manually designed assets~\citep{DBLP:journals/corr/abs-2412-15200}, offering scalability and control in game design~\citep{DBLP:journals/tomccap/HendrikxMVI13}, urban planning~\citep{DBLP:journals/corr/abs-2407-17572,DBLP:journals/tog/YangWVW13}, and architecture~\citep{DBLP:journals/tog/TaltonLLDMK11,DBLP:journals/cgf/WuFLW18}.
However, their reliance on predefined rules and deterministic algorithms limits diversity, requiring extensive human intervention for realistic or varied scenes~\citep{DBLP:journals/tog/YuYTTCO11}.
Advances in deep generative models, such as GANs~\citep{DBLP:journals/corr/GoodfellowPMXWOCB14} and diffusion models~\citep{DBLP:conf/nips/HoJA20}, enable neural networks to synthesize diverse, realistic spatial structures by learning real-world distributions. 
Combined with innovations in 3D representations like NeRF~\citep{DBLP:conf/eccv/MildenhallSTBRN20} and 3D Gaussians~\citep{DBLP:journals/tog/KerblKLD23}, neural 3D-based generation methods enhance geometry fidelity, rendering efficiency, and view consistency, making them ideal for photorealistic scene synthesis and immersive virtual environments.
Starting from a single image, image-based scene generation methods leverage camera pose transformations and image outpainting to iteratively synthesize perpetual views~\citep{DBLP:conf/iccv/LiuM0SJK21,DBLP:conf/eccv/LiWSK22} or panoramic local environments~\citep{DBLP:journals/tog/ChenWL22,DBLP:conf/nips/TangZCWF23}. 
Benefiting from the rapid advancement of video diffusion models~\citep{DBLP:journals/corr/abs-2311-15127,DBLP:journals/corr/abs-2402-17177}, video generation quality has significantly improved, leading to a surge in 3D scene generation research over the past two years.
These methods formulate 3D scene generation as a form of video generation and enhance view consistency through temporal modeling~\citep{DBLP:conf/iclr/0001CXHLY024}.
The integration of dynamic 3D representations~\citep{DBLP:conf/cvpr/WuYFX0000W24,DBLP:conf/cvpr/YangGZJ0024} further facilitates the synthesis of immersive and dynamic environments~\citep{DBLP:journals/corr/abs-2411-04928,DBLP:conf/iclr/abs-2406-13527}.

Compared to generating 3D objects and avatars, generating 3D scenes presents significantly greater challenges across several dimensions.
\textbf{1) Scale:}
Objects and avatars typically exist within a fixed, limited spatial extent, while scenes must accommodate multiple entities across a much larger and more variable spatial scale.
\textbf{2) Structural complexity:}
Scenes involve complex spatial and semantic relationships among diverse objects, requiring the model to ensure both functional coherence and overall plausibility.
\textbf{3) Data availability:}
While large-scale datasets for object- and avatar-level generation are abundant, high-quality, annotated 3D scene datasets remain scarce and expensive to collect.
\textbf{4) Fine-grained control:}
Scene generation often demands user control over attributes like object placement, zoning, and style, which remain difficult to incorporate in a flexible and interpretable way.

Despite rapid progress in 3D scene generation, the field lacks a comprehensive survey that systematically categorizes existing approaches, highlights key challenges, and identifies future directions. 
Prior surveys focus on narrow domains such as procedural generation~\citep{DBLP:journals/cgf/SmelikTBB14,DBLP:journals/cgf/CogoKPBORM24}, indoor scenes~\citep{DBLP:journals/jcst/ZhangZLH19,DBLP:journals/cgf/PatilPLFSZ24}, autonomous driving~\citep{Ayyildiz2024ASO}, and text-driven generation~\citep{DBLP:journals/corr/abs-2502-14799,DBLP:conf/pais/GhorabL22}, offering limited perspectives.
Broader surveys on general 3D or 4D content generation~\citep{DBLP:journals/corr/abs-2210-15663,DBLP:journals/corr/abs-2305-06131,DBLP:journals/corr/abs-2401-17807,DBLP:journals/corr/abs-2402-01166,DBLP:journals/corr/abs-2410-04738,DBLP:journals/corr/abs-2503-14501} often treat scene generation only peripherally, leading to fragmented coverage.
Although many existing works explore aspects of scene generation, their broader focus often leads to a fragmented understanding that overlooks critical components.
Some works focus on specific subdomains, such as diffusion models~\citep{DBLP:journals/corr/abs-2410-04738}, text-driven scene generation~\citep{DBLP:journals/corr/abs-2305-06131}, or 4D generation~\citep{DBLP:journals/corr/abs-2503-14501}, 
while others neglect key representations like 3D Gaussians~\citep{DBLP:journals/corr/abs-2210-15663} and image sequences~\citep{DBLP:journals/corr/abs-2401-17807,DBLP:journals/corr/abs-2402-01166}, as well as important paradigms like procedural and video-based generation~\citep{DBLP:journals/corr/abs-2210-15663,DBLP:journals/corr/abs-2401-17807,DBLP:journals/corr/abs-2402-01166}.
%
%
Surveys on world models~\citep{DBLP:journals/corr/abs-2405-03520,DBLP:journals/csur/DingZSZZFYSLSXL26,DBLP:journals/corr/abs-2501-11260} primarily address video prediction in driving scenarios, offering only a partial view.
These limitations motivate a comprehensive survey that unifies recent advances and captures the rapidly evolving landscape of 3D scene generation.

\noindent \textbf{Contributions.}
This survey offers a structured overview of recent advances in 3D scene generation.
We categorize existing methods into four types: procedural, neural 3D-based, image-based, and video-based generation, highlighting their paradigms and trade-offs. 
We also review key applications in scene editing, human-scene interaction, embodied navigation, robotics, and autonomous driving.
Additionally, we examine commonly used scene representations, datasets, and evaluation protocols, and identify current limitations in generative capacity, controllability, and realism.
Finally, we outline future directions including higher fidelity, physics-aware and interactive generation, and unified perception-generation models.

\noindent \textbf{Scope.}
This survey primarily focuses on approaches for generating 3D scenes in 3D scene representations. 
In contrast to generative methods that model scene distributions and synthesize diverse samples, 3D reconstruction methods recover a specific observed scene from one or more inputs.
For a review of reconstruction approaches, readers may refer to~\citep{DBLP:journals/pami/HanLB21,DBLP:journals/pami/HuangWWRJ24}.
Furthermore, this survey excludes general video generation~\citep{DBLP:journals/corr/abs-2402-17177,DBLP:journals/csur/XingFCDHXWJ25} and general 3D object generation~\citep{DBLP:conf/icml/SingerSPAMKGVP023,DBLP:journals/corr/abs-2403-16993,DBLP:conf/cvpr/ZhengLNLHM24} methods, even though they have demonstrated some capability in 3D scene generation.
\rev{Instead, we focus on methods that explicitly model scene-level environments, including scene-centric world models that simulate the spatial and temporal evolution of 3D scenes for prediction, planning, and data generation in downstream tasks such as robotics and autonomous driving.}
This survey complements existing reviews on 3D generative models~\citep{DBLP:journals/corr/abs-2210-15663,DBLP:journals/corr/abs-2305-06131,DBLP:journals/corr/abs-2401-17807,DBLP:journals/corr/abs-2402-01166,DBLP:journals/corr/abs-2410-04738}, as none provide a comprehensive overview of 3D scene generation or its relevant insights.

\noindent \textbf{Organization.}
A summary of this survey’s structure is presented in Figure~\ref{fig:organization}.
Section~\ref{sec:preliminaries} provides the foundational concepts, including task definition and formulation, 3D scene representations, and generative models.
Section~\ref{sec:methods} categorizes existing approaches into four types, detailing each category's paradigm, strengths, and weaknesses.
Section~\ref{sec:datasets-and-evaluation} introduces relevant datasets and evaluation metrics.
\rev{Section~\ref{sec:analysis} presents a comparative analysis of existing approaches, covering quantitative results, computational cost, and failure modes.}
Section~\ref{sec:task-applications} reviews various downstream tasks related to 3D scene generation.
Finally, Section~\ref{sec:challenges-and-future} discusses current challenges, future directions, and concludes the survey.

%% file: figures/annual-stat.tex
\begin{tikzpicture}
  \pgfplotstableread[row sep=\\,col sep=&]{
    Year            & PG   & IG   & VG  & N3G \\
    Before          & 37   & 0    & 0   & 3 \\
    2018            & 3    & 0    & 0   & 5 \\
    2019            & 1    & 4    & 0   & 2 \\
    2020            & 1    & 8    & 0   & 4 \\
    2021            & 4    & 6    & 0   & 8 \\
    2022            & 2    & 9    & 0   & 4 \\
    2023            & 5    & 17   & 2   & 16 \\
    2024            & 14   & 26   & 26  & 26 \\
    2025            & 29   & 23   & 59  & 31 \\
  }
  \paperData

  \begin{axis}[
    ybar stacked,
    width             = \linewidth,
    height            = .65\linewidth,
    ymajorgrids       = true,
    symbolic x coords = {
      Before,
      2018,
      2019,
      2020,
      2021,
      2022,
      2023,
      2024,
      2025
    },
    ymin              = 0,
    ymax              = 160,
    x                 = 6 ex,
    bar width         = 4 mm,
    ylabel            = {\# Papers},
    ylabel style      = {
      font            = \fontsize{9}{9}\selectfont
    },
    y label style     = {at = {(-0.08, 0.5)}},
    x tick style      = {draw = none},
    xticklabel style  = {
      align           = center,
      font            = \fontsize{9}{9}\selectfont,
    },
    ytick             = {0, 10, 20, 30, 40, 50, 60, 70, 80, 90, 100, 110, 120, 130, 140, 150, 160},
    yticklabels       = {0,, 20,, 40,, 60,, 80,, 100,, 120,, 140,, 160},
    yticklabel style  = {
      align           = center,
      font            = \fontsize{9}{9}\selectfont,
    },
    legend image code/.code = {
      \draw [#1] (0 mm, -0.8 mm) rectangle (2 mm, 1.2 mm);
    },
    legend cell align = {left},
    legend style      = {
      at              = {(0.31, 0.99)},
      anchor          = north,
      draw            = none,
      font            = \fontsize{8}{8}\selectfont,
      legend columns  = 1,
      fill opacity    = 0.8,
      /tikz/every even column/.append style={column sep = 1 mm}
    }, 
    ]
    \definecolor{colorN3G}{RGB}{251, 188, 5}
    \definecolor{colorPG}{RGB}{234, 67, 53}
    \definecolor{colorIG}{RGB}{52, 168, 83}
    \definecolor{colorVG}{RGB}{66, 133, 244}
    \addplot [ybar, fill=colorPG]  table[x = Year,y = PG]{\paperData};
    \addplot [ybar, fill=colorN3G] table[x = Year,y = N3G]{\paperData};
    \addplot [ybar, fill=colorIG]  table[x = Year,y = IG]{\paperData};
    \addplot [ybar, fill=colorVG]  table[x = Year,y = VG]{\paperData};
    \node at (axis cs:Before, 40) [above, font=\small] {40};
    \node at (axis cs:2018, 8) [above, font=\small] {8};
    \node at (axis cs:2019, 7) [above, font=\small] {7};
    \node at (axis cs:2020, 13) [above, font=\small] {13};
    \node at (axis cs:2021, 18) [above, font=\small] {18};
    \node at (axis cs:2022, 15) [above, font=\small] {15};
    \node at (axis cs:2023, 40) [above, font=\small] {40};
    \node at (axis cs:2024, 89) [above, font=\small] {89};
    \node at (axis cs:2025, 142) [above, font=\small] {142};

    \addlegendentry{Procedural Generation};
    \addlegendentry{Neural 3D-based Generation};
    \addlegendentry{Image-based Generation};
    \addlegendentry{Video-based Generation};
  \end{axis}
\end{tikzpicture}

%% file: figures/organization.tex
\sf
\definecolor{colorRoot}{RGB}{77, 134, 219}
\definecolor{colorSec1}{RGB}{211, 189, 108}
\definecolor{colorSec2}{RGB}{173, 182, 107}
\definecolor{colorSec3}{RGB}{145, 180, 123}
\definecolor{colorSec4}{RGB}{118, 177, 138}
\definecolor{colorSec5}{RGB}{84, 163, 155}
\definecolor{colorSec6}{RGB}{81, 146, 164}
\definecolor{colorSec7}{RGB}{70, 125, 140}

\begin{forest}
    for tree={
        grow=east,
        reversed=true,
        anchor=base west,
        parent anchor=east,
        child anchor=west,
        base=left,
        font=\normalsize,
        rectangle,
        draw=gray!50,
        rounded corners,
        align=left,
        minimum width=4em,
        edge+={black, line width=0.5pt},
        s sep=3pt,
        l sep = 8mm,
        inner xsep=2pt,
        inner ysep=3pt,
        edge path={
          \noexpand\path [draw, \forestoption{edge}]
          (!u.parent anchor) -- ++(3.5mm,0) |- (.child anchor) \forestoption{edge label};
        },
        ver/.style={rotate=90, child anchor=north, parent anchor=south, anchor=center},
    },
    where level=1{text width=9.5em,font=\small,}{},
    where level=2{text width=11.5em,font=\small,}{},
    where level=3{text width=14.5em,font=\small,}{},
    [
        3D Scene Generation, ver, color=colorRoot!80, fill=colorRoot!50, text=black, font=\normalsize, text centered
        [Introduction (\S~\ref{sec:intro}), color=colorSec1!80, fill=colorSec1!50, text=black]
        [
            Preliminaries (\S~\ref{sec:preliminaries}), color=colorSec2!80, fill=colorSec2!50, text=black
            [
                Task Definition and Formu-\\lation (\S~\ref{sec:task-definition}), color=colorSec2!80, fill=colorSec2!30, text=black
            ]
            [
                3D Scene Representations\\
                (\S~\ref{sec:scene-rep}), color=colorSec2!80, fill=colorSec2!30, text=black
                [Voxel Grid, color=colorSec2!80, fill=colorSec2!15, text=black]
                [Point Cloud, color=colorSec2!80, fill=colorSec2!15, text=black]
                [Mesh, color=colorSec2!80, fill=colorSec2!15, text=black]
                [Neural Fields, color=colorSec2!80, fill=colorSec2!15, text=black]
                [3D Gaussians, color=colorSec2!80, fill=colorSec2!15, text=black]
                [Image Sequence, color=colorSec2!80, fill=colorSec2!15, text=black]
            ]
            [
                Generative Models (\S~\ref{sec:generative-models}), color=colorSec2!80, fill=colorSec2!30, text=black
                [Autoregressive Model, color=colorSec2!80, fill=colorSec2!15, text=black]
                [VAE, color=colorSec2!80, fill=colorSec2!15, text=black]
                [GAN, color=colorSec2!80, fill=colorSec2!15, text=black]
                [Diffusion Model, color=colorSec2!80, fill=colorSec2!15, text=black]
                [Procedural Generator, color=colorSec2!80, fill=colorSec2!15, text=black]
            ]
        ]
        [
            Methods: A Hierarchi-\\calTaxonomy (\S~\ref{sec:methods}), color=colorSec3!80, fill=colorSec3!40, text=black
            [Paradigm-level Trade-offs \\(\S~\ref{sec:paradigm-tradeoffs}), color=colorSec3!80, fill=colorSec3!30, text=black]
            [
                Procedural Gen. (\S~\ref{sec:pcg-methods}), color=colorSec3!80, fill=colorSec3!30, text=black
                [Rule-based Gen. (\S~\ref{sec:pcg-rule-methods}), colorSec3!80, fill=colorSec3!15, text=black]
                [Optm.-based Gen. (\S~\ref{sec:pcg-optm-methods}), colorSec3!80, fill=colorSec3!15, text=black]
                [LLM-based Gen. (\S~\ref{sec:pcg-llm-methods}), colorSec3!80, fill=colorSec3!15, text=black]
            ]
            [
                Neural 3D-based Gen. \\(\S~\ref{sec:neural-3d-methods}), colorSec3!80, fill=colorSec3!30, text=black
                [Scene Parameters (\S~\ref{sec:n3d-param-methods}), color=colorSec3!80, fill=colorSec3!15, text=black]
                [Scene Graph (\S~\ref{sec:n3d-sg-methods}), color=colorSec3!80, fill=colorSec3!15, text=black]
                [Semantic Layout (\S~\ref{sec:n3d-semlay-methods}), color=colorSec3!80, fill=colorSec3!15, text=black]
                [Implicit Layout (\S~\ref{sec:n3d-implay-methods}), color=colorSec3!80, fill=colorSec3!15, text=black]
            ]
            [
                Image-based Gen. (\S~\ref{sec:image-methods}), colorSec3!80, fill=colorSec3!30, text=black
                [Holistic Gen. (\S~\ref{sec:img-holi-methods}), color=colorSec3!80, fill=colorSec3!15, text=black]
                [Iterative Gen. (\S~\ref{sec:img-iter-methods}), color=colorSec3!80, fill=colorSec3!15, text=black]
            ]
            [
                Video-based Gen. (\S~\ref{sec:video-methods}), color=colorSec3!80, fill=colorSec3!30, text=black
                [Two-stage Gen. (\S~\ref{sec:vid-two-methods}), color=colorSec3!80, fill=colorSec3!15, text=black]
                [One-stage Gen. (\S~\ref{sec:vid-one-methods}), color=colorSec3!80, fill=colorSec3!15, text=black]
            ]
        ] 
        [
            Datasets and Evalua-\\tion Metrics (\S~\ref{sec:datasets-and-evaluation}), color=colorSec4!80, fill=colorSec4!50, text=black
            [
                Datasets (\S~\ref{sec:datasets}), color=colorSec4!80, fill=colorSec4!30, text=black
                [Indoor Datasets (\S~\ref{sec:indoor-datasets}), color=colorSec4!80, fill=colorSec4!15, text=black]
                [Natural Datasets (\S~\ref{sec:natural-datasets}), color=colorSec4!80, fill=colorSec4!15, text=black]
                [Urban Datasets (\S~\ref{sec:urban-datasets}), color=colorSec4!80, fill=colorSec4!15, text=black]
            ]
            [
                Evaluation (\S~\ref{sec:evaluation-metrics}), color=colorSec4!80, fill=colorSec4!30, text=black
                [Metrics-based Eval. (\S~\ref{sec:metrics-evaluation}), color=colorSec4!80, fill=colorSec4!15, text=black]
                [Benchmark-based Eval. (\S~\ref{sec:benchmark-evaluation}), color=colorSec4!80, fill=colorSec4!15, text=black]
                [Human Eval. (\S~\ref{sec:human-evaluation}), color=colorSec4!80, fill=colorSec4!15, text=black]
            ]
        ]
        [
            Comparative Results \\and Analysis (\S~\ref{sec:analysis}), color=colorSec5!80, fill=colorSec5!50, text=black
            [Quantitative Comparisons \\(\S~\ref{sec:quantitative-comparison}), color=colorSec5!80, fill=colorSec5!30, text=black]
            [Computational Cost (\S~\ref{sec:computational-cost}), color=colorSec5!80, fill=colorSec5!30, text=black]
            [Failure Analysis (\S~\ref{sec:critical-analysis}), color=colorSec5!80, fill=colorSec5!30, text=black]
        ]
        [
            Tasks and Applications\\(\S~\ref{sec:task-applications}), color=colorSec6!80, fill=colorSec6!50, text=black
            [
                Downstream Tasks (\S~\ref{sec:downstream-tasks}), color=colorSec6!80, fill=colorSec6!30, text=black
                [3D Scene Editing (\S~\ref{sec:scene-editing}), color=colorSec6!80, fill=colorSec6!15, text=black]
                [Human-Scene Interaction (\S~\ref{sec:hsi}), color=colorSec6!80, fill=colorSec6!15, text=black]
                [Embodied Navigation (\S~\ref{sec:navigation}), color=colorSec6!80, fill=colorSec6!15, text=black]
            ]
            [
                Application Domains\\(\S~\ref{sec:applications}), color=colorSec6!80, fill=colorSec6!30, text=black
                [Robotics (\S~\ref{sec:robotics}), color=colorSec6!80, fill=colorSec6!15, text=black]
                [Autonomous Driving (\S~\ref{sec:autonomus-driving}), color=colorSec6!80, fill=colorSec6!15, text=black]
            ]
            [
                Paradigm Choice from an \\Application Perspective \\(\S~\ref{sec:app-driven-paradigms}), color=colorSec6!80, fill=colorSec6!30, text=black
            ]
        ]
        [
            Challenges and \\Future Directions (\S~\ref{sec:challenges-and-future}), color=colorSec7!80, fill=colorSec7!50, text=black
            [Challenges (\S~\ref{sec:challenges}), color=colorSec7!80, fill=colorSec7!30, text=black]
            [Future Directions (\S~\ref{sec:future-directions}), color=colorSec7!80, fill=colorSec7!30, text=black]
        ]
    ]
\end{forest}

%% file: sections/preliminaries.tex
\section{Preliminaries}
\label{sec:preliminaries}

\subsection{Task Definition and Formulation}
\label{sec:task-definition}

3D scene generation maps an input $\mathbf{x}$ (\textit{e.g.,} random noise, text, images, or other conditions) to a \textbf{3D scene representation} $\mathbf{S}$ (Sec.~\ref{sec:scene-rep}) using a \textbf{generative model} $\mathcal{G}$ (Sec.~\ref{sec:generative-models}).
\begin{equation}
  \mathcal{G}: \mathbf{x} \to \mathbf{S}
\end{equation}
The generated scene $\mathbf{S}$ is spatially coherent, defines 3D geometry implicitly or explicitly, and supports multi-view rendering, simulation, or downstream interaction.

\subsection{3D Scene Representations}
\label{sec:scene-rep}

Various 3D scene representations have been developed and utilized in computer vision and graphics.
In this section, we provide an overview of the key 3D scene representations, discussing their structures, properties, and suitability for 3D scene generation.

\noindent \textbf{Voxel Grid.}
A voxel grid is a 3D array $\mathbf{V} \in \mathbb{R}^{H \times W \times D}$, where $H$, $W$, and $D$ represent the height, width, and depth of the grid, respectively. 
Each voxel stores properties such as occupancy or signed distance values~\citep{DBLP:conf/siggraph/CurlessL96}, enabling structured volumetric scene representation.

\noindent \textbf{Point Cloud.}
A point cloud is an unordered set of $N$ 3D points $\mathbf{P} = \{ \mathbf{p}_i \mid \mathbf{p}_i \in \mathbb{R}^3 \}_{i=1}^{N}$ that approximates an object's surface. 
Point clouds are sparse, unstructured, and memory-efficient representations, typically obtained from depth sensors, LiDAR, or structure-from-motion~\citep{DBLP:conf/cvpr/SchonbergerF16}.

\noindent \textbf{Mesh.}
A polygonal mesh $\mathbf{M} = \{ \mathbf{M}_V, \mathbf{M}_E, \mathbf{M}_F \}$ defines a 3D surface through vertices $\mathbf{M}_V$, edges $\mathbf{M}_E$, and faces $\mathbf{M}_F$ (\eg, triangles or quads), explicitly encoding connectivity and making it well-suited for surface modeling in 3D scenes. 

\noindent \textbf{Neural Fields.}
Signed Distance Fields (SDF)~\citep{DBLP:conf/cvpr/ParkFSNL19} and Neural Radiance Fields (NeRF)~\citep{DBLP:conf/eccv/MildenhallSTBRN20} are continuous implicit representations parameterized by neural networks.
SDF maps a 3D position $\mathbf{x} \in \mathbb{R}^3$ to a signed distance $s(\mathbf{x}) \in \mathbb{R}$, with surfaces defined as the zero-level set, while NeRF maps $\mathbf{x}$ and a view direction $\mathbf{r} \in \mathbb{R}^3$ to a volume density $\sigma(\mathbf{x})$ and color $\mathbf{c}(\mathbf{x}, \mathbf{r})$.
Rendering is typically performed via sphere tracing for SDF~\citep{DBLP:journals/vc/Hart96} and differentiable volume rendering for NeRF~\citep{DBLP:conf/siggraph/KajiyaH84,DBLP:journals/tvcg/Max95a}.

\noindent \textbf{3D Gaussians.}
3D Gaussians~\citep{DBLP:journals/tog/KerblKLD23} represent scenes using a set of $N$ Gaussian primitives $\mathbf{G}=\{ (\boldsymbol{\mu}_i, \Sigma_i, \mathbf{c}i, \alpha_i) \}_{i=1}^{N}$, where $\boldsymbol{\mu}_i \in \mathbb{R}^3$ denotes the center, $\Sigma_i \in \mathbb{R}^{3 \times 3}$ the anisotropic covariance, $\mathbf{c}_i \in \mathbb{R}^3$ the color, and $\alpha_i \in [0,1]$ the opacity.
Images are generated by projecting and rasterizing these Gaussians onto the image plane.

{\noindent{\bf{Image Sequence.}}}
An image sequence, $\mathbf{C}=\{ \mathbf{I}_i \in \mathbb{R}^{H \times W \times 3} \}_{i=1}^N$, captures a scene from multiple viewpoints and implicitly encodes its 3D structure. 
It is widely adopted in image- and video-based 3D scene generation, where geometry can be recovered from multi-view reconstruction.

\subsection{Generative Models}
\label{sec:generative-models}

Generative models for 3D scene generation fall into two categories: procedural generators~\citep{DBLP:journals/cgf/SmelikTBB14} and data-driven methods (e.g., autoregressive models, VAEs~\citep{DBLP:journals/corr/KingmaW13}, GANs~\citep{DBLP:journals/corr/GoodfellowPMXWOCB14}, and diffusion models~\citep{DBLP:conf/nips/HoJA20}).
The former synthesizes scenes via predefined rules and parameters, while the latter learns scene distributions from data.

\noindent \rev{\textbf{Autoregressive} (AR) models generate data sequentially by factorizing the joint distribution as $p(\mathbf{x}) = \prod_{t=1}^T p(\mathbf{x}_t \mid \mathbf{x}_{<t})$. 
They capture dependencies among elements through conditional prediction, typically implemented with neural networks~\citep{DBLP:conf/nips/VaswaniSPUJGKP17,DBLP:journals/corr/abs-1803-03324}.}

\noindent \rev{\textbf{Variational Autoencoders} (VAEs)
~\citep{DBLP:journals/corr/KingmaW13} encode data into a probabilistic latent space and reconstruct it through a decoder. 
They support smooth latent interpolation but often produce blurrier samples due to likelihood-based optimization~\citep{DBLP:journals/corr/abs-1810-00597,DBLP:conf/nips/LucasTG019}.}

\noindent \rev{\textbf{Generative Adversarial Networks} (GANs)~\citep{DBLP:journals/corr/GoodfellowPMXWOCB14} train a generator and discriminator adversarially to produce realistic data.
They yield high-quality samples but can be unstable and suffer from mode collapse~\citep{DBLP:conf/nips/GulrajaniAADC17}.}

\noindent \rev{\textbf{Diffusion Models}~\citep{DBLP:conf/nips/HoJA20} generate data by learning to reverse a gradual noising process. 
They provide strong generation quality and stability, but usually require expensive iterative sampling~\citep{DBLP:conf/nips/DhariwalN21}.}

\noindent \rev{\textbf{Procedural Generators}~\citep{DBLP:journals/cgf/SmelikTBB14} synthesize 3D scenes through rule-based or stochastic operations. 
They offer strong controllability and reproducibility, making them particularly suitable for structured scene creation.}

%% file: sections/methods.tex
\section{Methods: A Hierarchical Taxonomy}
\label{sec:methods}

\begin{figure}
  \includegraphics[width=\linewidth]{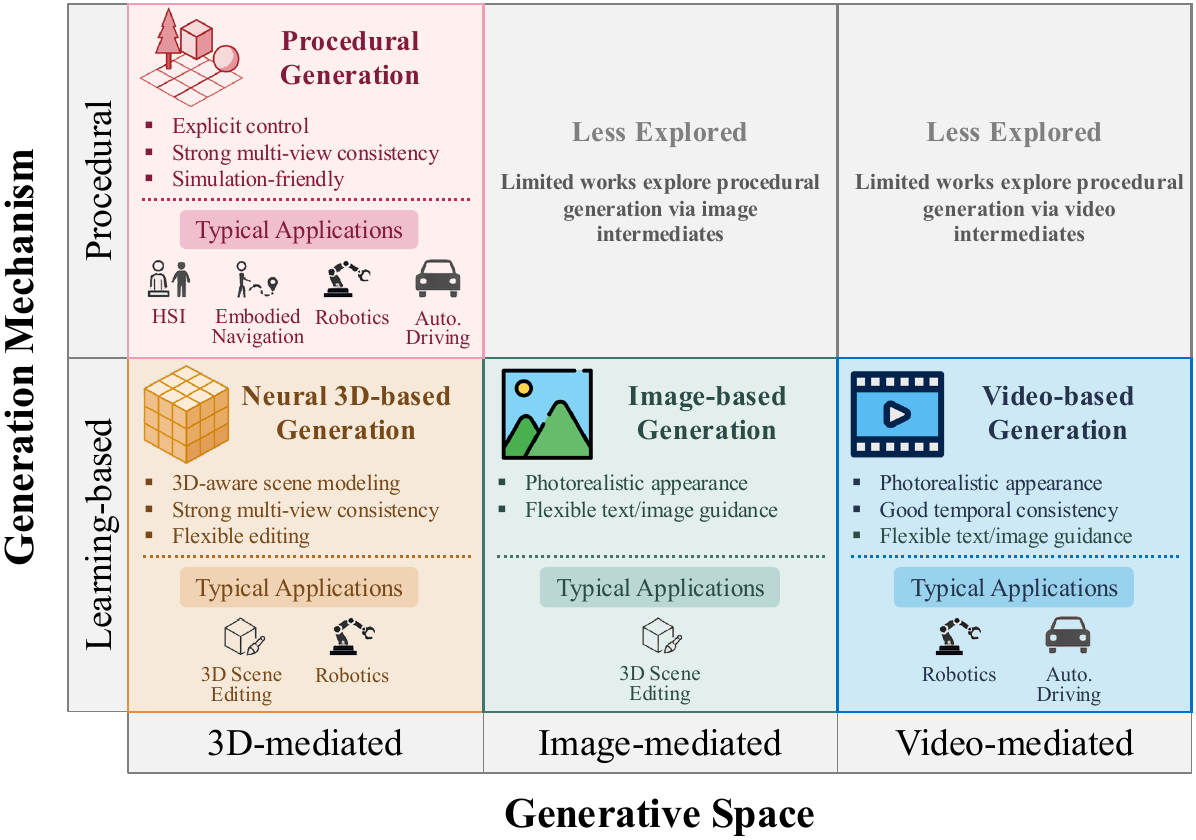}
  \caption{\rev{\textbf{A two-axis view of 3D scene generation methods.} Methods are organized along two orthogonal dimensions: generation mechanism and generative space, leading to four representative paradigms. For each paradigm, we highlight its characteristic capabilities and the typical applications they enable.}}
  \label{fig:taxonomy}
\end{figure}

\rev{Methods for 3D scene generation differ in how scene layouts and geometry are determined during generation.
We view existing approaches along two orthogonal dimensions (Figure~\ref{fig:taxonomy}):
\textbf{(1) generation mechanism}, \ie, rule-based (procedural) versus data-driven (learning-based), and
\textbf{(2) generative space}, \ie, whether scene structure is generated directly in 3D space or mediated through 2D observations (images or videos).
Under this view, learning-based methods can be grouped into \textit{native 3D} and \textit{2D-mediated} generation.
In practice, most methods correspond to specific combinations of these dimensions, forming recurring design patterns.
These should be understood as dominant regimes rather than strictly disjoint classes, as many recent methods combine multiple mechanisms and representations.
For clarity, we organize existing methods into four representative paradigms based on these dominant combinations:}
\begin{itemize}
\item \rev{\textbf{Procedural Generation} constructs 3D scenes using predefined rules, constraints, or simulation processes.
Scene layouts and often explicit meshes are determined by rule systems or programmatic pipelines, rather than learned from data, typically operating directly in 3D space.}

\item \rev{\textbf{Neural 3D-based Generation} uses neural networks to predict scene layouts or geometry directly in 3D representations, such as object parameters, scene graphs, semantic layouts, or 3D fields.
Scene layouts and geometry are generated within the 3D domain during the generation process, representing learning-based methods in native 3D space.}

\item \rev{\textbf{Image-based Generation} formulates 3D scene synthesis as an image generation or outpainting problem.
Scene layouts and geometry are inferred implicitly by enforcing spatial consistency across synthesized views, representing learning-based methods mediated through image space.}

\item \rev{\textbf{Video-based Generation} formulates scene synthesis as a video generation problem.
Scene layouts, geometry, and dynamics are inferred implicitly by enforcing temporal consistency across frames, representing learning-based methods mediated through video space.}
\end{itemize}

\input{tables/representative-methods}

\begin{figure*}
    \centering
    \includegraphics[width=\linewidth]{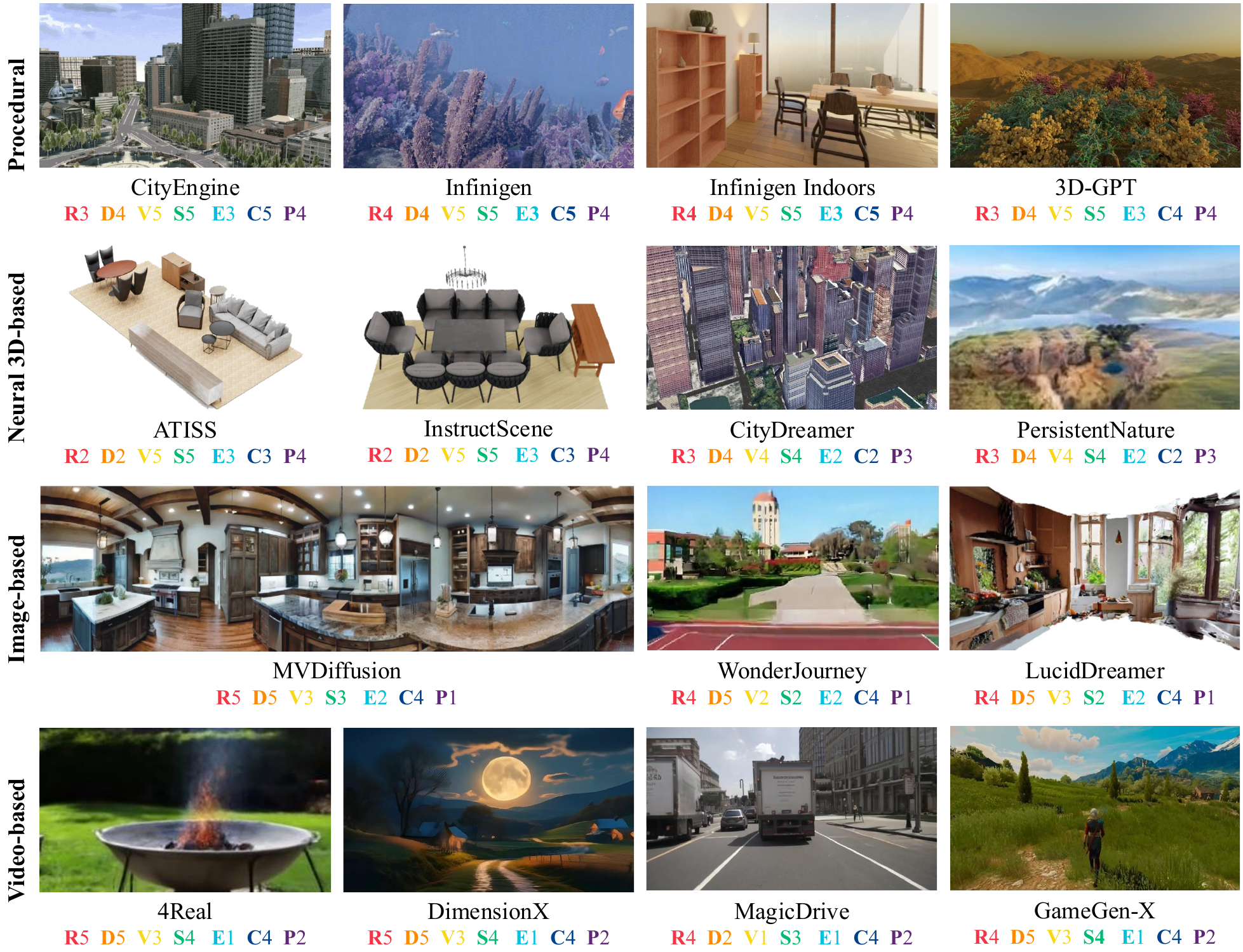}
    \caption{\rev{\textbf{Summary and comparison of representative works for 3D scene generation}. We score each method on a scale from 1 to 5, where a higher score indicates stronger performance on the corresponding dimension. 
    Specifically, 
    \textcolor[rgb]{0.96078431, 0.20784314, 0.27058824}{\bf R}, 
    \textcolor[rgb]{1, 0.54117647, 0}{\bf D}, 
    \textcolor[rgb]{0.98039216, 0.84313725, 0.08235294}{\bf V}, 
    \textcolor[rgb]{0, 0.72941176, 0.44705882}{\bf S}, 
    \textcolor[rgb]{0, 0.76078431, 0.87058824}{\bf E}, 
    \textcolor[rgb]{0, 0.25882353, 0.55294118}{\bf C}, and 
    \textcolor[rgb]{0.37647059, 0.14509804, 0.47058824}{\bf P} denote 
    \textcolor[rgb]{0.96078431, 0.20784314, 0.27058824}{Realism}, 
    \textcolor[rgb]{1, 0.54117647, 0}{Diversity}, 
    \textcolor[rgb]{0.98039216, 0.84313725, 0.08235294}{View Consistency}, 
    \textcolor[rgb]{0, 0.72941176, 0.44705882}{Semantic Consistency}, 
    \textcolor[rgb]{0, 0.76078431, 0.87058824}{Efficiency}, 
    \textcolor[rgb]{0, 0.25882353, 0.55294118}{Controllability}, and 
    \textcolor[rgb]{0.37647059, 0.14509804, 0.47058824}{Physical Plausibility}, respectively.}
    As their importance varies by application, Table~\ref{tab:app-driven-paradigms} lists the must-have dimensions for representative use cases, enabling direct cross-reference with the per-method scores shown here.}
    \label{fig:representative-methods}
\end{figure*}

\subsection{Paradigm-level Trade-offs}
\label{sec:paradigm-tradeoffs}

\noindent \textbf{Realism.}
Procedural generation spans a wide spectrum of visual fidelity, from stylized content to photorealistic scenes used in film and visual effects, largely depending on the complexity of procedural rules, asset quality, and rendering pipelines~\citep{DBLP:conf/cvpr/RaistrickLMMWZK23}.
Neural 3D-based generation can produce geometrically coherent scenes, while its visual realism is often influenced by the scale and fidelity of available 3D training data.
Image-based generation is highly effective at producing photorealistic results by leveraging large-scale real-world imagery~\citep{DBLP:conf/nips/SahariaCSLWDGLA22, DBLP:conf/icml/NicholDRSMMSC22}, although it does not always ensure accurate and globally consistent geometry.
Video-based generation further enhances perceptual realism through temporal coherence, but it can still be affected by accumulated geometric drift~\rev{\citep{DBLP:journals/corr/abs-2604-10578}}.

\noindent \textbf{Diversity.}
Procedural generation achieves diversity through rule variation and combinatorial asset composition, which can be limited by predefined design spaces~\citep{DBLP:books/daglib/0034882}.
Neural 3D-based generation inherits diversity from training distributions and may struggle to generalize beyond them~\citep{DBLP:conf/nips/GuoNFZKYLW24}.
Image- and video-based generation benefit from the long-tail diversity present in real-world visual data, enabling rich appearance variations, though coverage remains implicitly constrained by dataset biases~\citep{DBLP:conf/cvpr/DombrowskiZCRK25}.

\noindent \textbf{View Consistency.}
Procedural and neural 3D-based generation explicitly models scene structure, which naturally enforces consistency across multiple views~\citep{DBLP:books/daglib/0034882, DBLP:conf/cvpr/Lee0L24}. 
In contrast, image- and video-based methods operate in the view domain and only infer geometry implicitly, making global cross-view consistency harder to achieve unless an explicit 3D representation is introduced as an intermediate or post-processing stage~\citep{DBLP:conf/iclr/LiuLZLLKW24}.

\noindent \textbf{Semantic Consistency.}
Procedural and neural 3D-based generation preserve semantic consistency through a globally defined scene representation that maintains elements across viewpoints~\citep{DBLP:conf/iccv/JainTA21}. 
Image-based generation lacks global context and is prone to semantic drift under large viewpoint changes~\citep{DBLP:conf/siggrapha/PatashnikGCZT24}. Video-based methods improve short-term consistency via temporal coherence, but long-range stability across large viewpoint shifts remains challenging~\citep{DBLP:journals/corr/abs-2502-17863}.

\noindent \textbf{Efficiency.}
Procedural generation may involve costly rule evaluation, simulation, or asset preparation, though it can scale efficiently once pipelines are established~\citep{DBLP:journals/tomccap/HendrikxMVI13}.
Neural 3D-based generation can be computationally intensive, especially for volumetric rendering methods requiring dense sampling and repeated network evaluations per ray~\citep{DBLP:conf/eccv/MildenhallSTBRN20}, though recent representations~\citep{DBLP:journals/tog/MullerESK22, DBLP:journals/tog/KerblKLD23} improve efficiency with trade-offs.
Image- and video-based generation benefit from efficient feed-forward inference when generating individual views or short sequences, but may suffer from limited reuse of intermediate structure across viewpoints or time~\citep{DBLP:conf/iclr/YuNHLSA24}.

\noindent \textbf{Controllability.}
Procedural generation offers fine-grained and interpretable control through explicit rules, parameters, and constraints, at the cost of manual design effort~\citep{DBLP:journals/tomccap/HendrikxMVI13}.
Neural 3D-based generation provides indirect control through structured representations or learned priors, but often lacks intuitive high-level interfaces~\citep{DBLP:conf/cvpr/KaniaYKTT22}.
Image- and video-based generation is controlled by text, images, or reference videos, enabling flexible semantic guidance while offering less explicit control over precise geometry or physical constraints~\citep{DBLP:journals/pami/CaoZSY26}.

\noindent \textbf{Physical Plausibility.}
\rev{Physical plausibility is closely tied to the simulatability of the output representation. 
Explicit object-level meshes equipped with material properties, articulation, and collision geometry are directly usable by physics engines. 
By contrast, neural fields, 3D Gaussians, occupancy grids, and image sequences typically do not directly provide object-decomposed collision geometry, articulation, or material and physical parameters required for simulation. 
Procedural and retrieval-based methods can inherit simulatability from structured assets when these assets include the necessary geometry and physical metadata, enabling physics-based validation but limiting diversity through predefined rules or asset libraries~\citep{DBLP:conf/aiide/StephensonR16}.}
Neural 3D-based methods capture geometric regularities but are not necessarily simulation-ready, making physical plausibility encouraged rather than enforced~\citep{DBLP:conf/cvpr/XuPYSZ22}.
Image- and video-based methods rely on learned physical priors, achieving visual realism while struggling with hard constraints and rare interactions ~\citep{DBLP:journals/corr/abs-2502-07001}.

\noindent
Overall, these characteristics reflect common tendencies of each paradigm under typical formulations rather than strict limitations. 
In practice, hybrid designs and implementation choices can significantly blur these boundaries. 
\rev{Moreover, the relative importance of these seven dimensions is application-dependent: a dimension that is a hard constraint for one consumer may be a soft preference for another. 
Section~\ref{sec:app-driven-paradigms} and Table~\ref{tab:app-driven-paradigms} provide this demand-side view, identifying the must-have dimensions for representative application scenarios.}
\rev{To make these trade-offs more concrete, Figure~\ref{fig:representative-methods} compares representative methods across the seven dimensions, while Figure~\ref{fig:failure} illustrates typical failure modes, which are further analyzed in Section~\ref{sec:critical-analysis}.}

\subsection{Procedural Generation}
\label{sec:pcg-methods}

\begin{figure}[!t]
  \includegraphics[width=\linewidth]{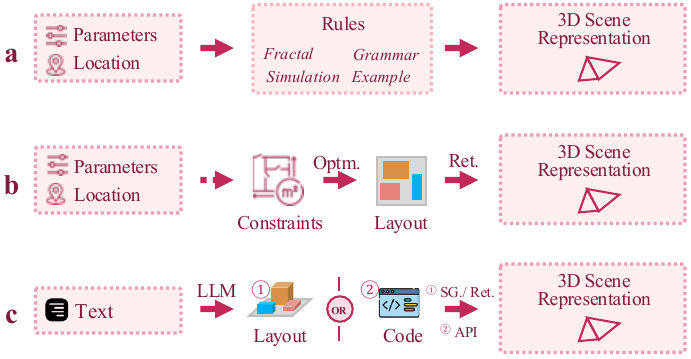}
  \caption{\textbf{The paradigms of procedural methods for 3D scene generation.} \textbf{(a)} Rule-based generation methods follow predefined rules to generate 3D scenes. \textbf{(b)} Optimization-based generation finds an optimized scene under different constraints. \textbf{(c)} LLM-based generation uses large language models (LLMs) for tasks like layout design and object selection, or to generate code that controls other generators. Note that dashed arrows denote optional operations. ``Optm.'', ``Ret.'', and ``SG.'' denote ``Optimization'', ``Retrieval'', and ``Shape Generation'', respectively. 
  }
  \label{fig:paradigm-pcg-based}
\end{figure}

Procedural generation methods create 3D scenes automatically by adhering to predefined rules or constraints.
They are widely employed in computer graphics to generate diverse environments, such as terrains, vegetation, rivers, roads, interior spaces, buildings, and even entire cities.
As summarized in Section~\ref{sec:paradigm-tradeoffs}, procedural methods are efficient and structurally consistent by design, but their expressiveness is constrained by handcrafted rules and parameters.
The paradigms of procedural generation are illustrated in Figure~\ref{fig:paradigm-pcg-based}, and can be broadly categorized into rule-based, optimization-based, and LLM-based approaches.

\subsubsection{Rule-based Generation}
\label{sec:pcg-rule-methods}

Rule-based procedural generation includes various approaches that create 3D scenes using clearly defined rules and algorithms. 
These methods produce scene geometry directly, which is subsequently rendered for visualization. 
Common techniques in this category include fractal-based, grammar-based, simulation-driven, and example-based generation.

Fractals~\citep{doi:10.1126/science.156.3775.636, mandelbrot1983fractal,DBLP:journals/cacm/FournierFC82} are mathematical structures that exhibit self-similarity across scales.
Fractal-based methods are widely applied in terrain modeling and texture synthesis, as they efficiently generate visually complex patterns while requiring minimal storage.
Techniques such as midpoint displacement~\citep{DBLP:conf/graphicsinterface/PrzemyslawM93,DBLP:conf/graphite/BelhadjA05} and fractional Brownian motion~\citep{mandelbrot1968fractional} (fBM) generate multi-scale details that resemble natural landscapes. 

Grammar-based methods consist of an alphabet of symbols, an initial axiom, and a set of rewriting rules.
Each generated symbol encodes geometric commands for complex shape generation.
CityEngine~\citep{DBLP:conf/siggraph/ParishM01} extends L-systems~\citep{lindenmayer1968mathematical} for the generation of road networks and building geometry to create cities. 
\cite{DBLP:journals/tog/MullerWHUG06} build upon shape grammars~\citep{DBLP:conf/ifip/StinyG71} to model highly detailed 3D buildings.

Simulation-based procedural generation creates realistic 3D environments by modeling natural and artificial processes. 
Some methods simulate erosion effects~\citep{DBLP:conf/siggraph/KelleyMN88,DBLP:conf/siggraph/MusgraveKM89,DBLP:journals/tog/CordonnierJPBCBGGG23} and hydrology~\citep{DBLP:journals/tog/GenevauxGGPB13,DBLP:journals/tog/SchottPFGG23,DBLP:journals/tog/ParisGCG23} to generate terrain with high fidelity.
Vegetation simulations model plant growth under resource competition~\citep{DBLP:conf/siggraph/DeussenHLMPP98,DBLP:journals/tog/CordonnierGGBGP17,DBLP:journals/tog/MakowskiHSMPP19} and climate change~\citep{DBLP:journals/tog/PalubickiMGHMP22}.
%
In urban contexts, ecosystem-based approaches populate cities with vegetation~\citep{DBLP:conf/si3d/BenesAJAV11}, while others simulate city growth and resource distribution to generate settlements that evolve organically over time~\citep{DBLP:journals/tog/VanegasABW09,DBLP:journals/cgf/WeberMWG09}.

To improve controllability, example-based procedural methods have been introduced. 
These approaches take a small user-provided example and synthesize larger scenes by either extending its boundaries~\citep{DBLP:conf/si3d/Merrell07,DBLP:journals/tog/MerrellM08} or matching structural and geometric features~\citep{DBLP:journals/tvcg/ZhouSTR07,DBLP:journals/cgf/NishidaGA16}.
Inverse procedural generation further enhances high-level control over the synthesis process. 
Such methods employ optimization objectives to infer procedural parameters~\citep{DBLP:journals/tog/TaltonLLDMK11,DBLP:journals/tog/VanegasGABW12} or learn global distributions that govern scene layout and organization~\citep{DBLP:journals/tog/EmilienVCPB15}.

The aforementioned techniques are frequently combined to leverage their complementary strengths in generating large-scale and diverse scenes. 
For instance, Citygen~\citep{kelly2007citygen} integrates road network synthesis with building generation to produce urban environments, while Infinigen~\citep{DBLP:conf/cvpr/RaistrickLMMWZK23} unifies material, terrain, vegetation, and creature generators to create unbounded natural scenes.

\subsubsection{Optimization-based Generation}
\label{sec:pcg-optm-methods}

Optimization-based generation formulates scene synthesis as an optimization problem that minimizes objectives encoding predefined constraints. 
These constraints, typically derived from physics rules, functionality, or design principles, are embedded into cost functions and optimized using stochastic or sampling-based methods. 
Alternatively, statistical approaches learn spatial relationships from data and guide the layout process through probabilistic sampling. 
Some systems support user-defined constraints and user interactions to enable controllable and semantically meaningful generation. 

Some methods model physical and spatial constraints as cost functions and employ stochastic optimization techniques for scene synthesis. 
Physical-level constraints typically address object interpenetration, stability, and friction~\citep{DBLP:conf/graphicsinterface/XuSF02}. 
At the layout level, constraints such as functional relationships (\eg, co-occurrence and accessibility), interior design principles (\eg, symmetry, alignment, and co-circularity), and human behavior patterns have been extensively explored~\citep{DBLP:journals/tog/MerrellSLAK11,DBLP:journals/tog/YuYTTCO11,DBLP:conf/vr/KanK18}.
In addition, users can specify high-level constraints, such as scene type, scale, and layout, to enable more controllable and semantically meaningful scene generation~\citep{DBLP:journals/cgf/WuFLW18,DBLP:journals/corr/abs-2111-05527,DBLP:conf/nips/DeitkeVHWESHKKM22}. 
Building upon existing procedural generation pipelines, Infinigen Indoors~\citep{DBLP:conf/cvpr/RaistrickMKY0HW24} introduces a constraint specification API that allows users to define custom constraints and achieve highly controllable scene synthesis.

Other methods adopt data-driven models to learn object arrangement patterns from annotated data, transforming scene generation into a probabilistic sampling problem. 
Bayesian networks are commonly used~\citep{DBLP:journals/tog/MerrellSK10,DBLP:journals/tog/FisherRSFH12,DBLP:journals/tvcg/YuYT16} to capture conditional dependencies between objects, while graph-based models~\citep{DBLP:conf/cvpr/QiZHJZ18,DBLP:journals/tvcg/ZhangZXLYF22,DBLP:conf/mm/ZhangLHYZ21} model spatial hierarchies or relational structures to improve spatial reasoning and object placement accuracy.

\subsubsection{LLM-based Generation}
\label{sec:pcg-llm-methods}

Large Language Models~\citep{DBLP:conf/nips/Ouyang0JAWMZASR22} (LLMs) and Vision-language models~\citep{DBLP:journals/corr/abs-2303-08774} (VLMs) have introduced a new paradigm in procedural generation by enabling text-driven scene synthesis, allowing users to specify environments through natural language descriptions, offering greater flexibility and user control over scene design.

Several approaches use LLMs to generate scene layouts, such as object parameters~\citep{DBLP:conf/nips/FengZFJAHBWW23, DBLP:conf/eccv/FuWLS24, DBLP:conf/eccv/OcalTKG24, DBLP:conf/icml/ZhouRXHLWS024} and scene graphs~\citep{DBLP:journals/corr/abs-2404-02838, DBLP:conf/cvpr/abs-2503-18476, DBLP:conf/cvpr/GaoLC0S24, DBLP:conf/siggrapha/LiL0M024}.
Based on these layouts, 3D assets can be obtained through object retrieval or shape generation.
Specifically, 
LayoutGPT~\citep{DBLP:conf/nips/FengZFJAHBWW23} guides LLMs using generation prompts and structural templates to produce object parameters for retrieving assets.
CityCraft~\citep{DBLP:journals/corr/abs-2406-04983} guides land-use planning with LLMs and retrieves building assets from a database to construct detailed urban environments.
I-Design~\citep{DBLP:journals/corr/abs-2404-02838} and~\cite{DBLP:conf/cvpr/abs-2503-18476} use graph-based object representations to model inter-object semantics more effectively.
To support more stylized and versatile scene generation, 
GraphDreamer~\citep{DBLP:conf/cvpr/GaoLC0S24} and Cube~\citep{DBLP:journals/corr/abs-2503-15475} generate scene graphs via LLMs, treating nodes as objects and enabling compositional scene generation through 3D object generation models.
The Scene Language~\citep{DBLP:conf/cvpr/abs-2410-16770} introduces a language-based scene representation composed of a program, words, and embeddings, which can be generated by LLMs and rendered using traditional, neural, or hybrid graphics pipelines.

Other methods utilize LLMs as agents to control procedural generation by adjusting parameters of rule-based systems or modifying operations within procedural generation software.
\cite{DBLP:conf/mm/LiuZZZ24} employ LLMs to fine-tune parameters in rule-based landscape generation, optimizing procedural workflows with learned priors. 
3D-GPT~\citep{DBLP:conf/3dim/abs-2310-12945} and SceneCraft~\citep{DBLP:conf/icml/HuIJKYRSF24} generate Python scripts to control existing procedural frameworks, such as Infinigen~\citep{DBLP:conf/cvpr/RaistrickLMMWZK23} and Blender\footnote{\url{https://www.blender.org/}}, allowing direct manipulation of procedural assets.
Holodeck~\citep{DBLP:conf/cvpr/YangSWVHH0HKLCY24} generates 3D environments through multiple rounds of conversation with an LLM, including floor and wall texturing, door and window generation, object selection and placement.
City$\mathcal{X}$~\citep{DBLP:journals/corr/abs-2407-17572} and Scene$\mathcal{X}$~\citep{DBLP:conf/aaai/abs-2403-15698} use a multi-agent system for different stages of generation, producing Python code for layout, terrain, building, and road generation through Blender rendering. 
WorldCraft~\citep{DBLP:journals/corr/abs-2502-15601} further incorporates object generation and animation modules.

\subsection{Neural 3D-based Generation}
\label{sec:neural-3d-methods}

\begin{figure}[!t]
  \includegraphics[width=\linewidth]{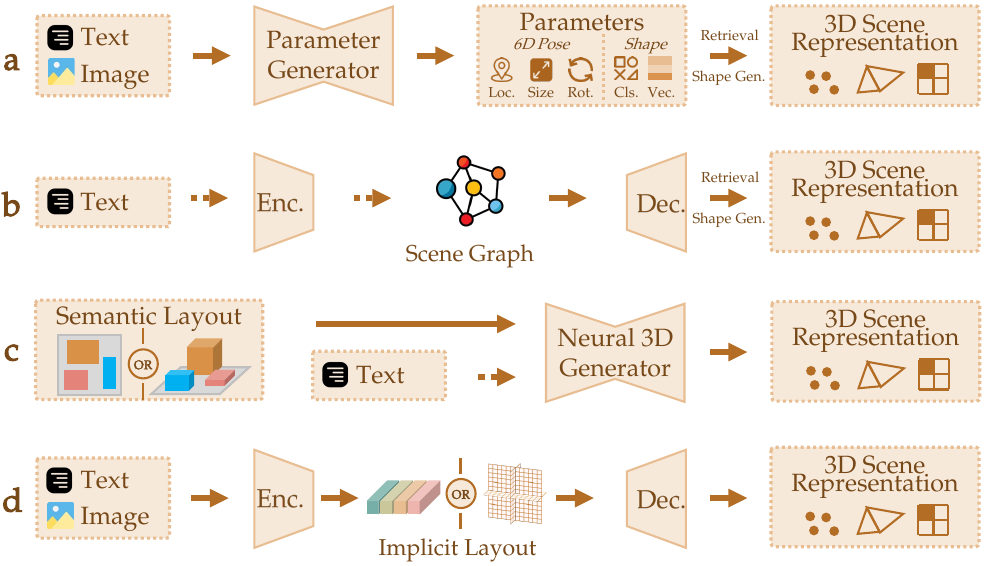}
  \caption{\textbf{The paradigms of neural 3D-based methods for 3D scene generation.} 
  These paradigms use \textbf{(a)} scene parameters, \textbf{(b)} scene graphs, \textbf{(c)} semantic layouts, and \textbf{(d)} implicit layouts as intermediate representations to control the spatial arrangement of generated 3D scenes. These representations, either user-provided or produced by generative models, are then converted into 3D scene representations (\eg, voxel grid, mesh, NeRF, or 3D Gaussians) via retrieval or decoding.
  Note that dashed arrows denote optional operations. ``Enc.'' and ``Dec.'' stand for ``Encoder'' and ``Decoder'', respectively. ``Shape Gen.'' represents ``Shape Generation''.}
  \label{fig:paradigm-n3d-based}
\end{figure}

Neural 3D-based methods generate 3D scene representations using generative models trained on datasets with 3D annotations.
Recent advancements in NeRF and 3D Gaussians have further enhanced the fidelity and realism.
As discussed in Section~\ref{sec:paradigm-tradeoffs}, these methods provide strong view consistency and semantic consistency due to their explicit 3D modeling, while their controllability and scalability are constrained by the underlying representations.
As shown in Figure~\ref{fig:paradigm-n3d-based}, the methods are categorized into four types based on the spatial arrangement that controls the layout of generated 3D scenes: scene parameters, scene graph, semantic layout, and implicit layout.

\subsubsection{Scene Parameters}
\label{sec:n3d-param-methods}

Scene parameters offer a compact way to represent object arrangements, implicitly capturing inter-object relationships without relying on explicit scene graphs. 
These parameters typically encompass an object's location, size, orientation, class, and shape latent code.
As illustrated in Figure~\ref{fig:paradigm-n3d-based}\hyperref[fig:paradigm-n3d-based]{a}, these methods first generate scene parameters as an intermediate representation, which is then used to synthesize the final 3D scene.

DeepSynth~\citep{DBLP:journals/tog/WangSCR18}, FastSynth~\citep{DBLP:conf/cvpr/Ritchie0L19}, \cite{DBLP:journals/tog/ZhangYMLHVH20}, and Sync2Gen~\citep{DBLP:conf/iccv/0005ZYHMZBH21} adopt CNN-based architectures that utilize top-down image-based scene representations, sequentially inserting objects by predicting their parameters.
Subsequent works explore more advanced models, such as transformers and diffusion models.
ATISS~\citep{DBLP:conf/nips/PaschalidouKSKG21}, SceneFormer~\citep{DBLP:conf/3dim/WangYN21}, COFS~\citep{DBLP:conf/siggraph/ParaGMW23}, and \cite{DBLP:conf/cvpr/NieD0N23} use transformers to autoregressively generate object parameters.
RoomDesigner~\citep{DBLP:conf/3dim/ZhaoZLDG24} refines this process by decoupling layout and shape generation, ensuring shape compatibility in indoor scenes.
CASAGPT~\citep{DBLP:conf/cvpr/abs-2504-19478} leverages cuboids as intermediate object representations to better avoid object collisions.
DeBaRA~\citep{DBLP:conf/nips/MaillardSDO24} adopts a diffusion model for object parameter generation, while PhyScene~\citep{DBLP:conf/cvpr/YangJZH24} further integrates physical constraints for physical plausibility and interactivity.

To improve controllability in text-driven scene generation, 
RelScene~\citep{DBLP:conf/mm/YeZLP24} employs BERT~\citep{DBLP:conf/naacl/DevlinCLT19} to align spatial relationships with textual descriptions in latent space. 
DiffuScene~\citep{DBLP:conf/cvpr/TangNMDTN24} leverages latent diffusion models~\citep{DBLP:conf/cvpr/RombachBLEO22} to generate object parameters from text inputs, followed by object retrieval.
Ctrl-Room~\citep{DBLP:conf/3dim/abs-2310-03602} and SceneFactor~\citep{DBLP:conf/cvpr/abs-2412-01801} employ LDMs to generate coarse object layouts from text prompts, with fine-grained appearance obtained via panorama generation and geometric diffusion model, respectively.
\cite{DBLP:conf/icml/EpsteinPMEH24}, SceneWiz3D~\citep{DBLP:journals/corr/abs-2312-08885}, and DreamScene~\citep{DBLP:conf/eccv/LiSZWLWLZ24} adopt a multi-stage approach, first generating an initial object layout, then refining object geometry using Score Distillation Sampling (SDS)~\citep{DBLP:conf/iclr/PooleJBM23}, followed by a global refinement step to improve compositional consistency.

Human movement and interactions often influence the organization of environments, where motion patterns and physical contact inform the arrangement of objects and scene layouts.
Pose2Room~\citep{DBLP:conf/eccv/NieD0N22} introduces an end-to-end generative model that predicts the bounding boxes of furniture in a room from human motion.
SUMMON~\citep{DBLP:conf/siggrapha/YeWLPLX022} and MIME~\citep{DBLP:conf/cvpr/YiHTHTB23} further improve semantic consistency and physical affordances by generating objects with meshes that align with human-scene contact.
\cite{DBLP:conf/nips/AnVNHNVN23} propose a multi-conditional diffusion model that integrates text prompts to enhance controllability. 
To ensure physically plausible layouts free from interpenetration and unintended collisions, INFERACT~\citep{DBLP:conf/siggraph/LiHZW24} optimizes scene layout generation while simultaneously simulating human movement in a physics-based environment using reinforcement learning.

\subsubsection{Scene Graph}
\label{sec:n3d-sg-methods}

Scene graphs offer a structured, symbolic representation of 3D scenes, with nodes representing objects and edges capturing their spatial relationships. 
Incorporating scene graphs allows generative models to enforce spatial constraints and preserve relational consistency, facilitating the creation of well-structured 3D environments.
Following the paradigm illustrated in Figure~\ref{fig:paradigm-n3d-based}\hyperref[fig:paradigm-n3d-based]{b}, scene graphs, whether generated by models or provided as input, function as layout priors that guide the decoding process to create 3D scene representations by object retrieval or shape generation.

Early data-driven approaches~\citep{DBLP:conf/emnlp/ChangSM14,DBLP:journals/cgf/KermaniLTZ16,DBLP:journals/tog/FuCWWZF17,DBLP:journals/tog/MaPFLPHYTGZ18} represent spatial relationships between objects using scene graphs, which serve as a blueprint for 3D scene generation through object retrieval and placement.
Subsequent works enhance graph representations and introduce advanced generative models. 
PlanIT~\citep{DBLP:journals/tog/WangLWSCR19} employs a deep graph generative model to synthesize scene graphs, followed by an image-based network for object instantiation.
GRAINS~\citep{DBLP:journals/tog/LiPXCKSTCCZ19} adopts a recursive VAE to learn scene structures as hierarchical graphs, which can be decoded into object bounding boxes.
3D-SLN~\citep{DBLP:conf/cvpr/LuoZ0T20} utilizes scene graphs as a structural prior for 3D scene layout generation, ensuring spatial coherence, and further incorporates differentiable rendering to synthesize realistic images.
Meta-Sim~\citep{DBLP:conf/iccv/KarPLCYRA0F19} and Meta-Sim2~\citep{DBLP:conf/eccv/DevaranjanKF20} use scene graphs to structure scene generation, optimizing parameters for visual realism and synthesizing diverse 3D scenes using rendering engines.

Previous methods enable scene generation from scene graphs but rely on object retrieval or direct synthesis, limiting geometric diversity.
To address this, Graph-to-3D~\citep{DBLP:conf/iccv/DhamoMNT21} introduces a graph-based VAE that jointly optimizes layout and shape.
SceneHGN~\citep{DBLP:journals/pami/GaoSMLGY23} models scenes as hierarchical graphs (layout → object geometry) with a hierarchical VAE.
CommonScenes~\citep{DBLP:conf/nips/ZhaiOWDTNB23} and EchoScene~\citep{DBLP:conf/eccv/ZhaiOCLDNTB24} use dual-branch graph diffusion to jointly model global scene–object relations and local inter-object interactions.
MMGDreamer~\citep{DBLP:conf/aaai/YangLZQJMYXXXLL25} employs a mixed-modality graph for fine-grained geometric control.

Recent methods improve controllability by integrating human input.
SEK~\citep{DBLP:conf/eccv/WuFWXDMM24} encodes scene knowledge as a scene graph within a conditioned diffusion model for sketch-driven scene generation.
InstructScene~\citep{DBLP:conf/iclr/LinM24} integrates text encoders with graph-based generative models for text-driven scene synthesis.
To generalize scene-graph-based generation to broader scenes, \cite{DBLP:conf/cvpr/abs-2503-07152} map scene graphs onto a Bird's Eye View (BEV) embedding map, which guides a diffusion model for large-scale outdoor scene synthesis.
HiScene~\citep{DBLP:journals/corr/abs-2504-13072} leverages VLM-guided occlusion reasoning and video diffusion-based amodal completion to generate editable 3D scenes with compositional object identities from a single isometric view.

\subsubsection{Semantic Layout}
\label{sec:n3d-semlay-methods}

Semantic layouts serve as an intermediate representation that encodes the structural and semantic organization of a 3D scene. 
They provide high-level guidance for 3D scene generation, ensuring controllability and coherence in the placement of objects and scene elements. 
As shown in Figure~\ref{fig:paradigm-n3d-based}\hyperref[fig:paradigm-n3d-based]{c}, semantic layouts, whether user-provided or generated, act as precise constraints for generative models, guiding 3D scene generation while enabling optional textural prompts for style control.

A 2D semantic layout consists of a 2D semantic map, sometimes including additional maps such as height maps, viewed from a top-down perspective.
CC3D~\citep{DBLP:conf/iccv/BahmaniPPYWGT23} generates a 3D feature volume conditioned on a 2D semantic map, which serves as a NeRF for neural rendering. 
BerfScene~\citep{DBLP:conf/cvpr/ZhangXS0ZY24} incorporates positional encoding and low-pass filtering to make the 3D representation equivariant to the BEV map, enabling controllable and scalable 3D scene generation.
Frankenstein~\citep{DBLP:conf/siggrapha/YanLWCSSLCDMLJ24} encodes scene components into a compact triplane~\citep{DBLP:conf/cvpr/ChanLCNPMGGTKKW22}, generated via a diffusion process conditioned on a 2D semantic layout. 
BlockFusion~\citep{DBLP:journals/tog/WuLYSSWCLSLJ24} introduces a latent triplane extrapolation mechanism for unbounded scene expansion.
Incorporating a height map with the semantic map enables the direct conversion of 2D layouts into 3D voxel worlds, essential for urban and natural scenes where building structures and terrain elevation provide important priors.
InfiniCity~\citep{DBLP:conf/iccv/LinLMCS0T23} utilizes InfinityGAN~\citep{DBLP:conf/iclr/LinLCT022} to generate infinite-scale 2D layouts, which are then used to create a watertight semantic voxel world, with textures synthesized through neural rendering.
For natural scenes, SceneDreamer~\citep{DBLP:journals/pami/ChenWL23} uses a neural hash grid to model diverse landscapes in a space- and scene-varied hyperspace.
For urban scenes, CityDreamer~\citep{DBLP:conf/cvpr/Xie0H024} and GaussianCity~\citep{DBLP:journals/corr/abs-2406-06526} separate background and building generation to better handle structural diversity, while CityDreamer4D~\citep{DBLP:journals/corr/abs-2501-08983} further incorporates dynamic traffic for large-scale 4D city generation.

A 3D semantic layout offers enhanced capability to represent more complex 3D layouts compared to 2D, improving controllability, typically by using voxels or 3D bounding boxes.
GANcraft~\citep{DBLP:conf/iccv/HaoMB021} uses voxels as the 3D semantic layout, optimizing a neural field with pseudo-ground truth and adversarial training.
UrbanGIRAFFE~\citep{DBLP:conf/iccv/YangYGX0L23} and DisCoScene~\citep{DBLP:conf/cvpr/XuCSPSSYSLZT23} break down the scene into stuff, objects, and sky, and adopt compositional neural fields for scene generation.
By incorporating score distillation sampling (SDS)~\citep{DBLP:conf/iclr/PooleJBM23}, 3D semantic layouts offer better control over text-guided scene generation, improving the alignment of generated scenes with textual descriptions.
Comp3D~\citep{DBLP:conf/3dim/PoW24}, CompoNeRF~\citep{DBLP:journals/corr/abs-2303-13843}, Set-the-Scene~\citep{DBLP:conf/iccvw/Cohen-BarRMGC23}, and Layout-your-3D~\citep{DBLP:journals/corr/abs-2410-15391} generate 3D scenes with compositional NeRFs using pre-defined customizable layouts as object proxies.
SceneCraft~\citep{DBLP:conf/nips/YangMCW24} and Layout2Scene~\citep{DBLP:journals/corr/abs-2501-02519} generate indoor scenes by distilling the pretrained diffusion models.
Urban Architect~\citep{DBLP:journals/corr/abs-2404-06780} integrates geometric and semantic constraints with SDS, leveraging the scalable hash grid to ensure better view-consistency in urban scene generation.

\subsubsection{Implicit Layout}
\label{sec:n3d-implay-methods}

Implicit layouts are feature maps that encode the spatial structure of a 3D scene.
As shown in Figure~\ref{fig:paradigm-n3d-based}\hyperref[fig:paradigm-n3d-based]{d}, these layouts manifest as latent features of different dimensions.
Encoders learn to embed 3D scene layout information into latent feature maps, which are then used by the decoder to generate 3D scenes in the form of NeRF, 3D Gaussians, or voxel grids.

Recent advances in representations like NeRFs and 3D Gaussians have enabled neural networks to directly generate and render high-fidelity RGB images from latent feature maps. 
Some methods leverage these representations to produce appearance-consistent 3D scenes with photorealistic quality.
NeRF-VAE~\citep{DBLP:conf/icml/KosiorekSZMSMR21} encodes shared information across multiple scenes using a VAE.
GIRAFFE~\citep{DBLP:conf/cvpr/Niemeyer021} represents scenes as compositional generative neural fields to disentangle objects from background.
GSN~\citep{DBLP:conf/iccv/DeVries0STS21} and Persistent Nature~\citep{DBLP:conf/cvpr/Chai0LIS23} adopt GAN-based architectures to generate 2D latent grids as implicit scene layouts, which are sampled along camera rays to guide NeRF rendering.
GAUDI~\citep{DBLP:conf/nips/0001GATTCDZGUDS22} employs a diffusion model to learn scene features and camera poses jointly, decoding them into a tri-plane and pose for NeRF-based rendering control.
NeuralField-LDM~\citep{DBLP:conf/cvpr/0001BYKSLR0F23} decomposes NeRF scenes into a hierarchical latent structure that includes 3D voxel, 2D BEV, and 1D global representations. 
Hierarchical diffusion models are then trained on this tri-latent space for generation.
Director3D~\citep{DBLP:conf/nips/LiLXQCZ0J24} uses a Gaussian-driven multi-view latent diffusion model to generate pixel-aligned and unbounded 3D Gaussians along a generated trajectory, followed by SDS refinement.
Prometheus~\citep{DBLP:journals/corr/abs-2412-21117} and SplatFlow~\citep{DBLP:conf/cvpr/abs-2411-16443} learn a compressed latent space from multi-view images, and decode this latent space into pixel-aligned 3DGS representations.

Another branch of work focuses more on generating semantic structure and scene geometry, typically using voxel grids as representations. 
These methods are not immediately renderable but can be textured through external rendering pipelines.
\cite{DBLP:journals/corr/abs-2301-00527} introduce discrete and latent diffusion models to generate and complete 3D scenes consisting of multiple objects, represented as semantic voxel grids.
Due to the computational challenges posed by voxel grids, DiffInDScene~\citep{DBLP:conf/cvpr/JuHL00024}, PDD~\citep{DBLP:conf/eccv/LiuLLQLY24}, $\mathcal{X}^3$~\citep{DBLP:conf/cvpr/RenHZMFW24}, and LT3SD~\citep{DBLP:conf/cvpr/abs-2409-08215} use a hierarchical diffusion pipeline to generate large-scale and fine-grained 3D scenes efficiently.
SemCity~\citep{DBLP:conf/cvpr/LeeLJISY24} employs a tri-plane representation for 3D semantic scenes, allowing for generation and editing by manipulating the tri-plane space during diffusion.
NuiScene~\citep{DBLP:journals/corr/abs-2503-16375} encodes the local scene chunks into vector sets, and uses a diffusion model to generate neighboring chunks for unbounded outdoor scenes.
%
\rev{More recently, this line has evolved from static semantic scene generation toward occupancy-based world modeling, where the scene is represented as an explicit 3D or 4D semantic state that can be rolled forward over time. 
Representative examples include OccWorld~\citep{DBLP:conf/eccv/ZhengCHZDL24}, which formulates scene evolution directly in 3D occupancy space, OccSora~\citep{DBLP:journals/corr/abs-2405-20337} and Drive-OccWorld~\citep{DBLP:conf/aaai/YangMMDCQFL25}, which extend this idea to 4D occupancy generation for simulation and planning, and UniScene~\citep{DBLP:conf/cvpr/LiGLZDCZTZW0Z0Z25}, which uses semantic occupancy as a meta scene representation for unified video, LiDAR, and occupancy generation. 
DynamicCity~\citep{DBLP:conf/iclr/abs-2410-18084} further pushes this direction toward large-scale 4D scene generation by modeling long-horizon dynamic occupancy states.}

\subsection{Image-based Generation}
\label{sec:image-methods}

\begin{figure}[!t]
  \includegraphics[width=\linewidth]{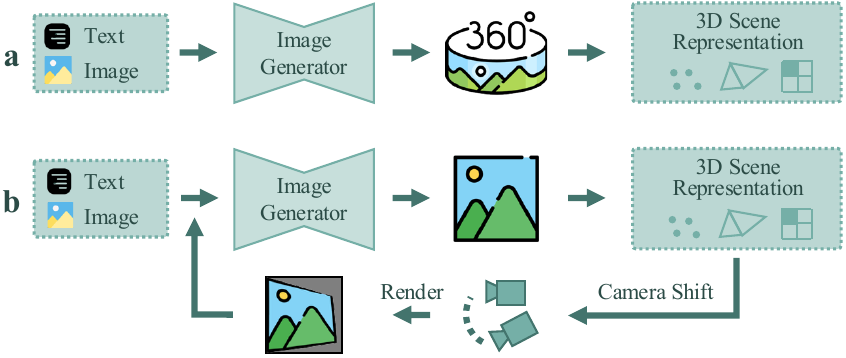}
  \caption{\textbf{The paradigms of image-based methods for 3D scene generation.} \textbf{(a)} Holistic generation creates an entire scene image in one step. \textbf{(b)} Iterative generation progressively extends the scene by extrapolating a sequence of images.}
  \label{fig:paradigm-img-based}
\end{figure}

Image-based generation bridges 2D and 3D scene synthesis by generating images while implicitly inferring underlying scene structure.
As discussed in Section~\ref{sec:paradigm-tradeoffs}, by shifting the generation process to the image domain, these methods naturally favor photorealistic appearance and diverse visual content.
As a trade-off, relying on implicit structural inference makes it challenging to maintain accurate depth, global semantic consistency, and coherent geometry across views.
The methods fall into two categories: holistic and iterative generation, as illustrated in Figure~\ref{fig:paradigm-img-based}. 
Holistic generation produces a complete scene image in a single step, while iterative generation gradually expands the scene through extrapolation, generating a sequence of images.

\subsubsection{Holistic Generation}
\label{sec:img-holi-methods}

As shown in Figure~\ref{fig:paradigm-img-based}\hyperref[fig:paradigm-img-based]{a}, holistic generation in 3D scene generation often relies on panoramic images, which provide a full 360{\textdegree} $\times$ 180{\textdegree} field of view, ensuring spatial continuity and explicit geometric constraints. 
This makes them particularly effective in mitigating scene inconsistencies that arise in perspective views.

Given an RGB image, early methods~\citep{DBLP:conf/icip/AkimotoKHA19,DBLP:conf/wacv/SumantriP20,DBLP:conf/cvpr/SomanathK21,DBLP:conf/aaai/HaraMH21,DBLP:journals/pami/HaraMH23,DBLP:conf/eccv/OhCCPWY22} use GANs for image outpainting to fill masked regions in panoramas.
More recent approaches employ advanced generative models (\eg, CoModGAN~\citep{DBLP:conf/iclr/ZhaoCSDLCX21} and VQGAN~\citep{DBLP:conf/cvpr/EsserRO21}) for greater diversity and content control.
ImmerseGAN~\citep{DBLP:conf/3dim/DastjerdiHEKL22} leverages CoModGAN for user-controlled generation.
OmniDreamer~\citep{DBLP:conf/cvpr/AkimotoMA22} and Dream360~\citep{DBLP:journals/tvcg/AiCLCMZKHW24} use VQGAN to generate diverse and high-resolution panoramas.
Leveraging advances in latent diffusion models (LDM)~\citep{DBLP:conf/cvpr/RombachBLEO22}, PanoDiffusion~\citep{DBLP:conf/iclr/WuZC24} enhances scene structure awareness by integrating depth into a bi-modal diffusion framework.

Text-to-image models (\eg, CLIP~\citep{DBLP:conf/icml/RadfordKHRGASAM21}, LDM~\citep{DBLP:conf/cvpr/RombachBLEO22}) enable text-driven panorama generation.
Text2Light~\citep{DBLP:journals/tog/ChenWL22} uses CLIP for text-based generation and hierarchical samplers to extract and piece together panoramic patches based on the input text.
Some approaches~\citep{DBLP:conf/nips/LeeKKS23,DBLP:conf/icml/Bar-TalYLD23} leverage diffusion models to generate high-resolution planar panoramas.
However, they fail to guarantee the continuity at image boundaries, which is essential in creating a seamless viewing experience. 
To address this, MVDiffusion~\citep{DBLP:conf/nips/TangZCWF23}, DiffCollage~\citep{DBLP:conf/cvpr/ZhangSHC023}, and CubeDiff~\citep{DBLP:journals/corr/abs-2501-17162} generate multi-view consistent images and align them into a closed-loop panorama for smooth transitions.
StitchDiffusion~\citep{DBLP:conf/wacv/WangXFX24}, Diffusion360~\citep{DBLP:journals/corr/abs-2311-13141}, PanoDiff~\citep{DBLP:conf/mm/WangCLX023}, and PanFusion~\citep{DBLP:conf/cvpr/ZhangWGH0O024} adopt padding and cropping strategies at boundaries to maintain the continuity.

Recent methods extend single-view panorama generation to multi-view for immersive scene exploration, following two main strategies: one directly generates multi-view panoramic images with diffusion models~\citep{DBLP:conf/nips/YeJ00HZOHZ024}, while the other applies 3D reconstruction (\eg, surface reconstruction~\citep{DBLP:journals/corr/abs-2305-10853,DBLP:conf/3dim/abs-2310-03602,DBLP:conf/cvpr/SchultTHWWMLWWH24}, NeRF~\citep{DBLP:journals/pami/WangWCWLL24}, and 3D Gaussian Splatting~\citep{DBLP:conf/eccv/ZhouFXCCBYWK24,DBLP:conf/ijcai/MaZJ24,DBLP:journals/corr/abs-2407-15187,DBLP:journals/corr/abs-2408-13711,DBLP:journals/corr/abs-2408-13252}) as post-processing.
In this context, LayerPano3D~\citep{DBLP:journals/corr/abs-2408-13252} breaks the generated panorama into depth-based layers, filling in unseen content to help create complex scene hierarchies.

Another research direction focuses on generating geometrically consistent street-view panoramas from satellite images. 
Some methods~\citep{DBLP:conf/cvpr/LuLCOPQ20, DBLP:journals/pami/ShiCYL22, DBLP:journals/tmm/WuTJZQSY23} integrate geometric priors into GAN-based frameworks to learn cross-view mappings.
Others~\citep{DBLP:conf/iccv/LiLCQPO21, DBLP:conf/iccv/QianXX023, DBLP:conf/eccv/XuQ24} infer 3D structure from satellite images and synthesize textures for street-view panoramas.

\subsubsection{Iterative Generation}
\label{sec:img-iter-methods}

As shown in Figure~\ref{fig:paradigm-img-based}\hyperref[fig:paradigm-img-based]{b}, iterative generation starts with an initial 2D image, either provided by the user or generated from text prompts. 
To generate large-scale 3D scenes, these methods progressively extrapolate the scene along a predefined trajectory. 
By expanding and refining content step by step, they continuously optimize the 3D scene representation, enhancing geometric and structural coherence.

Given a single image, early methods infer 3D scene representations and use them to render novel views. 
These representations include point clouds~\citep{DBLP:journals/tog/NiklausMYL19,DBLP:conf/cvpr/WilesGS020,DBLP:conf/iccv/RockwellF021,DBLP:conf/aaai/KohABTWLYBA23}, multi-plane images~\citep{DBLP:conf/cvpr/TuckerS20,DBLP:conf/nips/HabtegebrialJGS20}, depth maps~\citep{DBLP:conf/cvpr/ShihSKH20}, and meshes~\citep{DBLP:conf/iccv/HuRBP21}.
Despite enabling fast rendering, these representations limit camera movement due to their finite spatial extent.
To enable unrestricted camera movement, Infinite Nature~\citep{DBLP:conf/iccv/LiuM0SJK21}, InfiniteNature-Zero~\citep{DBLP:conf/eccv/LiWSK22}, Pathdreamer~\citep{DBLP:conf/iccv/KohLYBA21}, and SGAM~\citep{DBLP:conf/nips/ShenMW22} follow a ``render-refine-repeat'' manner, iteratively warping previous views and outpainting missing regions.
DiffDreamer~\citep{DBLP:conf/iccv/CaiCPSOGW23} improves multi-view consistency by conditioning on multiple past and future frames using a diffusion model.
Rather than using explicit 3D representations, GFVS~\citep{DBLP:conf/iccv/RombachEO21} and LOTR~\citep{DBLP:conf/cvpr/RenW22} encode images and camera poses directly, using transformers to generate novel views.
\cite{DBLP:conf/cvpr/TsengL0A0023}, Photoconsistent-NVS~\citep{DBLP:conf/iccv/YuFDB23}, and ODIN~\citep{DBLP:conf/nips/WallingfordBKRD24} improve long-term view synthesis consistency with a pose-guided diffusion model.
CAT3D~\citep{DBLP:conf/nips/GaoHHBMSBP24} uses a multi-view LDM to generate novel views from input images, followed by 3D reconstruction for interactive rendering.
Similarly, Bolt3D~\citep{DBLP:journals/corr/abs-2503-14445} generates scene appearance and geometry through multi-view diffusion but directly outputs 3D Gaussians to avoid time-consuming optimization.

Text-driven scene generation boosts diversity and controllability by leveraging pretrained text-to-image diffusion models~\citep{DBLP:conf/cvpr/RombachBLEO22,DBLP:conf/iccv/ZhangRA23}. 
Without requiring extensive domain-specific training, these methods iteratively shift the camera view, outpaint images based on text prompts.
PanoGen~\citep{DBLP:conf/nips/LiB23}, AOG-Net~\citep{DBLP:conf/aaai/LuHW0024}, PanoFree~\citep{DBLP:conf/eccv/LiuLCLXP24}, OPa-Ma~\citep{DBLP:journals/corr/abs-2407-10923}, and Invisible Stitch~\citep{DBLP:conf/3dim/abs-2404-19758} iteratively outpaint images in perspective view and seamlessly stitch them into a panoramic scene.
Other approaches leverage depth estimator~\citep{DBLP:journals/pami/RanftlLHSK22,DBLP:journals/corr/abs-2302-12288,DBLP:conf/cvpr/MiangolehDMPA21} 
to merge RGB images into a unified 3D scene. 
SceneScape~\citep{DBLP:conf/nips/FridmanAKD23}, Text2Room~\citep{DBLP:conf/iccv/HolleinCO0N23}, and iControl3D~\citep{DBLP:conf/mm/LiWCPWXWCL24} use 3D meshes as an intermediary proxy to fuse diffusion-generated images into a coherent 3D scene representation iteratively.
WonderJourney~\citep{DBLP:conf/cvpr/YuDHSRFCSS0H24} adopts a point cloud representation and leverages a VLM-guided re-generation strategy to ensure visual fidelity.
Text2NeRF~\citep{DBLP:journals/tvcg/ZhangLWWL24} and 3D-SceneDreamer~\citep{DBLP:conf/cvpr/ZhangZ0MHBXZ24} adopt NeRF-based representations to mitigate error accumulation in geometry and appearance, improving adaptability across scenarios.
Scene123~\citep{DBLP:journals/corr/abs-2408-05477} further enhances photorealism by using a GAN framework, where the discriminator compares outputs from the video generator with those from the scene generator.
By introducing 3D Gaussian Splatting~\citep{DBLP:journals/tog/KerblKLD23}, LucidDreamer~\citep{DBLP:journals/corr/abs-2311-13384}, 
WonderWorld~\citep{DBLP:journals/corr/abs-2406-09394}, RealmDreamer~\citep{DBLP:journals/corr/abs-2404-07199}, WonderTurbo~\citep{DBLP:journals/corr/abs-2504-02261}, and Hunyuan-World~\citep{DBLP:journals/corr/abs-2507-21809, DBLP:journals/corr/abs-2604-14268} adopt 3D Gaussians as 3D scene representations for higher quality and faster rendering.
\rev{%
Leveraging recent advancements in powerful large 3D foundation models~\citep{DBLP:conf/cvpr/abs-2409-12957,DBLP:conf/cvpr/abs-2412-01506}, SynCity~\citep{DBLP:journals/corr/abs-2503-16420} enables training-free generation of high-quality 3D scenes by iteratively performing image outpainting, 3D object generation, and stitching.
An alternative line of work~\citep{DBLP:conf/cvpr/abs-2412-03558, DBLP:journals/corr/abs-2511-16624} generates 3D scenes through object-centric composition, constructing scenes by assembling multiple generated instances while preserving spatial relationships and global coherence.}

Another research direction conducts iterative view synthesis and image animation simultaneously to build a dynamic 3D scene from a single image.
3D Cinemagraphy~\citep{DBLP:conf/cvpr/Li0S0XL23} and Make-It-4D~\citep{DBLP:conf/mm/ShenLSPX0L23} use layered depth images (LDIs) to build feature point clouds and animate scenes via motion estimation and 3D scene flow. 
3D-MOM~\citep{DBLP:conf/iclr/jincjhkkk25} first optimizes 3D Gaussians by generating multi-view images from a single image, and then optimizes 4D Gaussians~\citep{DBLP:conf/cvpr/WuYFX0000W24} by estimating consistent motion across views.

\subsection{Video-based Generation}
\label{sec:video-methods}

\begin{figure}[!t]
  \includegraphics[width=\linewidth]{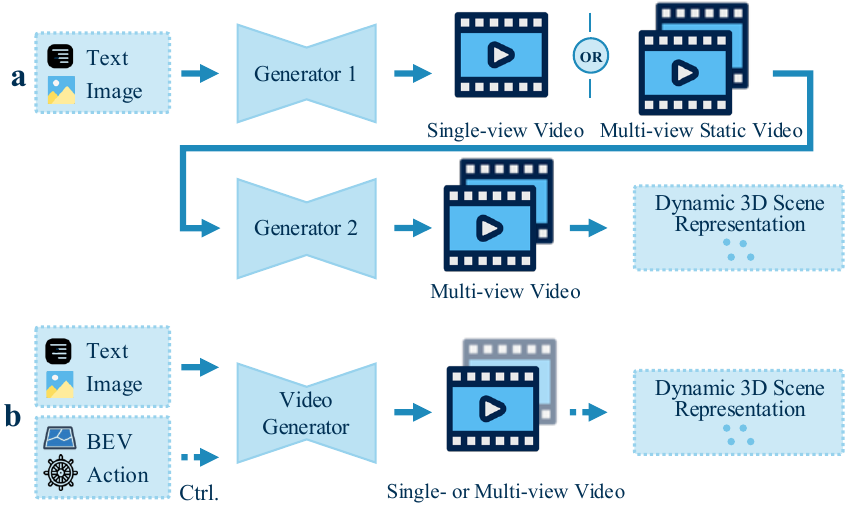}
  \caption{\textbf{The paradigms of video-based methods for 3D scene generation.} 
  \textbf{(a)} Two-stage methods employ two generators, at least one being a video generator, to synthesize multi-view videos while utilizing dynamic 3D scene representations for consistency and exploration.
 \textbf{(b)} One-stage methods produce single or multi-view videos within a unified process and model, optionally optimizing dynamic 3D scene representations.
 Note that dashed arrows denote optional operations. ``Ctrl.'' stands for ``Control''.}
  \label{fig:paradigm-video-based}
\end{figure}

Recent advances in video diffusion models~\citep{DBLP:journals/corr/abs-2402-17177,DBLP:journals/csur/XingFCDHXWJ25} have demonstrated significant progress in generating high-quality video content. 
Building on these advancements, video-based 3D scene generation synthesizes sequences of images, enabling the creation of immersive and dynamic environments.
As discussed in Section~\ref{sec:paradigm-tradeoffs}, temporal coherence enhances perceptual realism and motion consistency, but when scenes are generated primarily as sequences of views, implicit 3D modeling often leads to accumulated geometric drift and limited long-range global consistency.
These methods can be divided into two-stage and one-stage categories, with their paradigms illustrated in Figure~\ref{fig:paradigm-video-based}.

\subsubsection{Two-stage Generation}
\label{sec:vid-two-methods}

As shown in Figure~\ref{fig:paradigm-video-based}\hyperref[fig:paradigm-video-based]{a}, two-stage generation divides the generation into two stages, each targeting multi-view spatial consistency and multi-frame temporal coherence separately.
To further improve view consistency, these generated sequences are subsequently used to optimize a dynamic 3D scene representation (\eg, 4D Gaussians~\citep{DBLP:conf/cvpr/WuYFX0000W24}, Deformable Gaussians~\citep{DBLP:conf/cvpr/YangGZJ0024}).
%
VividDream~\citep{DBLP:journals/corr/abs-2405-20334} first constructs a static 3D scene through iterative image outpainting, then renders multi-view videos covering the entire scene and applies time-reversal~\citep{DBLP:conf/eccv/FengDXNABZ24} to animate them, creating dynamic videos across viewpoints.
%
PaintScene4D~\citep{DBLP:journals/corr/abs-2412-04471} first generates a video from a text description using video diffusion, then refines it through iterative warping and inpainting at each timestamp to maintain multi-view consistency.
Similarly, 4Real~\citep{DBLP:conf/nips/YuWZMSCJT024}, DimensionX~\citep{DBLP:journals/corr/abs-2411-04928}, and Free4D~\citep{DBLP:journals/corr/abs-2503-20785} first generate a coherent reference video, then expand viewpoints using frame-conditioned video generation.

\subsubsection{One-stage Generation}
\label{sec:vid-one-methods}

As shown in Figure~\ref{fig:paradigm-video-based}\hyperref[fig:paradigm-video-based]{b}, one-stage generation consolidates generation into a single process, implicitly capturing spatio-temporal consistency to produce single- or multi-view videos from any viewpoint and timestep within a unified model.
Some approaches~\citep{DBLP:journals/corr/abs-2409-02048, DBLP:journals/cvpr/abs-2504-02764, DBLP:conf/cvpr/abs-2412-12091, DBLP:conf/cvpr/abs-2504-01956} adopt video diffusion models for iterative view extrapolation, followed by 3DGS optimization to build a static scene.
To generate dynamic scenes, 
Gen$\mathcal{X}$D~\citep{DBLP:journals/corr/abs-2411-02319} and CAT4D~\citep{DBLP:conf/cvpr/abs-2411-18613} adopt distinct multiview-temporal strategies to construct multi-view video models capable of generating all views across all timestamps.
StarGen~\citep{DBLP:journals/corr/abs-2501-05763} and Streetscapes~\citep{DBLP:conf/siggraph/Deng0LGSW24} use past frames as guidance for video generation, enhancing long-range scene synthesis through an autoregressive approach.
%
By utilizing the natural multi-view 3D prior of panoramic images, 4K4DGen~\citep{DBLP:conf/iclr/abs-2406-13527} samples perspective images from a static panorama, animates them, and aligns them into a dynamic panorama.
360DVD~\citep{DBLP:conf/cvpr/WangLMCZ24}, Imagine360~\citep{DBLP:journals/corr/abs-2412-03552}, Genex~\citep{DBLP:journals/corr/abs-2411-11844}, and DynamicScaler~\citep{DBLP:conf/cvpr/abs-2412-11100} integrate panoramic constraints into video diffusion models to generate spherical-consistent panoramic videos.

%
\rev{In interactive domains such as video games and autonomous driving, one-stage video generation is further developed into scene-centric world modeling, where control signals such as user actions, BEV maps, bounding boxes, or driver commands govern the temporal evolution of environments.}
In open-world gaming environments, vast datasets comprising user inputs and rendered videos enable models like DIAMOND~\citep{DBLP:conf/nips/AlonsoJMKSPF24}, GameNGen~\citep{DBLP:conf/iclr/abs-2408-14837}, Oasis~\citep{oasis2024}, GameGen-$\mathbb{X}$~\citep{DBLP:journals/corr/abs-2411-00769}, and WORLDMEM~\citep{DBLP:journals/corr/abs-2504-12369} to predict future frames based on user interactions, creating responsive virtual environments as neural game engines.
In autonomous driving, models such as DriveDreamer~\citep{DBLP:conf/eccv/WangZHCZL24}, MagicDrive~\citep{DBLP:conf/iclr/0001CXHLY024}, Drive-WM~\citep{DBLP:conf/cvpr/00010FLC024}, and GAIA-1~\citep{DBLP:journals/corr/abs-2309-17080} utilize inputs like text, bounding boxes, Bird's Eye View (BEV) maps, and driver actions to control video generation for complex driving scenarios.
\rev{Recent extensions push this line of work toward stronger world modeling capabilities, including improved view consistency~\citep{DBLP:conf/eccv/LiZY24, DBLP:conf/cvpr/abs-2412-13188, DBLP:conf/cvpr/WenZLJWLZWS024, DBLP:journals/corr/abs-2503-15208, DBLP:journals/corr/abs-2503-20523}, richer control interfaces~\citep{DBLP:conf/nips/GaoY0CQ0Z024, DBLP:journals/corr/abs-2408-00415, DBLP:conf/aaai/ZhaoWZCHB025, DBLP:conf/cvpr/HassanSRRHBKZCS25}, explicit 3D-level control via occupancy~\citep{DBLP:conf/eccv/LuHYZZ24, DBLP:journals/corr/abs-2412-03934,  DBLP:journals/corr/abs-2411-11252}, multimodal output~\citep{DBLP:journals/corr/abs-2407-05679, DBLP:journals/corr/abs-2412-01407, DBLP:conf/cvpr/abs-2503-13265}, faster generation~\citep{DBLP:conf/cvpr/abs-2409-05463}, and longer sequences~\citep{DBLP:journals/corr/abs-2406-01349, DBLP:journals/corr/abs-2411-13807, DBLP:journals/corr/abs-2412-19505, DBLP:conf/cvpr/NiGLCLW25}.}

%% file: tables/representative-methods.tex
\definecolor{crIndoor}{RGB}{255, 200, 196}
\definecolor{crNature}{RGB}{197, 225, 222}
\definecolor{crUrban}{RGB}{250, 234, 219}
\definecolor{crNone}{RGB}{238, 237, 234}    
\definecolor{crText}{RGB}{243, 240, 161}    
\definecolor{crImage}{RGB}{247, 173, 190}   
\definecolor{crImageTD}{RGB}{240, 179, 210} 
\definecolor{crCstrn}{RGB}{191, 155, 203}   
\definecolor{crMotion}{RGB}{202, 210, 219}  
\definecolor{crGraph}{RGB}{112, 206, 234}   
\definecolor{crVolume}{RGB}{160, 215, 209}  
\definecolor{crSmMap}{RGB}{219, 210, 201}   
\definecolor{crLiDAR}{RGB}{139, 218, 225}   
\definecolor{crBBox}{RGB}{239, 208, 151}    
\definecolor{crCamera}{RGB}{162, 180, 156}  
\definecolor{crAction}{RGB}{240, 212, 209}  

\begin{table*}[!t]
    \setlength\extrarowheight{1pt}
    \centering
    \caption{\textbf{Summary and comparison of representative works for 3D scene generation}. 
    The table compares scene types (\eg, \crbx{crIndoor}{I} indoor, \crbx{crNature}{N} nature, \crbx{crUrban}{U} urban) and conditioning modalities (\eg, \crbx{crNone}{X} unconditioned, \crbx{crText}{T} text, \crbx{crImage}{I} image, \crbx{crImageTD}{I\textsubscript{D}} top-down image, \crbx{crCstrn}{C} constraint, \crbx{crMotion}{M} motion, \crbx{crGraph}{G} scene graph, \crbx{crVolume}{V} semantic volume, \crbx{crSmMap}{S\textsubscript{D}} semantic map, \crbx{crLiDAR}{L} LiDAR, \crbx{crBBox}{B} bounding box, \crbx{crCamera}{P} camera pose, \crbx{crAction}{A} user action) across various 3D scene representations.
    Note that ``Optm.'', ``Gen.'', ``Rep.'', and ``Seq.'' are short for ``Optimization'', ``Generative'', ``Representation'', and ``Sequence'', respectively.}
    \label{tab:representative-methods}
    \resizebox{\linewidth}{!}{
    \begin{tabular}{llccccccc}
    \toprule
      \multicolumn{2}{l}{\bf{Category}}      & \bf{Method} & 
      \bf{Venue}                             & \bf{Gen. Model} & 
      \bf{Scene Type}                        & \bf{Condition} & 
      \bf{3D Scene Rep.} \\
    \midrule 
      \multirow{11}{*}{\rotatebox{90}{Procedural}} &
      \multirow{4}{*}{Rule-based}
        & \cite{DBLP:conf/siggraph/MusgraveKM89} &
        SIGGRAPH'89                          & Procedural &
        \crbx{crNature}{N}                   & \crbx{crNone}{X} & 
        Mesh \\
        & & CityEngine~\citep{DBLP:conf/siggraph/ParishM01} &
        SIGGRAPH'01                          & Procedural & 
        \crbx{crUrban}{U}                    & \crbx{crImageTD}{I\textsubscript{D}} & 
        Mesh \\
        & & ~\cite{DBLP:journals/tog/CordonnierGGBGP17} & 
        TOG'17                               & Procedural & 
        \crbx{crNature}{N}                   & \crbx{crVolume}{V} & 
        Mesh \\
        & & Infinigen~\citep{DBLP:conf/cvpr/RaistrickLMMWZK23} &
        CVPR'23                              & Procedural & 
        \crbx{crNature}{N}                   & \crbx{crNone}{X} & 
        Mesh \\
      \cline{2-8}
      & \multirow{4}{*}{Optm.-based} 
        & Make it home~\citep{DBLP:journals/tog/YuYTTCO11} &
        TOG'11                               & Procedural & 
        \crbx{crIndoor}{I}                   & \crbx{crNone}{X} & 
        Mesh \\
        & & ~\cite{DBLP:journals/cgf/WuFLW18} &
        CGF'18                               & Procedural & 
        \crbx{crIndoor}{I}                   & \crbx{crCstrn}{C} & 
        Mesh \\
        & & ProcTHOR\citep{DBLP:conf/nips/DeitkeVHWESHKKM22} &
        NeurIPS'22                           & Procedural &
        \crbx{crIndoor}{I}                   & \crbx{crImageTD}{I\textsubscript{D}} & 
        Mesh \\
        & & Infinigen Indoors~\citep{DBLP:conf/cvpr/RaistrickMKY0HW24} &
        CVPR'24                              & Procedural & 
        \crbx{crIndoor}{I}                   & \crbx{crCstrn}{C} & 
        Mesh \\
      \cline{2-8}
      & \multirow{3}{*}{LLM-based} 
        & LayoutGPT~\citep{DBLP:conf/nips/FengZFJAHBWW23} &
        NeurIPS'23                           & Procedural & 
        \crbx{crIndoor}{I}                   & \crbx{crText}{T} & 
        Mesh \\
        & & 3D-GPT\citep{DBLP:conf/3dim/abs-2310-12945} 
        & 3DV'25                             & Procedural & 
        \crbx{crNature}{N}                   & \crbx{crText}{T} & 
        Mesh \\
        & & Scene$\mathcal{X}$~\citep{DBLP:conf/aaai/abs-2403-15698} 
        & AAAI'25                            & Procedural & 
        \crbx{crNature}{N}\crbx{crUrban}{U}  & \crbx{crText}{T} & 
        Mesh \\
    \midrule 
      \multirow{21}{*}{\rotatebox{90}{Neural 3D-based}} &
      \multirow{4}{*}{Scene Parameters}
        & DeepSynth~\citep{DBLP:journals/tog/WangSCR18} &
        TOG'18                               & GAN & 
        \crbx{crIndoor}{I}                   & \crbx{crImageTD}{I\textsubscript{D}} & 
        Mesh \\
        & & ATISS~\citep{DBLP:conf/nips/PaschalidouKSKG21} &
        NeurIPS'21                           & Autoregressive & 
        \crbx{crIndoor}{I}                   & \crbx{crImageTD}{I\textsubscript{D}} & 
        Mesh \\
        & & MIME~\citep{DBLP:conf/cvpr/YiHTHTB23} &
        CVPR'23                              & Autoregressive & 
        \crbx{crIndoor}{I}                   & \crbx{crImageTD}{I\textsubscript{D}}+\crbx{crMotion}{M} & 
        Mesh \\
        & & DiffuScene~\citep{DBLP:conf/cvpr/TangNMDTN24} &
        CVPR'24                              & Diffusion & 
        \crbx{crIndoor}{I}                   & \crbx{crText}{T} & 
        Mesh \\
      \cline{2-8}
      & \multirow{5}{*}{Scene Graph}
        & PlanIT~\citep{DBLP:journals/tog/WangLWSCR19} &
        TOG'19                               & Autoregressive & 
        \crbx{crIndoor}{I}                   & \crbx{crImageTD}{I\textsubscript{D}} & 
        Mesh \\
        & & GRAINS~\citep{DBLP:journals/tog/LiPXCKSTCCZ19} &
        TOG'19                               & VAE & 
        \crbx{crIndoor}{I}                   & \crbx{crNone}{X} & 
        Mesh \\
        & & Graph-to-3D~\citep{DBLP:conf/iccv/DhamoMNT21} &
        ICCV'21                              & VAE & 
        \crbx{crIndoor}{I}                   & \crbx{crGraph}{G} & 
        SDF \\
        & & CommonScenes~\citep{DBLP:conf/nips/ZhaiOWDTNB23} & 
        NeurIPS'23                           & Diffusion & 
        \crbx{crIndoor}{I}                   & \crbx{crGraph}{G} & 
        Mesh \\
        & & InstructScene~\citep{DBLP:conf/iclr/LinM24} 
        & ICLR'24                            & Diffusion & 
        \crbx{crIndoor}{I}                   & \crbx{crText}{T} & 
        Mesh \\
      \cline{2-8}
      & \multirow{7}{*}{Semantic Layout}
        & GANcraft~\citep{DBLP:conf/iccv/HaoMB021} & 
        ICCV'21                              & GAN & 
        \crbx{crNature}{N}                   & \crbx{crVolume}{V} & 
        NeRF \\
        & & CC3D~\citep{DBLP:conf/iccv/BahmaniPPYWGT23} &
        ICCV'23                              & GAN & 
        \crbx{crIndoor}{I}\crbx{crUrban}{U}  & \crbx{crSmMap}{S\textsubscript{D}} & 
        NeRF \\
        & & InfiniCity~\citep{DBLP:conf/iccv/LinLMCS0T23} &
        ICCV'23                              & GAN & 
        \crbx{crUrban}{U}                    & \crbx{crNone}{X} & 
        NeRF \\
        & & SceneDreamer~\citep{DBLP:journals/pami/ChenWL23} &
        TPAMI'23                             & GAN & 
        \crbx{crNature}{N}                   & \crbx{crNone}{X} & 
        NeRF \\
        & & CityDreamer~\citep{DBLP:conf/cvpr/Xie0H024} &
        CVPR'24                              & GAN & 
        \crbx{crUrban}{U}                    & \crbx{crNone}{X} & 
        NeRF \\
        & & Comp3D\citep{DBLP:conf/3dim/PoW24} &
        3DV'24                               & Diffusion & 
        \crbx{crNature}{N}                   & \crbx{crText}{T}+\crbx{crVolume}{V} & 
        NeRF \\
        & & BlockFusion~\citep{DBLP:journals/tog/WuLYSSWCLSLJ24} &
        TOG'24                               & Diffusion & 
        \crbx{crIndoor}{I}\crbx{crNature}{N}\crbx{crUrban}{U} & 
        \crbx{crSmMap}{S\textsubscript{D}} & 
        SDF \\
      \cline{2-8}
      & \multirow{5}{*}{Implicit Layout}
        & GSN~\citep{DBLP:conf/iccv/DeVries0STS21} & 
        ICCV'21                              & GAN & 
        \crbx{crIndoor}{I}                   & \crbx{crNone}{X}/\crbx{crImage}{I} & 
        NeRF \\
        & & GAUDI~\citep{DBLP:conf/nips/0001GATTCDZGUDS22} &
        NeurIPS'22                           & Diffusion & 
        \crbx{crIndoor}{I}                   & 
        \crbx{crNone}{X}/\crbx{crText}{T}/\crbx{crImage}{I} & 
        NeRF \\
        & & NeuralField-LDM~\citep{DBLP:conf/cvpr/0001BYKSLR0F23} &
        CVPR'23                              & Diffusion & 
        \crbx{crUrban}{U}                    & \crbx{crNone}{X}/\crbx{crImage}{I} & 
        NeRF \\
        & & $\mathcal{X}^3$~\citep{DBLP:conf/cvpr/RenHZMFW24} 
        & CVPR'24                            & VAE\&Diffusion & 
        \crbx{crUrban}{U}                    & \crbx{crNone}{X}/\crbx{crLiDAR}{L} & 
        Voxel Grid \\
        & & Director3D~\citep{DBLP:conf/nips/LiLXQCZ0J24} 
        & NeurIPS'24                         & Diffusion & 
        \crbx{crIndoor}{I}\crbx{crNature}{N}\crbx{crUrban}{U} & \crbx{crText}{T} 
        & 3D Gaussians\\
    \midrule 
      \multirow{13}{*}{\rotatebox{90}{Image-based}} &
      \multirow{5}{*}{Holistic}
        & ImmerseGAN~\citep{DBLP:conf/3dim/DastjerdiHEKL22} &
        3DV'22                               & GAN & 
        \crbx{crIndoor}{I}\crbx{crNature}{N}\crbx{crUrban}{U} & \crbx{crText}{T}/\crbx{crImage}{I} & 
        Image Seq. \\
        & & MVDiffusion~\citep{DBLP:conf/nips/TangZCWF23} &
        NeurIPS'23                           & Diffusion & 
        \crbx{crIndoor}{I}\crbx{crNature}{N}\crbx{crUrban}{U} & \crbx{crText}{T} & 
        Image Seq. \\
        & & PanFusion~\citep{DBLP:conf/cvpr/ZhangWGH0O024} & 
        CVPR'24                              & Diffusion & 
        \crbx{crIndoor}{I}\crbx{crNature}{N}\crbx{crUrban}{U} & \crbx{crText}{T} & 
        Image Seq. \\
        & & PERF~\citep{DBLP:journals/pami/WangWCWLL24} & 
        TPAMI'24                             & Diffusion & 
        \crbx{crIndoor}{I}                   & \crbx{crImage}{I} & 
        NeRF \\
        & & LayerPano3D~\citep{DBLP:journals/corr/abs-2408-13252} & 
        SIGGRAPH'25                          & Diffusion & 
        \crbx{crIndoor}{I}\crbx{crNature}{N}\crbx{crUrban}{U} & \crbx{crText}{T} & 
        3D Gaussians\\
      \cline{2-8}
      & \multirow{8}{*}{Iterative}
        & PixelSynth~\citep{DBLP:conf/iccv/RockwellF021} & 
        ICCV'21                              & VAE & 
        \crbx{crIndoor}{I}                   & \crbx{crImage}{I} & 
        Point Cloud\\
        & & GFVS~\citep{DBLP:conf/iccv/RombachEO21} & 
        ICCV'21                              & GAN & 
        \crbx{crIndoor}{I}\crbx{crNature}{N} & \crbx{crImage}{I} & 
        Image Seq. \\
        & & Infinite Nature~\citep{DBLP:conf/iccv/LiuM0SJK21} & 
        ICCV'21                              & GAN & 
        \crbx{crNature}{N}                   & \crbx{crImage}{I} & 
        Image Seq. \\
        & & 3D Cinemagraphy~\citep{DBLP:conf/cvpr/Li0S0XL23} & 
        CVPR'23                              & GAN & 
        \crbx{crNature}{N}                   & \crbx{crImage}{I} & 
        Point Cloud \\
        & & Text2Room~\citep{DBLP:conf/iccv/HolleinCO0N23} & 
        ICCV'23                              & Diffusion & 
        \crbx{crIndoor}{I}                   & \crbx{crText}{T}/\crbx{crImage}{I} & 
        Mesh \\
        & & Text2NeRF~\citep{DBLP:journals/tvcg/ZhangLWWL24} & 
        CVPR'24                              & Diffusion & 
        \crbx{crIndoor}{I}\crbx{crNature}{N}\crbx{crUrban}{U} & \crbx{crText}{T}/\crbx{crImage}{I} & 
        NeRF \\
        & & WonderJourney~\citep{DBLP:conf/cvpr/YuDHSRFCSS0H24} & 
        CVPR'24                              & Diffusion & 
        \crbx{crIndoor}{I}\crbx{crNature}{N}\crbx{crUrban}{U} & \crbx{crText}{T}/\crbx{crImage}{I} & 
        Point Cloud \\
        & & LucidDreamer~\citep{DBLP:journals/corr/abs-2311-13384} & 
        TVCG'25                              & Diffusion & 
        \crbx{crIndoor}{I}\crbx{crNature}{N}\crbx{crUrban}{U} & \crbx{crText}{T}/\crbx{crImage}{I} & 
        3D Gaussians\\
    \midrule 
      \multirow{7}{*}{\rotatebox{90}{Video-based}} &
      \multirow{2}{*}{Two-stage}
        & 4Real~\citep{DBLP:conf/nips/YuWZMSCJT024} & 
        NeurIPS'24                           & Diffusion & 
        \crbx{crIndoor}{I}\crbx{crNature}{N}\crbx{crUrban}{U} & \crbx{crText}{T} & 
        3D Gaussians\\
        & & DimensionX~\citep{DBLP:journals/corr/abs-2411-04928} & 
        ICCV'25                              & Diffusion & 
        \crbx{crIndoor}{I}\crbx{crNature}{N}\crbx{crUrban}{U} & \crbx{crText}{T}/\crbx{crImage}{I} & 
        3D Gaussians \\
      \cline{2-8}
      & \multirow{5}{*}{One-stage}
        & MagicDrive~\citep{DBLP:conf/iclr/0001CXHLY024} &
        ICLR'24                              & Diffusion & 
        \crbx{crUrban}{U}                    & 
        \crbx{crText}{T}+\crbx{crSmMap}{S\textsubscript{D}}+\crbx{crBBox}{B}+\crbx{crCamera}{P} & 
        Image Seq. \\
        & & Vista~\citep{DBLP:conf/nips/GaoY0CQ0Z024} & 
        NeurIPS'24                           & Diffusion & 
        \crbx{crUrban}{U}                    & 
        \crbx{crText}{T}/\crbx{crImage}{I}/\crbx{crAction}{A} & 
        Image Seq. \\
        & & Gen$\mathcal{X}$D~\citep{DBLP:journals/corr/abs-2411-02319} & 
        ICLR'25                              & Diffusion & 
        \crbx{crIndoor}{I}\crbx{crNature}{N}\crbx{crUrban}{U} & \crbx{crImage}{I}/\crbx{crCamera}{P} & 
        Image Seq. \\
        & & 4K4DGen~\citep{DBLP:conf/iclr/abs-2406-13527} & 
        ICLR'25                              & Diffusion & 
        \crbx{crNature}{N}\crbx{crUrban}{U}  & \crbx{crImage}{I} & 
        3D Gaussians \\
        & & GameGen-$\mathbb{X}$~\citep{DBLP:journals/corr/abs-2411-00769} & 
        ICLR'25                              & Diffusion & 
        \crbx{crNature}{N}\crbx{crUrban}{U}  & \crbx{crText}{T}/\crbx{crAction}{A} & 
        Image Seq. \\
    \bottomrule
    \end{tabular}
    }
\end{table*}

%% file: sections/datasets.tex
\section{Datasets and Evaluation Metrics}
\label{sec:datasets-and-evaluation}

\input{tables/representative-dataset}

\subsection{Datasets}
\label{sec:datasets}

We summarize the commonly used datasets for 3D scene generation in Table~\ref{tab:comp-dataset}, grouping them by scene type into three categories: indoor, natural, and urban.

\subsubsection{Indoor Datasets}
\label{sec:indoor-datasets}

Existing indoor datasets are either captured from real-world scenes using RGB or RGB-D sensors or professionally designed with curated 3D CAD furniture models.

Real-world datasets are captured from physical scenes using sensors like depth, DSLR, or panoramic cameras. 
Early datasets provide RGB-D or panoramic images with semantic labels (\eg, NYUv2~\citep{DBLP:conf/eccv/SilbermanHKF12}, 2D-3D-S~\citep{DBLP:journals/corr/ArmeniSZS17}), 
while recent ones like ScanNet~\citep{DBLP:conf/cvpr/DaiCSHFN17} and Matterport3D~\citep{DBLP:conf/3dim/ChangDFHNSSZZ17} offer 3D reconstructions with dense meshes and instance-level annotations.

\begin{itemize}
\item \textbf{SUN360}~\citep{DBLP:conf/cvpr/XiaoEOT12} contains 67,583 high-res 360\textdegree $\times$ 180\textdegree ~panoramic images in equirectangular format, manually categorized into 80 scene types.

\item \textbf{NYUv2}~\citep{DBLP:conf/eccv/SilbermanHKF12} provides 1,449 densely annotated RGB-D images from 464 indoor scenes, covering per-pixel semantics and instance-level objects.

\item \textbf{SUN-RGBD}~\citep{DBLP:conf/cvpr/SongLX15} offers 10,335 RGB-D images and reconstructed point cloud, with rich annotations including room types, 2D polygons, 3D bounding boxes, and camera poses.

\item \textbf{SceneNN}~\citep{DBLP:conf/3dim/HuaPNTYY16} offers 502K RGB-D frames from 100 indoor scenes with reconstructed meshes, textured models, camera poses, and both object-oriented and axis-aligned bounding boxes.

\item \textbf{2D-3D-S}~\citep{DBLP:journals/corr/ArmeniSZS17} includes over 70,000 panoramic images from six indoor areas, with aligned depth, surface normals, semantic labels, point clouds, meshes, global XYZ maps, and full camera metadata.

\item \textbf{Laval Indoor}~\citep{DBLP:journals/tog/GardnerSYSGGL17} offers 2.2K high-res indoor panoramas (7768$\times$3884) with HDR lighting from various settings such as homes, offices, and factories.

\item \textbf{Matterport3D}~\citep{DBLP:conf/3dim/ChangDFHNSSZZ17} contains 10,800 panoramic images from 194,400 RGB-D views in 90 buildings, with dense camera trajectories, aligned depth maps, and semantic labels.

\item \textbf{ScanNet}~\citep{DBLP:conf/cvpr/DaiCSHFN17} offers 2.5M RGB-D frames in 1,513 scans from 707 distinct spaces with camera poses, surface reconstructions, dense 3D semantic labels, and aligned CAD models.

\item \textbf{Replica}~\citep{DBLP:journals/corr/abs-1906-05797} provides high-quality 3D reconstructions of 35 rooms across 18 scenes, featuring PBR textures, HDR lighting, and semantic annotations.

\item \textbf{RealEstate10K}~\citep{DBLP:journals/tog/ZhouTFFS18} contains 10 million frames from 10K YouTube videos, featuring both indoor and outdoor scenes with per-frame camera parameters.

\item \textbf{3DSSG}~\citep{DBLP:conf/cvpr/WaldDNT20} provides scene graphs for 478 indoor rooms from 3RScan~\citep{DBLP:conf/iccv/WaldANTN19}, with 93 object attributes, 40 relationship types, and 534 semantic classes.

\item \textbf{HM3D}~\citep{DBLP:conf/nips/RamakrishnanGWM21} offers 1,000 high-res 3D reconstructions of indoor spaces across residential, commercial, and civic buildings.

\item \textbf{ScanNet++}~\citep{DBLP:conf/iccv/YeshwanthLND23} includes 1,000+ scenes captured with laser scanner, DSLR, and iPhone RGB-D, featuring fine-grained semantics and long-tail categories.

\item \textbf{DL3DV-10K}~\citep{DBLP:conf/cvpr/LingSTZXWYGYLLS24} contains 51.2M frames from 10,510 video sequences across 65 indoor and semi-outdoor locations, featuring varied visual conditions such as reflections and different lighting.
\end{itemize}

Synthetic indoor datasets overcome real-world limitations like limited diversity, occlusion, and costly annotation. 
Using designed layouts and textured 3D assets, datasets like SUNCG~\citep{DBLP:conf/cvpr/SongYZCSF17} and 3D-FRONT~\citep{DBLP:conf/iccv/FuC0ZWLZSJZ021} offer large-scale, diverse scenes. 
Some~\citep{DBLP:conf/eccv/ZhengZLTGZ20,DBLP:conf/iccv/RobertsRRK0PWS21} leverage advanced rendering for photorealistic images with accurate 2D labels.

\begin{itemize}
\item \textbf{SceneSynth}~\citep{DBLP:journals/tog/FisherRSFH12} includes 130 indoor scenes (\eg, studies, kitchens, living rooms) with 1,723 models from Google 3D Warehouse.

\item \textbf{SUNCG}~\citep{DBLP:conf/cvpr/SongYZCSF17} offers 45,622 manually designed scenes, featuring 404K rooms and 5.7M objects from 2,644 meshes across 84 categories.

\item \textbf{Structured3D}~\citep{DBLP:conf/eccv/ZhengZLTGZ20} includes 196.5K images from 3.5K professionally designed houses with 3D annotations (\eg, lines, planes).

\item \textbf{Hypersim}~\citep{DBLP:conf/iccv/RobertsRRK0PWS21} provides 77.4K photorealistic renders with PBR materials and lighting for realistic view synthesis.

\item \textbf{3D-FRONT}~\citep{DBLP:conf/iccv/FuC0ZWLZSJZ021} offers 6,813 professionally designed houses and 18,797 diversely furnished rooms, populated with high-quality textured 3D objects from 3D-FUTURE~\citep{DBLP:journals/ijcv/FuJGGZMT21}.

\item \textbf{SG-FRONT}~\citep{DBLP:conf/nips/ZhaiOWDTNB23} augments 3D-FRONT with scene graph annotations.
\end{itemize}

\subsubsection{Natural Datasets}
\label{sec:natural-datasets}

Datasets for natural scenes are still limited, mainly due to the difficulties of large-scale collection and annotation in open outdoor environments. However, several notable efforts have been made to advance research in this area.

\begin{itemize}

\item \textbf{Laval Outdoor}~\citep{DBLP:conf/cvpr/Hold-GeoffroyAL19} provides 205 high-res HDR panoramas of diverse natural and urban scenes.

\item \textbf{LHQ}~\citep{DBLP:conf/iccv/SkorokhodovSE21} offers 91,693 curated landscape images from Unsplash and Flickr, designed for high-quality image generation tasks.

\item \textbf{ACID}~\citep{DBLP:conf/iccv/LiuM0SJK21} features 2.1M drone-captured frames from 891 YouTube videos of coastal regions, with 3D camera trajectories obtained via structure-from-motion.
\end{itemize}

\subsubsection{Urban Datasets}
\label{sec:urban-datasets}

Urban datasets are built from real-world imagery or synthesized using game engines, providing images and annotations in 2D or 3D.

Real-world datasets mainly focus on driving scenes, represented by KITTI~\citep{DBLP:conf/cvpr/GeigerLU12}, Waymo~\citep{DBLP:conf/cvpr/SunKDCPTGZCCVHN20}, and nuScenes~\citep{DBLP:conf/cvpr/CaesarBLVLXKPBB20}, due to the significant attention autonomous driving has received over the past decade. 
Another major source is Google's street views and aerial views, exemplified by HoliCity~\citep{DBLP:journals/corr/abs-2008-03286} and GoogleEarth~\citep{DBLP:conf/cvpr/Xie0H024}. 
These datasets provide rich annotations, such as semantic segmentation and instance segmentation.

\begin{itemize}

\item \textbf{KITTI}~\citep{DBLP:conf/cvpr/GeigerLU12}, captured in Karlsruhe, includes stereo and optical flow pairs, 39.2 km of visual odometry, and 200K+ 3D object annotations, using a Velodyne LiDAR, GPS/IMU, and a stereo camera rig with grayscale and color cameras.

\item \textbf{SemanticKITTI}~\citep{DBLP:conf/iccv/BehleyGMQBSG19} extends KITTI with dense point-wise semantics for full 360\textdegree LiDAR scans.

\item\textbf{KITTI-360}~\citep{DBLP:journals/pami/LiaoXG23} extends KITTI with 73.7 km of driving, 150K images, 1B 3D points, and dense 2D/3D labels, using two 180\textdegree fisheye side cameras, a front stereo camera, and two LiDARs.

\item \textbf{Cityscapes}~\citep{DBLP:conf/cvpr/CordtsORREBFRS16} provides street-view videos from 50 cities, with 5K pixel-level and 20K coarse annotations.

\item \textbf{Waymo}~\citep{DBLP:conf/cvpr/SunKDCPTGZCCVHN20} offers 1M frames from 1,150 20-second scenes (6.4 hour total) with 12M 3D and 9.9M 2D boxes, collected in San Francisco, Mountain View, and Phoenix using 5 LiDARs and 5 high-res pinhole cameras.

\item \textbf{nuScenes}~\citep{DBLP:conf/cvpr/CaesarBLVLXKPBB20} provides 1.4M images and 390K LiDAR sweeps from 1,000 20s scenes in Boston and Singapore, using 6 cameras, 1 LiDAR, 5 RADARs, GPS, and IMU, with 3D box tracking for 23 classes.

\item \textbf{HoliCity}~\citep{DBLP:journals/corr/abs-2008-03286} aligns 6,300 high-res panoramas (13312$\times$6656) with CAD models of downtown London for image-CAD fusion.

\item \textbf{OmniCity}~\citep{DBLP:conf/cvpr/LiLXXYHXL23} provides 100K+ pixel-annotated street, satellite, and panorama images from 25K locations in New York City.

\item \textbf{GoogleEarth}~\citep{DBLP:conf/cvpr/Xie0H024} offers 24K New York images from 400 Google Earth\footnote{\url{https://earth.google.com/studio}} trajectories with 2D/3D semantic and instance masks plus camera parameters.

\item \textbf{OSM} dataset~\citep{DBLP:conf/cvpr/Xie0H024}, sourced from Open Street Map\footnote{\url{https://openstreetmap.org}}, provides bird's eye view semantic maps, height fields, and vector data of roads, buildings, and land use across 80+ cities worldwide.
\end{itemize}

Real-world annotations are costly and viewpoint-limited. 
Synthetic datasets like CARLA~\citep{DBLP:conf/corl/DosovitskiyRCLK17} and CityTopia~\citep{DBLP:journals/corr/abs-2501-08983}, built in game engines, provide diverse street and drone views with rich 2D/3D annotations.

\begin{itemize}
\item \textbf{CARLA}~\citep{DBLP:conf/corl/DosovitskiyRCLK17} is a simulator based on Unreal Engine, offering diverse urban environments, sensor simulations (camera, LiDAR, radar), and customizable driving scenarios with control over weather, lighting, traffic, and pedestrian behaviors, enabling unlimited rendering of RGB images and corresponding 2D/3D annotations.

\item \textbf{CarlaSC}~\citep{DBLP:journals/ral/WilsonSFZCJBG22} offers 43.2K frames of semantic scenes from 24 sequences across 8 maps, captured by virtual LiDAR sensors in the CARLA simulator under varying traffic conditions.

\item \textbf{Virtual-KITTI-2}~\citep{DBLP:journals/corr/abs-2001-10773} replicates 5 KITTI sequences using Unity, offering photorealistic video under varying conditions with dense annotations for depth, segmentation, optical flow, and object tracking.

\item \textbf{CityTopia}~\citep{DBLP:journals/corr/abs-2501-08983} provides 37.5K photorealistic frames with 2D/3D annotations from procedural cities in Unreal Engine, featuring varied lighting and aerial/street-view perspectives.
\end{itemize}

\subsection{Evaluation Metrics}
\label{sec:evaluation-metrics}

Evaluating 3D scene generation methods is essential for comparing different methods across different domains. 
Various metrics have been proposed to assess key aspects of generated scenes, including geometric accuracy, structural consistency, visual realism, diversity, and physical plausibility.
This section summarizes and discusses commonly used evaluation metrics in 3D scene generation, highlighting their relevance to different generation paradigms and focuses.

\subsubsection{Metrics-based Evaluation}
\label{sec:metrics-evaluation}

\noindent{\textbf{Fidelity}}
is commonly evaluated using metrics originally developed for image and video generation, which assess the visual quality and realism of rendered outputs such as NeRFs, 3D Gaussians, or image sequences.
Distribution-based metrics, most notably Fréchet Inception Distance (FID)~\citep{DBLP:conf/nips/HeuselRUNH17} and Kernel Inception Distance (KID)~\citep{DBLP:conf/iclr/BinkowskiSAG18}, are the de facto standard for fidelity evaluation.
These metrics measure statistical distances between feature distributions extracted from a pre-trained Inception network.
Inception Score (IS)~\citep{DBLP:conf/nips/SalimansGZCRCC16} is also used, although it is known to be less reliable due to its sensitivity to classifier confidence and limited coverage of mode diversity.
Several variants~\citep{DBLP:conf/iclr/MorozovVB21,DBLP:conf/nips/SteinCHSRVLCTL23,DBLP:conf/iclr/KynkaanniemiKAA23} replace the Inception feature space with alternative backbones, including self-supervised or multimodal encoders, to improve correlation with human perception.
Perceptual no-reference metrics, such as NIQE~\citep{DBLP:journals/tip/MittalMB12} and BRISQUE~\citep{DBLP:journals/spl/MittalSB13}, estimate image quality directly from natural image statistics and are occasionally used as complementary signals.
For 3D-specific fidelity assessment, F3D~\citep{DBLP:conf/eccv/LiuLLQLY24} extends FID by operating on 3D feature representations learned by a 3D convolutional autoencoder.
In addition, set-level metrics including MMD, Coverage (COV)~\citep{DBLP:conf/icml/AchlioptasDMG18}, and 1-NNA~\citep{DBLP:conf/iclr/Lopez-PazO17} are often adopted to analyze distributional alignment, coverage, and mode collapse.

\noindent \textbf{Spatial Consistency}
metrics evaluate the geometric correctness and multi-view alignment of generated 3D scenes.
They are typically defined in terms of depth and camera pose errors.
Depth accuracy is assessed by comparing predicted depth maps with pseudo or reconstructed ground-truth, using metrics such as L2 distance, RMSE, and Scale-Invariant RMSE (SI-RMSE)~\citep{DBLP:conf/nips/EigenPF14}.
Camera pose consistency is evaluated by measuring the deviation between estimated and reference camera trajectories, where standard Structure-from-Motion pipelines (e.g., COLMAP~\citep{DBLP:conf/cvpr/SchonbergerF16}) are commonly used for pose estimation.

\noindent \textbf{Temporal Coherence}
assesses the consistency of generated scenes across time, which is particularly important for dynamic scenes and video-based outputs.
Among existing metrics, Fr{\'e}chet Video Distance (FVD)~\citep{DBLP:conf/iclr/UnterthinerSKMM19} is the most widely adopted, extending FID with spatiotemporal video features to jointly capture frame quality and temporal coherence.
Flow warping error (FE)~\citep{DBLP:conf/eccv/LaiHWSYY18} measures temporal stability by evaluating optical-flow-based frame alignment and is often used as a complementary low-level metric.
To better capture complex motion patterns, Fr{\'e}chet Video Motion Distance (FVMD)~\citep{DBLP:journals/corr/abs-2407-16124} introduces explicit motion features derived from keypoint tracking and compares their distributions using the Fr{\'e}chet distance.

\noindent \textbf{Controllability}
evaluates how effectively the generated results respond to user-specified conditions or inputs.
CLIP Score~\citep{DBLP:conf/emnlp/HesselHFBC21} is the most commonly used metric for text-conditioned generation, measuring the semantic alignment between generated images and conditioning text using a pre-trained CLIP model.
Higher CLIP scores indicate better adherence to user prompts, although the metric primarily captures semantic consistency rather than fine-grained structural control.

\noindent \textbf{Diversity}
reflects the ability of a generative model to produce varied and non-collapsing outputs.
Diversity is often evaluated by comparing the distribution of generated scenes with that of real or training data.
Category distribution KL divergence (CKL)~\citep{DBLP:conf/cvpr/Ritchie0L19} measures the discrepancy between object category distributions, where lower divergence indicates higher diversity.
Scene Classification Accuracy (SCA)~\citep{DBLP:conf/cvpr/Ritchie0L19} assesses whether a classifier can distinguish generated scenes from real ones, indirectly reflecting distributional coverage and potential mode collapse.

\noindent \textbf{Plausibility}
measures whether generated scenes obey basic physical, geometric, and semantic constraints.
This aspect is commonly evaluated using rule-based or task-specific metrics.
Collision rate quantifies the proportion of objects involved in physical intersections, while Out-of-Bounds Object Area (OBA) measures the extent to which objects violate scene boundaries.
Such metrics are typically domain-dependent and are used as complementary indicators of physical and semantic validity rather than general-purpose evaluation measures.

\begin{table*}[!t]
  \input{tables/3d-front}
\end{table*}

\subsubsection{Benchmark-based Evaluation}
\label{sec:benchmark-evaluation}

To promote fair, reproducible, and comprehensive evaluation of diverse 3D scene generation methods, recent research has increasingly embraced standardized benchmark suites that integrate multiple metrics, task configurations, and quality dimensions. 
This trend marks a shift from relying only on isolated quantitative metrics to adopting more holistic, task-aligned evaluations that better reflect the complexity of real-world applications.

\noindent\textbf{Q-Align}~\citep{DBLP:conf/icml/0001ZZCLLGW0SYM24} adopts large multi-modal models (LMMs) to predict visual quality scores that align with human judgment. 
It covers three core dimensions: Image Quality Assessment (IQA), Image Aesthetic Assessment (IAA), and Video Quality Assessment (VQA). 
During inference, the mean opinion scores are collected and re-weighted to obtain the LMM-predicted score.

\noindent\textbf{VideoScore}~\citep{DBLP:journals/corr/abs-2406-15252} enables video quality evaluation by training on a large-scale human-feedback dataset. It provides assessment across five aspects: Visual Quality (VQ), Temporal Consistency (TC), Dynamic Degree (DD), Text-to-Video Alignment (TVA), and Factual Consistency (FC).

\noindent\textbf{VBench}~\citep{DBLP:conf/cvpr/HuangHYZS0Z0JCW24} and \textbf{VBench++}~\citep{DBLP:journals/corr/abs-2411-13503} are comprehensive and versatile benchmark suites for video generation. 
They comprise 16 dimensions in video generation (\eg, subject identity inconsistency, motion smoothness, temporal flickering, and spatial relationship, etc).
\textbf{VBench-2.0}~\citep{DBLP:journals/corr/abs-2503-21755} further addresses more complex challenges associated with intrinsic faithfulness, including commonsense reasoning, physics-based realism, human motion, and creative composition. 

\noindent\textbf{WorldScore}~\citep{DBLP:journals/corr/abs-2504-00983} unifies the evaluation of 3D, 4D, and video models on their ability to generate a world following instructions.
It formulates the evaluation of 3D scene generation as a sequence of next-scene generation tasks guided by camera trajectories, jointly measures controllability, quality, and dynamics in various fine-grained features.

\subsubsection{Human Evaluation}
\label{sec:human-evaluation}

User studies remain an essential component for capturing subjective qualities of 3D scene generation that are difficult to quantify through automated metrics, such as visual appeal, realism, and perceptual coherence. 

Participants are typically asked to rank or rate generated scenes based on multiple aspects, including photorealism, aesthetics, input alignment (\eg, text or layout), 3D consistency across views, and physical or semantic plausibility. 
Ideally, participants should include both domain experts (\eg, 3D artists, designers, researchers) and normal users. 
Their feedback offers complementary perspectives: experts may provide more critical and structured insights, while non-experts better reflect general user impressions. 

Although human evaluations are resource-intensive and inherently subjective, they provide essential qualitative insights that complement other evaluation methods by capturing human preferences in real-world contexts. 
Platforms like Prolific\footnote{\url{https://www.prolific.com}} and Amazon Mechanical Turk (AMT)\footnote{\url{https://www.mturk.com}} facilitate the recruitment of diverse participants and enable efficient scaling of user studies.

%% file: tables/representative-dataset.tex
\definecolor{crIndoor}{RGB}{255, 200, 196}
\definecolor{crNature}{RGB}{197, 225, 222}
\definecolor{crUrban}{RGB}{250, 234, 219}
\definecolor{crPCG}{RGB}{250, 198, 205}
\definecolor{crN3D}{RGB}{247, 233, 218}
\definecolor{crImg}{RGB}{170, 206, 201}
\definecolor{crVdo}{RGB}{187, 226, 243}
\definecolor{crSem}{RGB}{202, 200, 229}      
\definecolor{crIns}{RGB}{243, 240, 161}      
\definecolor{crBBox}{RGB}{239, 208, 151}     
\definecolor{crSG}{RGB}{191, 155, 203}       
\definecolor{crCamera}{RGB}{162, 180, 156}   
\definecolor{crMotion}{RGB}{160, 215, 209}   

\begin{table*}[!t]
  \rowcolors{2}{gray!3}{gray!12}
  \setlength\tabcolsep{4pt}
  \setlength\extrarowheight{2pt}
  \caption{\textbf{Summary and comparison of popular datasets for 3D scene generation.} The scene types \crbx{crIndoor}{I}, \crbx{crNature}{N}, and \crbx{crUrban}{U} represent ``Indoor'', ``Nature'', and ``Urban''. The annotations \crbx{crSG}{G}, \crbx{crCamera}{P}, and \crbx{crMotion}{M} denote scene graph, camera pose, and motion annotations (\eg, optical flow), respectively. \crbx{crSem}{\tf{S}{2}}/\crbx{crIns}{\tf{I}{2}}/\crbx{crBBox}{\tf{B}{2}} and \crbx{crSem}{\tf{S}{3}}/\crbx{crIns}{\tf{I}{3}}/\crbx{crBBox}{\tf{B}{3}} represent 2D/3D semantic maps, instance maps, and bounding boxes. \crbx{crPCG}{P}, \crbx{crN3D}{N}, \crbx{crImg}{I}, and \crbx{crVdo}{V} indicate procedural, neural 3D-based, image-based, and video-based generation, respectively. \ts{Mesh}{R} and \ts{PCD}{R} are reconstructed mesh and point clouds, respectively. Note that ``-'' in \#Images, 3D Model, and Annotations indicates datasets do not provide these; ``-'' in \#Scenes and Area means the information cannot be inferred.}
  \label{tab:comp-dataset}
  \resizebox{\linewidth}{!}{
  \begin{tabular}{lcccccccccc}
    \toprule
    \bf{Dataset}    & 
    \bf{Year}       & \bf{Type} & 
    \bf{Source}     & \bf{\#Images} & 
    \bf{\#Scenes}   & \bf{Area} &
    \bf{3D Model}   & \bf{Annotations} &
    \bf{Used by}    &
    \bf{URL} \\
    \midrule
    SUN360~\citep{DBLP:conf/cvpr/XiaoEOT12} &
    2012            & \crbx{crIndoor}{I}\crbx{crNature}{N}\crbx{crUrban}{U} &  
    Real            & 67.5K              &
    -               & -                  &
    -               & -                  &
    \crbx{crImg}{I} &  \\
    NYUv2~\citep{DBLP:conf/eccv/SilbermanHKF12} &
    2012            & \crbx{crIndoor}{I} &  
    Real            & 1.4K               & 
    464             & -                  &
    -               & 
    \crbx{crSem}{\tf{S}{2}}\crbx{crIns}{\tf{I}{2}} &
    \crbx{crN3D}{N} &
    \lnk{https://cs.nyu.edu/~fergus/datasets/nyu_depth_v2.html} \\
    Sun-RGBD~\citep{DBLP:conf/cvpr/SongLX15} &
    2015            & \crbx{crIndoor}{I} &  
    Real            & 10.3K              &
    -               & -                  &
    \ts{PCD}{R}     & 
    \crbx{crSem}{\tf{S}{2}}\crbx{crBBox}{\tf{B}{3}} &
    \crbx{crN3D}{N} &
    \lnk{https://rgbd.cs.princeton.edu/} \\
    SceneNN~\citep{DBLP:conf/3dim/HuaPNTYY16} &
    2016            & \crbx{crIndoor}{I} &  
    Real            & 502K               &
    100             & 2,260 \ts{m}{2}    &
    \ts{Mesh}{R}    & 
    \crbx{crSem}{\tf{S}{3}}\crbx{crIns}{\tf{I}{3}}\crbx{crBBox}{\tf{B}{3}}\crbx{crCamera}{P} &
    \crbx{crN3D}{N} &
    \lnk{https://hkust-vgd.github.io/scenenn/} \\
    2D-3D-S~\citep{DBLP:journals/corr/ArmeniSZS17} &
    2017            & \crbx{crIndoor}{I} &  
    Real            & 70.5K              &
    270             & 6,000 \ts{m}{2}    &
    \ts{Mesh}{R}, \ts{PCD}{R} & 
    \crbx{crSem}{\tf{S}{2}}\crbx{crIns}{\tf{I}{2}}\crbx{crSem}{\tf{S}{3}}\crbx{crIns}{\tf{I}{3}}\crbx{crCamera}{P} &
    \crbx{crImg}{I} &
    \lnk{https://github.com/alexsax/2D-3D-Semantics} \\
    Laval Indoor~\citep{DBLP:journals/tog/GardnerSYSGGL17} &
    2017            & \crbx{crIndoor}{I} &  
    Real            & 2.2K               &
    -               & -                  &
    -               & -                  &
    \crbx{crImg}{I} &
    \lnk{http://hdrdb.com/indoor/} \\
    Matterport3D~\citep{DBLP:conf/3dim/ChangDFHNSSZZ17} &
    2017            & \crbx{crIndoor}{I} &  
    Real            & 10.8K              &
    90              & 0.102 k\ts{m}{2}   &
    \ts{Mesh}{R}    & 
    \crbx{crSem}{\tf{S}{2}}\crbx{crIns}{\tf{I}{2}}\crbx{crSem}{\tf{S}{3}}\crbx{crIns}{\tf{I}{3}}\crbx{crCamera}{P} &
    \crbx{crN3D}{N}\crbx{crImg}{I} &
    \lnk{https://niessner.github.io/Matterport/} \\
    ScanNet~\citep{DBLP:conf/cvpr/DaiCSHFN17} &
    2017            & \crbx{crIndoor}{I} & 
    Real            & 2.5M               & 
    707             & 39,980 \ts{m}{2}   &
    \ts{Mesh}{R}, CAD & 
    \crbx{crSem}{\tf{S}{2}}\crbx{crIns}{\tf{I}{2}}\crbx{crSem}{\tf{S}{3}}\crbx{crIns}{\tf{I}{3}}\crbx{crCamera}{P} &
    \crbx{crN3D}{N} &
    \lnk{http://www.scan-net.org/} \\
    RealEstate10K~\citep{DBLP:journals/tog/ZhouTFFS18} & 
    2018            & \crbx{crIndoor}{I}\crbx{crUrban}{U} &  
    Real            & 10M                &
    -               & -                  &
    -               & \crbx{crCamera}{P} &
    \crbx{crN3D}{N}\crbx{crImg}{I}\crbx{crVdo}{V} &
    \lnk{https://google.github.io/realestate10k/} \\
    Replica~\citep{DBLP:journals/corr/abs-1906-05797} &
    2019            & \crbx{crIndoor}{I} &  
    Real            & -                  &
    18              & 2,190 \ts{m}{2}    &
    \ts{Mesh}{R}    & 
    \crbx{crSem}{\tf{S}{3}}\crbx{crIns}{\tf{I}{3}} &
    \crbx{crN3D}{N}\crbx{crImg}{I} &
    \lnk{https://github.com/facebookresearch/Replica-Dataset} \\
    3DSSG~\citep{DBLP:conf/cvpr/WaldDNT20} &
    2020            & \crbx{crIndoor}{I} &  
    Real            & 363K               &
    478             & -                  &
    \ts{Mesh}{R}    & 
    \crbx{crSem}{\tf{S}{3}}\crbx{crIns}{\tf{I}{3}}\crbx{crSG}{G}\crbx{crCamera}{P}  &
    \crbx{crN3D}{N} &
    \lnk{https://3dssg.github.io/} \\
    HM3D~\citep{DBLP:conf/nips/RamakrishnanGWM21} &
    2021            & \crbx{crIndoor}{I} &  
    Real            & -                  &
    1,000           & 0.365 k\ts{m}{2}   &
    \ts{Mesh}{R}    & -                  &
    \crbx{crImg}{I} &
    \lnk{https://aihabitat.org/datasets/hm3d/} \\
    ScanNet++~\citep{DBLP:conf/iccv/YeshwanthLND23} &
    2023            & \crbx{crIndoor}{I} &  
    Real            & 11.1M              &
    1006            & -                  &
    \ts{Mesh}{R}    & 
    \crbx{crSem}{\tf{S}{3}}\crbx{crIns}{\tf{I}{3}}\crbx{crCamera}{P} &
    \crbx{crN3D}{N} &
    \lnk{https://kaldir.vc.in.tum.de/scannetpp/} \\
    DL3DV-10K~\citep{DBLP:conf/cvpr/LingSTZXWYGYLLS24} & 
    2024            & \crbx{crIndoor}{I}\crbx{crNature}{N}\crbx{crUrban}{U} &  
    Real            & 51.2M              &
    -               & -                  &
    -               & \crbx{crCamera}{P} &
    \crbx{crN3D}{N}\crbx{crVdo}{V} &
    \lnk{https://dl3dv-10k.github.io/DL3DV-10K/} \\
    \midrule
    SceneSynth~\citep{DBLP:journals/tog/FisherRSFH12} &
    2012            & \crbx{crIndoor}{I} &  
    Synthetic       & -                  &
    130             & -                  &
    CAD             & -                  &
    \crbx{crN3D}{N} &
    \lnk{https://graphics.stanford.edu/projects/scenesynth/} \\
    SUNCG~\citep{DBLP:conf/cvpr/SongYZCSF17} &
    2017            & \crbx{crIndoor}{I} &  
    Synthetic       & -                  &
    45,622          & 24 k\ts{m}{2}      &
    CAD             & 
    \crbx{crSem}{\tf{S}{3}}\crbx{crIns}{\tf{I}{3}} &
    \crbx{crPCG}{P}\crbx{crN3D}{N} & \\
    Structured3D~\citep{DBLP:conf/eccv/ZhengZLTGZ20} &
    2020            & \crbx{crIndoor}{I} &  
    Synthetic       & 196.5K             &
    3500            & -                  &
    CAD             & 
    \crbx{crSem}{\tf{S}{2}}\crbx{crIns}{\tf{I}{2}}\crbx{crSem}{\tf{S}{3}}\crbx{crIns}{\tf{I}{3}} &
    \crbx{crN3D}{N}\crbx{crImg}{I} &
    \lnk{https://structured3d-dataset.org/} \\
    Hypersim~\citep{DBLP:conf/iccv/RobertsRRK0PWS21} &
    2021            & \crbx{crIndoor}{I} &  
    Synthetic       & 77.4K              &
    461             & -                  &
    CAD             & 
    \crbx{crSem}{\tf{S}{2}}\crbx{crIns}{\tf{I}{2}}\crbx{crCamera}{P} &
    \crbx{crN3D}{N} &
    \lnk{https://github.com/apple/ml-hypersim} \\
    3D-FRONT~\citep{DBLP:conf/iccv/FuC0ZWLZSJZ021} &
    2021            & \crbx{crIndoor}{I} &  
    Synthetic       & -                  &
    6,813           & 0.51 k\ts{m}{2}    &
    CAD             & -                  &
    \crbx{crPCG}{P}\crbx{crN3D}{N} &
    \lnk{https://tianchi.aliyun.com/specials/promotio-  libaba-3d-scene-dataset} \\
    SG-FRONT~\citep{DBLP:conf/nips/ZhaiOWDTNB23} &
    2023            & \crbx{crIndoor}{I} &  
    Synthetic       & -                  &
    6,813           & 0.51 k\ts{m}{2}    &
    CAD             & \crbx{crSG}{G}     &
    \crbx{crN3D}{N} &
    \lnk{https://sites.google.com/view/commonscenes/dataset} \\
    \midrule
    Laval Outdoor~\citep{DBLP:conf/cvpr/Hold-GeoffroyAL19} & 
    2017            & \crbx{crNature}{N}\crbx{crUrban}{U} &  
    Real            & 0.2K               &
    -               & -                  &
    -               & -                  &
    \crbx{crImg}{I} &
    \lnk{http://hdrdb.com/outdoor/} \\
    LHQ~\citep{DBLP:conf/iccv/SkorokhodovSE21} & 
    2021            & \crbx{crNature}{N} &  
    Real            & 91.7K              &
    -               & -                  &
    -               & -                  &
    \crbx{crN3D}{N}\crbx{crImg}{I} &
    \lnk{https://github.com/universome/alis} \\
    ACID~\citep{DBLP:conf/iccv/LiuM0SJK21} & 
    2021            & \crbx{crNature}{N} &  
    Real            & 2.1M               &
    -               & -                  &
    -               & \crbx{crCamera}{P} &
    \crbx{crN3D}{N}\crbx{crImg}{I}\crbx{crVdo}{V} &
    \lnk{https://infinite-nature.github.io/} \\
    \midrule
    KITTI~\citep{DBLP:conf/cvpr/GeigerLU12} &
    2012            & \crbx{crUrban}{U}  &  
    Real            & 15K                &
    1               & -                  &
    LiDAR           & 
    \crbx{crSem}{\tf{S}{2}}\crbx{crIns}{\tf{I}{2}}\crbx{crBBox}{\tf{B}{3}}\crbx{crCamera}{P}\crbx{crMotion}{M} &
    \crbx{crN3D}{N}\crbx{crImg}{I} &
    \lnk{https://www.cvlibs.net/datasets/kitti/} \\
    Cityscapes~\citep{DBLP:conf/cvpr/CordtsORREBFRS16} & 
    2016            & \crbx{crUrban}{U}  &  
    Real            & 25K                &
    50              & -                  &
    -               & 
    \crbx{crSem}{\tf{S}{2}}\crbx{crIns}{\tf{I}{2}} &
    \crbx{crImg}{I} &
    \lnk{https://www.cityscapes-dataset.com/} \\
    SemanticKITTI~\citep{DBLP:conf/iccv/BehleyGMQBSG19} & 
    2019            & \crbx{crUrban}{U}  &  
    Real            & -                  &
    1               & -                  &
    LiDAR           & \crbx{crSem}{\tf{S}{3}}\crbx{crCamera}{P} &
    \crbx{crN3D}{N} &
    \lnk{https://semantic-kitti.org/} \\
    Waymo~\citep{DBLP:conf/cvpr/SunKDCPTGZCCVHN20} & 
    2020            & \crbx{crUrban}{U}  &  
    Real            & 1M                 &
    3               & 76 k\ts{m}{2}      &
    LiDAR           & 
    \crbx{crBBox}{\tf{B}{2}}\crbx{crBBox}{\tf{B}{3}}\crbx{crCamera}{P} &
    \crbx{crN3D}{N}\crbx{crVdo}{V} &
    \lnk{https://waymo.com/open/} \\
    nuScenes~\citep{DBLP:conf/cvpr/CaesarBLVLXKPBB20} & 
    2020            & \crbx{crUrban}{U}  &  
    Real            & 1.4M              &
    2               & 5 k\ts{m}{2}       &
    LiDAR           &
    \crbx{crSem}{\tf{S}{3}}\crbx{crBBox}{\tf{B}{3}}\crbx{crCamera}{P} &
    \crbx{crN3D}{N}\crbx{crVdo}{V} &
    \lnk{https://www.nuscenes.org/} \\
    HoliCity~\citep{DBLP:journals/corr/abs-2008-03286} & 
    2020            & \crbx{crUrban}{U}  & 
    Real            & 6.3K               & 
    1               & 20 k\ts{m}{2}      & 
    CAD             & 
    \crbx{crSem}{\tf{S}{2}}\crbx{crCamera}{P} &
    \crbx{crN3D}{N} &
    \lnk{https://holicity.io/} \\
    OmniCity~\citep{DBLP:conf/cvpr/LiLXXYHXL23} & 
    2023            & \crbx{crUrban}{U}  & 
    Real            & 108.6K             &
    1               & -                  &
    -               & \crbx{crSem}{\tf{S}{2}}\crbx{crIns}{\tf{I}{2}} &
    \crbx{crN3D}{N} &
    \lnk{https://city-super.github.io/omnicity/} \\
    KITTI-360~\citep{DBLP:journals/pami/LiaoXG23} & 
    2023            & \crbx{crUrban}{U}  &
    Real            & 150K               &
    1               & -                  & 
    LiDAR           & 
    \crbx{crSem}{\tf{S}{2}}\crbx{crIns}{\tf{I}{2}}\crbx{crSem}{\tf{S}{3}}\crbx{crIns}{\tf{I}{3}}\crbx{crBBox}{\tf{B}{3}}\crbx{crCamera}{P} &
    \crbx{crN3D}{N} &
    \lnk{https://www.cvlibs.net/datasets/kitti-360/} \\
    GoogleEarth~\citep{DBLP:conf/cvpr/Xie0H024} & 
    2024            & \crbx{crUrban}{U}  &  
    Real            & 24K                &
    1               & 25 k\ts{m}{2}      &
    Voxel Grid      & 
    \crbx{crSem}{\tf{S}{2}}\crbx{crIns}{\tf{I}{2}}\crbx{crSem}{\tf{S}{3}}\crbx{crIns}{\tf{I}{3}}\crbx{crCamera}{P} &
    \crbx{crN3D}{N} &
    \lnk{https://gateway.infinitescript.com/s/GoogleEarth} \\
    OSM~\citep{DBLP:conf/cvpr/Xie0H024} & 
    2024            & \crbx{crUrban}{U}  & 
    Real            & -                  &
    80              & 6,000 k\ts{m}{2}   &
    -               & 
    \crbx{crSem}{\tf{S}{2}}\crbx{crIns}{\tf{I}{2}} &
    \crbx{crPCG}{P}\crbx{crN3D}{N} &
    \lnk{https://gateway.infinitescript.com/s/OSM} \\
    \midrule
    CARLA~\citep{DBLP:conf/corl/DosovitskiyRCLK17} & 
    2017            & \crbx{crUrban}{U}  &  
    Synthetic       & $\infty$           &
    13              & -                  &
    LiDAR           & 
    \crbx{crSem}{\tf{S}{2}}\crbx{crIns}{\tf{I}{2}}\crbx{crSem}{\tf{S}{3}}\crbx{crIns}{\tf{I}{3}}\crbx{crCamera}{P}\crbx{crMotion}{M} &
    \crbx{crN3D}{N}\crbx{crImg}{I}\crbx{crVdo}{V} &
    \lnk{https://carla.org/} \\
    Virtual-KITTI-2~\citep{DBLP:journals/corr/abs-2001-10773} & 
    2020            & \crbx{crUrban}{U}  &  
    Synthetic       & 42.5K              &
    1               & -                  &
    -               & 
    \crbx{crSem}{\tf{S}{2}}\crbx{crIns}{\tf{I}{2}}\crbx{crBBox}{\tf{B}{2}}\crbx{crBBox}{\tf{B}{3}}\crbx{crCamera}{P}\crbx{crMotion}{M} &
    \crbx{crImg}{I} &
    \lnk{https://europe.naverlabs.com/research/proxy-virtual-worlds/} \\
    CarlaSC~\citep{DBLP:journals/ral/WilsonSFZCJBG22} & 
    2022            & \crbx{crUrban}{U}  &  
    Synthetic       & 43.2K              &
    8               & -                  &
    Voxel Grid      & \crbx{crSem}{\tf{S}{3}}\crbx{crCamera}{P}\crbx{crMotion}{M} &
    \crbx{crN3D}{N} &
    \lnk{https://umich-curly.github.io/CarlaSC.github.io/} \\
    CityTopia~\citep{DBLP:journals/corr/abs-2501-08983} & 
    2025            & \crbx{crUrban}{U}  &
    Synthetic       &  37.5K             &
    11              &  36 k\ts{m}{2}     &
    Voxel Grid      & 
    \crbx{crSem}{\tf{S}{2}}\crbx{crIns}{\tf{I}{2}}\crbx{crSem}{\tf{S}{3}}\crbx{crIns}{\tf{I}{3}}\crbx{crCamera}{P} &
    \crbx{crN3D}{N} &
    \lnk{https://gateway.infinitescript.com/s/CityTopia} \\
    \bottomrule
  \end{tabular}
  }
\end{table*}

%% file: tables/3d-front.tex
  \setlength\tabcolsep{4pt}
  \setlength\extrarowheight{1pt}
  \caption{\rev{\textbf{Quantitative comparison of indoor scene generation methods on 3D-FRONT.} Results are reported for bedroom, dining room, and living room scenes using FID, KID, SCA, CKL, and iRecall when available. Lower is better for FID, KID, and CKL; for SCA, values closer to 50\% are better. Comparisons are valid within each block only, since different blocks use different conditioning signals and protocols.}}
  \label{tab:3dfront}
  \resizebox{\linewidth}{!}{
  \begin{tabular}{ccccccccccccccccc}
    \toprule
      & \multirow{2}{*}{Method}
      & \multicolumn{5}{c}{Bedroom}
      & \multicolumn{5}{c}{Dining Room}
      & \multicolumn{5}{c}{Living Room} \\
      \cmidrule(lr){3-7} \cmidrule(lr){8-12} \cmidrule(lr){13-17}
      & & FID\dwar 
        & KID\textsubscript{$\times 10^{-3}$}\dwar 
        & SCA\textsubscript{\%} 
        & CKL\dwar 
        & iRecall\textsubscript{\%}\upar 
        & FID\dwar 
        & KID\textsubscript{$\times 10^{-3}$}\dwar 
        & SCA\textsubscript{\%} 
        & CKL\dwar 
        & iRecall\textsubscript{\%}\upar
        & FID\dwar 
        & KID\textsubscript{$\times 10^{-3}$}\dwar 
        & SCA\textsubscript{\%}
        & CKL\dwar 
        & iRecall\textsubscript{\%}\upar \\
    \midrule 
      \multirow{6}{*}{\rotatebox{90}{Unconditioned}}  
      & FastSynth~\citep{DBLP:conf/cvpr/Ritchie0L19} 
      & 31.89       & -         & 83.61      & 2.40       & - 
      & 51.26       & -         & 90.12      & 5.26       & -
      & 57.22       & -         & 88.21      & 6.27       & - \\
      & Sceneformer~\citep{DBLP:conf/3dim/WangYN21} 
      & 33.61       & -         & 85.38      & 1.86       & -
      & 61.08       & -         & 85.94      & 5.18       & -
      & 63.54       & -         & 90.20      & 3.13       & - \\
      & Sync2Gen~\citep{DBLP:conf/iccv/0005ZYHMZBH21} 
      & 31.07       & -         & 82.97      & 2.24       & -
      & 46.05       & -         & 88.02      & 4.96       & -
      & 48.45       & -         & 84.57      & 7.52       & - \\
      & ATISS~\citep{DBLP:conf/nips/PaschalidouKSKG21} 
      & 18.60       & -         & 61.71      & 0.78       & -
      & 38.66       & -         & 71.34      & 0.64       & -
      & 40.83       & -         & 72.66      & 0.69       & - \\
      & ScenePrior~\citep{DBLP:conf/cvpr/NieD0N23}
      & 24.88       & -         & 83.26      & 0.43       & -
      & 46.25       & -         & 89.27      & 0.58       & -
      & 44.28       & -         & 88.07      & 0.31       & - \\
      & DiffuScene~\citep{DBLP:conf/cvpr/TangNMDTN24} 
      & \bf{18.29}  & -         & \bf{53.52} & \bf{0.35}  & -
      & \bf{32.60}  & -         & \bf{55.50} & \bf{0.22}  & -
      & \bf{36.18}  & -         & \bf{57.81} & \bf{0.21}  & - \\
    \midrule 
      \multirow{4}{*}{\rotatebox{90}{Floor Plan}} 
      & ATISS~\citep{DBLP:conf/nips/PaschalidouKSKG21} 
      & 56.63       & 4.76      & 54.00      & 0.75       & -
      & 72.43       & 5.52      & 59.00      & 0.65       & -
      & 74.52       & 5.67      & 59.00      & 0.51       & - \\
      & LayoutEnhancer~\citep{DBLP:conf/siggrapha/LeimerG0M22} 
      & 42.52       & 4.24      & 52.00      & 0.53       & -
      & 69.13       & 4.92      & 55.00      & 0.53       & -
      & 68.46       & 4.86      & 58.00      & 0.52       & - \\
      & DiffuScene~\citep{DBLP:conf/cvpr/TangNMDTN24} 
      & 39.42       & 4.35      & 53.00      & 0.40       & -
      & 49.66       & 4.26      & 54.00      & 0.23       & -
      & 50.70       & 3.79      & 56.00      & 0.23       & - \\
      & Forest2Seq~\citep{DBLP:conf/eccv/SunZZLL24}
      & \bf{38.41}  & \bf{3.93} & \bf{52.00} & \bf{0.34}  & -
      & \bf{49.13}  & \bf{2.57} & \bf{53.00} & \bf{0.23}  & -
      & \bf{48.24}  & \bf{1.89} & \bf{53.00} & \bf{0.23}  & - \\
    \midrule 
      \multirow{4}{*}{\rotatebox{90}{Text}} 
      & ATISS~\citep{DBLP:conf/nips/PaschalidouKSKG21} 
      & 122.37      & 8.23      & 58.77      & -          & 44.06
      & 120.10      & 7.43      & 57.02      & -          & 36.31
      & 134.75      & 9.82      & 59.83      & -          & 32.56 \\
      & DiffuScene~\citep{DBLP:conf/cvpr/TangNMDTN24} 
      & 129.34      & 9.66      & 59.13      & -          & 53.32
      & 135.93      & 10.71     & 58.98      & -          & 39.58
      & 142.37      & 11.76     & 60.88      & -          & 37.17 \\
      & InstructScene~\citep{DBLP:conf/iclr/LinM24} 
      & 114.86      & 6.52      & 56.37      & -          & 72.71
      & 111.52      & 5.91      & 55.32      & -          & 57.21
      & 129.13      & 8.24      & 58.93      & -          & 61.47 \\
      & FreeScene~\citep{DBLP:conf/cvpr/BaiBCWLM25} 
      & \bf{108.00} & \bf{6.07} & \bf{53.16} & -          & \bf{81.40}
      & \bf{108.22} & \bf{5.23} & \bf{54.05} & -          & \bf{71.81}
      & \bf{125.32} & \bf{7.14} & \bf{55.59} & -          & \bf{75.01} \\
    \bottomrule
  \end{tabular}
  }

%% file: sections/analysis.tex
\section{\rev{Comparative Results and Analysis}}
\label{sec:analysis}

\begin{table}[!t]
  \input{tables/nuscenes}
\end{table}

\begin{table*}[!t]
  \input{tables/eval-protocols}
\end{table*}

\subsection{\rev{Quantitative Comparisons}}
\label{sec:quantitative-comparison}

\rev{Quantitative comparison in 3D scene generation is nontrivial because existing methods are often evaluated under heterogeneous settings, including different conditioning signals, generation targets, room subsets, rendering pipelines, and metric implementations. 
Consequently, even the same baseline may yield different reported numbers across papers when re-evaluated under different protocols. 
To provide a more reliable summary, we adopt a best-effort protocol-consistent comparison and group methods only when they share the same evaluation setting. 
Tables~\ref{tab:3dfront} and~\ref{tab:nuscenes} therefore support within-block comparison rather than a single global ranking across all methods.}

\rev{Table~\ref{tab:3dfront} summarizes representative indoor scene generation results on 3D-FRONT~\citep{DBLP:conf/iccv/FuC0ZWLZSJZ021} under three settings: unconditioned, floor-plan-conditioned, and text-conditioned generation. 
Within each setting, recent methods generally outperform earlier baselines under matched protocols, with clear gains in FID, KID, and CKL. 
In particular, diffusion-based methods improve realism and distribution matching in unconditioned and floor-plan-conditioned generation. 
By contrast, text-conditioned generation remains more challenging, as shown by higher FID/KID and weaker semantic plausibility metrics.}

\rev{Table~\ref{tab:nuscenes} presents a best-effort comparison of urban driving scene generation methods on nuScenes~\citep{DBLP:conf/cvpr/CaesarBLVLXKPBB20}.
Since evaluation protocols vary more widely than in indoor generation, we restrict comparison to matched blocks such as single-view and multi-view generation. 
Even so, recent methods still show clear improvements in visual fidelity and temporal consistency over earlier baselines. 
Multi-view generation remains more challenging, as it requires both temporal coherence and cross-view geometric consistency.}

\rev{While Tables~\ref{tab:3dfront} and~\ref{tab:nuscenes} provide best-effort quantitative comparisons on two of the most commonly used benchmarks, standardized evaluation remains unavailable for many other 3D scene generation problems. 
Beyond these benchmarks, existing works often rely on diverse datasets, task-specific metrics, or human evaluation, which limits direct comparability. 
Table~\ref{tab:eval-protocols} summarizes this fragmented evaluation landscape across representative scene categories.}

\subsection{\rev{Computational Cost}}
\label{sec:computational-cost}

\rev{Computational cost varies substantially across different paradigms and is an important practical factor in choosing 3D scene generation methods. In general, the main trade-offs arise from training cost, inference cost, memory usage, and scalability. We therefore focus on paradigm-level trends, which are more informative than direct method-level comparison under the diverse settings used in existing works.}

\rev{Procedural methods are usually the most scalable once the generation pipeline has been established. 
Because assets are often pre-built and scene construction is driven by reusable rules, grammars, or simulators, their runtime typically grows relatively smoothly with scene size. 
However, this advantage depends on the subtype. 
Rule-based and optimization-based methods~\citep{DBLP:conf/cvpr/RaistrickLMMWZK23, DBLP:conf/cvpr/RaistrickMKY0HW24} usually avoid expensive model training, whereas LLM-based methods introduce additional inference cost, which further depends on whether the language model is locally deployed~\citep{DBLP:conf/nips/FengZFJAHBWW23} or accessed through external APIs~\citep{DBLP:conf/3dim/abs-2310-12945}. 
More broadly, procedural methods often replace GPU-intensive learning with manual effort in rule design, asset preparation, and pipeline engineering.}

\rev{For neural 3D-based methods, computational cost is largely determined by the underlying 3D representation. 
Voxel-based methods~\citep{DBLP:conf/iclr/abs-2410-18084} are memory-intensive because computation scales with volumetric resolution, whereas NeRF-based methods~\citep{DBLP:conf/cvpr/Xie0H024} are often time-intensive due to dense ray sampling and repeated network evaluations. 
By comparison, 3D Gaussian representations~\citep{DBLP:journals/corr/abs-2406-06526} are usually more efficient in rendering and optimization. 
Methods~\citep{DBLP:journals/tog/LiPXCKSTCCZ19} that predict object parameters and rely on asset retrieval are also often cheaper at inference than methods that explicitly synthesize geometry and appearance. 
The generative model further affects cost: diffusion-based methods~\citep{DBLP:conf/cvpr/TangNMDTN24} generally incur higher inference cost than autoregressive methods~\citep{DBLP:conf/nips/PaschalidouKSKG21} because they require iterative denoising at test time.
Their scalability is often constrained by the growth of representation complexity, rendering cost, and scene resolution.}

\rev{Image-based methods often reduce training cost by reusing mature 2D backbones, and some methods~\citep{DBLP:conf/iccv/HolleinCO0N23, DBLP:conf/cvpr/YuDHSRFCSS0H24} require little or no task-specific training.
However, this does not necessarily make them cheap in practice. 
Many image-based pipelines~\citep{DBLP:conf/iccv/HolleinCO0N23} remain slow at inference because they rely on iterative outpainting, lifting, stitching, or reconstruction, often making per-scene generation cost more significant than their training cost suggests. 
Their scalability is also limited by the fact that global 3D structure is not explicitly maintained, reducing reuse across viewpoints and long trajectories.}

\rev{Video-based methods usually have even larger overall compute requirements because they must model multiple frames jointly, which increases both memory usage and training cost. 
This is especially evident in diffusion-based video generation~\citep{DBLP:conf/nips/GaoY0CQ0Z024}, where cost grows with sequence length, frame resolution, and denoising steps. 
That said, this trend is not absolute: some methods~\citep{DBLP:conf/iclr/0001CXHLY024} can still achieve relatively efficient inference under specific settings. 
Overall, video-based methods generally require more compute than image-based ones, but better support temporal coherence and dynamic scene generation.}

\subsection{\rev{Failure Analysis}}
\label{sec:critical-analysis}

\begin{figure}
    \centering
    \includegraphics[width=\linewidth]{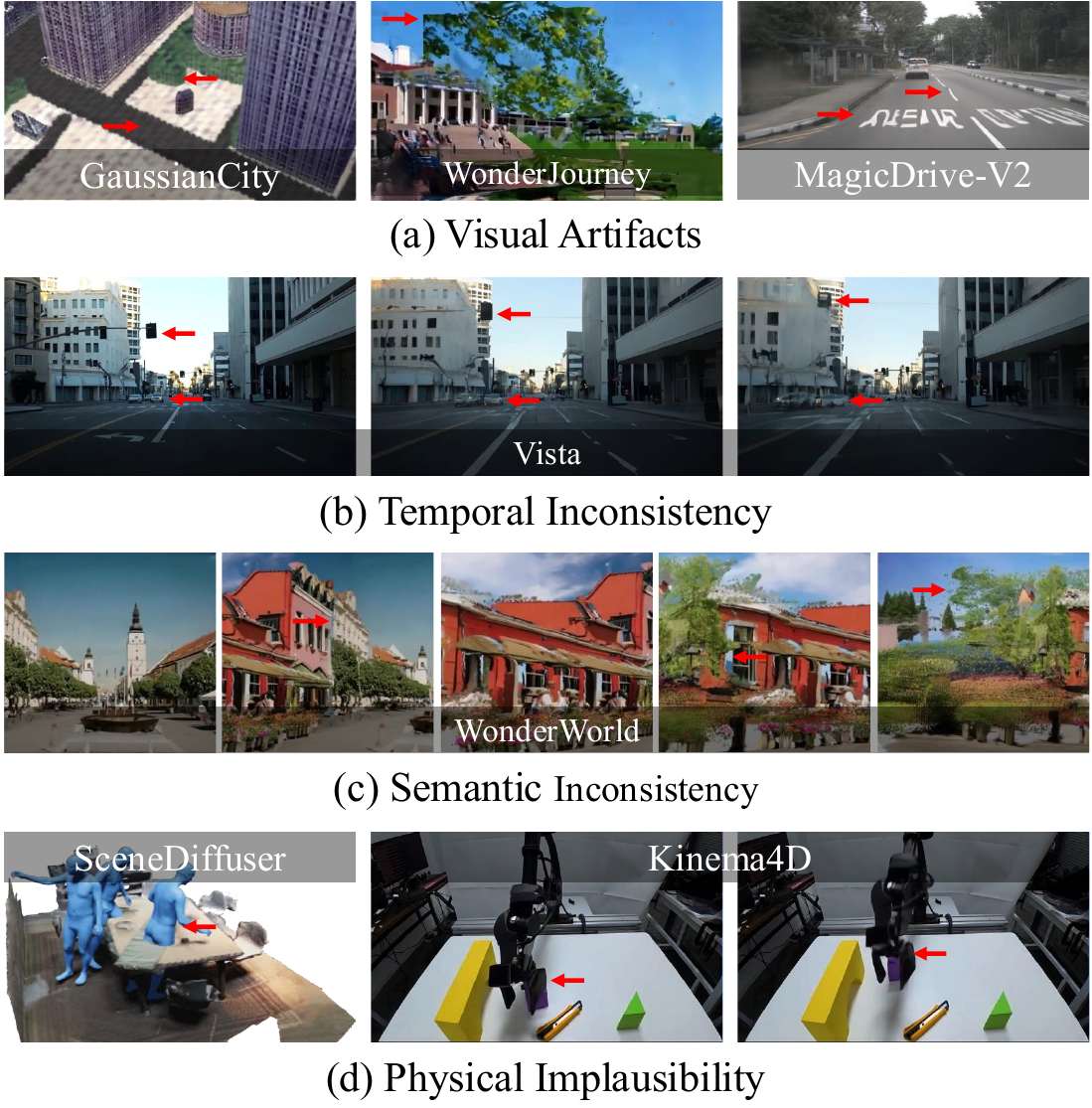}
    \caption{\rev{\textbf{Representative failure patterns in 3D scene generation.} 
    \textbf{(a) Visual artifacts:} geometric distortions and rendering artifacts, such as broken structures and blurred regions.
    \textbf{(b) Temporal inconsistency:} object drift and appearance changes across frames.
    \textbf{(c) Semantic inconsistency:} abrupt semantic shifts across frames, leading to discontinuous and implausible scene transitions.
    \textbf{(d) Physical implausibility:} violations of physical constraints, such as unstable configurations and unrealistic interactions.}}
    \label{fig:failure}
\end{figure}

\rev{We group common failure patterns into visual artifacts, temporal inconsistency, semantic inconsistency, and physical implausibility.}

\noindent 
\rev{\textbf{Visual Artifacts.}
As shown in Figure~\ref{fig:failure}\hyperref[fig:failure]{a}, visual artifacts are prevalent across most paradigms, except procedural methods with explicit modeling. 
In image-based iterative generation~\citep{DBLP:conf/cvpr/YuDHSRFCSS0H24}, they mainly arise from errors in monocular depth estimation, which are amplified under viewpoint changes and lead to distorted geometry. 
In contrast, neural 3D-based and video-based methods~\citep{DBLP:journals/corr/abs-2406-06526, DBLP:journals/corr/abs-2411-13807} exhibit artifacts from the generative process, such as blurred regions and shape distortions. 
Despite different origins, these artifacts degrade visual fidelity and indicate limitations in geometry and appearance modeling.}

\noindent 
\rev{\textbf{Temporal Inconsistency.}
As shown in Figure~\ref{fig:failure}\hyperref[fig:failure]{b}, temporal inconsistency is primarily observed in video-based methods~\citep{DBLP:conf/nips/GaoY0CQ0Z024}, where objects may drift, flicker, or change appearance across frames. 
Despite temporal modeling, maintaining stable representations over long sequences remains challenging.
This leads to unstable motion patterns and inconsistent scene evolution.}

\noindent 
\rev{\textbf{Semantic Inconsistency.}
As shown in Figure~\ref{fig:failure}\hyperref[fig:failure]{c}, semantic inconsistency is commonly observed in image-based methods~\citep{DBLP:journals/corr/abs-2406-09394}, where scene semantics change abruptly across views. 
This is largely due to the reliance on local view-conditioned generation (\eg, outpainting), which lacks global semantic constraints and long-horizon context.
As a result, generated content may exhibit discontinuous or implausible transitions in scene layout and composition.}

\noindent 
\rev{\textbf{Physical Implausibility.}
As shown in Figure~\ref{fig:failure}\hyperref[fig:failure]{d}, physical implausibility is commonly observed across generative paradigms. 
Neural 3D-based methods~\citep{DBLP:conf/cvpr/HuangWLJLZLZ23} may produce interpenetration artifacts due to inaccurate geometry or insufficient collision modeling. 
Video-based methods~\citep{DBLP:journals/corr/abs-2603-16669} often exhibit violations of physical interactions, where objects move without proper contact or force exchange. 
These issues arise from the lack of explicit modeling of physical constraints, leading to visually plausible but physically invalid scene dynamics.}

%% file: tables/nuscenes.tex
  \caption{\rev{\textbf{Quantitative comparison of urban driving scene generation methods on nuScenes.} Comparisons are valid within each block only, since different methods adopt different generation targets and evaluation settings. Note that ``Gen.'' denotes ``Generation''.}}
  \label{tab:nuscenes}
  \begin{tabularx}{\linewidth}{ccYY}
    \toprule
      & Method & FID\dwar & FVD\dwar \\
    \midrule 
      \multirow{9}{*}{\rotatebox{90}{Single-view Gen.}}
      & DriveGAN~\citep{DBLP:conf/cvpr/KimP0F21} 
      & 27.8     & 390.8 \\
      & DriveDreamer~\citep{DBLP:conf/eccv/WangZHCZL24}
      & 14.9     & 340.8 \\
      & Drive-WM~\citep{DBLP:conf/cvpr/00010FLC024}
      & 15.8     & 122.7 \\
      & WoVoGen~\citep{DBLP:conf/eccv/LuHYZZ24}
      & 27.6     & 417.7 \\
      & GenAD~\citep{DBLP:conf/eccv/ZhengSGZC24}
      & 15.4     & 184.0 \\
      & DriveDreamer-2~\citep{DBLP:conf/aaai/ZhaoWZCHB025}
      & 25.0     & 105.1 \\
      & GEM~\citep{DBLP:conf/cvpr/HassanSRRHBKZCS25} 
      & 10.5     & 158.5 \\
      & Vista~\citep{DBLP:conf/nips/GaoY0CQ0Z024}
      & 6.9      & 89.4 \\
      & ReSim~\citep{DBLP:journals/corr/abs-2506-09981}
      & \bf{5.2} & \bf{50.4} \\
    \midrule 
      \multirow{8}{*}{\rotatebox{90}{Multi-view Gen.}}
      & WoVoGen~\citep{DBLP:conf/eccv/LuHYZZ24}
      & 27.6     & 417.7 \\
      & Panacea~\citep{DBLP:conf/cvpr/WenZLJWLZWS024}
      & 17.0     & 139.0 \\
      & MagicDrive~\citep{DBLP:conf/iclr/0001CXHLY024}
      & 16.2     & 217.9 \\
      & Drive-WM~\citep{DBLP:conf/cvpr/00010FLC024}
      & 15.8     & 122.7 \\
      & MagicDrive-V2~\citep{DBLP:journals/corr/abs-2411-13807}
      & 20.9     & 94.8 \\
      & Vista~\citep{DBLP:conf/nips/GaoY0CQ0Z024}
      & 14.0     & 122.7 \\
      & UniScene~\citep{DBLP:conf/cvpr/LiGLZDCZTZW0Z0Z25}
      & \bf{6.5} & 71.9 \\
      & DiST-T~\citep{DBLP:journals/corr/abs-2503-15208}
      & 6.8      & \bf{22.7} \\
    \bottomrule
  \end{tabularx}

%% file: tables/eval-protocols.tex
\caption{\rev{\textbf{Evaluation protocols grouped by scene types.} Datasets, mainstream metrics, and representative methods are summarized for indoor, natural, and urban scene generation.}}
  \label{tab:eval-protocols}
  \begin{tabularx}{\linewidth}{cYYc}
    \toprule
      & Dataset & Mainstream Metrics
      & Representative Methods \\
    \midrule
      \multirow{15}{*}{\rotatebox{90}{Indoor}}
        & SUNCG~\citep{DBLP:conf/cvpr/SongYZCSF17}
        & User Study
        & \makecell{
            DeepSynth~\citep{DBLP:journals/tog/WangSCR18} \\
            PlanIT~\citep{DBLP:journals/tog/WangLWSCR19} \\
            GRAINS~\citep{DBLP:journals/tog/LiPXCKSTCCZ19} \\
          } \\
        \cmidrule(lr){2-4}
        & 3D-FRONT~\citep{DBLP:conf/iccv/FuC0ZWLZSJZ021}
        & FID, KID, SCA, CKL, iRecall
        & \makecell{%
            ATISS~\citep{DBLP:conf/nips/PaschalidouKSKG21} \\
            DiffuScene~\citep{DBLP:conf/cvpr/TangNMDTN24} \\ 
            InstructScene~\citep{DBLP:conf/iclr/LinM24} \\
          } \\
        \cmidrule(lr){2-4}
        & SG-FRONT~\citep{DBLP:conf/mm/SongCXKTYY23} 
        & FID, KID, MMD, COV, 1-NNA
        & \makecell{%
            CommonScenes~\citep{DBLP:conf/nips/ZhaiOWDTNB23} \\
            EchoScene~\citep{DBLP:conf/eccv/ZhaiOCLDNTB24} \\
            Planner3D~\citep{DBLP:journals/corr/abs-2403-12848} \\
          } \\
        \cmidrule(lr){2-4}
        & RealEstate10K~\citep{DBLP:journals/tog/ZhouTFFS18}
        & FID
        & \makecell{%
            SynSin~\citep{DBLP:conf/cvpr/WilesGS020}\\
            PixelSynth~\citep{DBLP:conf/iccv/RockwellF021}\\
            SE3DS~\citep{DBLP:conf/aaai/KohABTWLYBA23} \\
          } \\
        \cmidrule(lr){2-4}
        & Matterport3D~\citep{DBLP:conf/3dim/ChangDFHNSSZZ17}
        & FID, IS, CLIP Score
        & \makecell{%
            SynSin~\citep{DBLP:conf/cvpr/WilesGS020}\\
            MVDiffusion~\citep{DBLP:conf/nips/TangZCWF23}\\
            PanFusion~\citep{DBLP:conf/cvpr/ZhangWGH0O024} \\
          } \\
    \midrule
      \multirow{1}{*}{\rotatebox{90}{\parbox{7 mm}{\centering Natural}}}
        & \makecell{%
             ACID~\citep{DBLP:conf/iccv/LiuM0SJK21} \\
             LHQ~\citep{DBLP:conf/iccv/SkorokhodovSE21} \\
           }
        & FID, KID, IS
        & \makecell{%
            InfiniteNature~\citep{DBLP:conf/iccv/LiuM0SJK21} \\
            InfiniteNature-Zero~\citep{DBLP:conf/eccv/LiWSK22} \\
            PersistentNature~\citep{DBLP:conf/cvpr/Chai0LIS23} \\
            DiffDreamer~\citep{DBLP:conf/iccv/CaiCPSOGW23} \\
          } \\
    \midrule
      \multirow{9}{*}{\rotatebox{90}{Urban}}
        & nuScenes~\citep{DBLP:conf/cvpr/CaesarBLVLXKPBB20}
        & FID, FVD
        & \makecell{%
            MagicDrive~\citep{DBLP:conf/iclr/0001CXHLY024} \\
            Vista~\citep{DBLP:conf/nips/GaoY0CQ0Z024} \\
            DiST-4D~\citep{DBLP:journals/corr/abs-2503-15208} \\
          } \\
        \cmidrule(lr){2-4}
        & KITTI-360~\citep{DBLP:journals/pami/LiaoXG23}
        & FID, KID
        & \makecell{
            CC3D~\citep{DBLP:conf/iccv/BahmaniPPYWGT23} \\
            UrbanGIRAFFE~\citep{DBLP:conf/iccv/YangYGX0L23} \\
            GaussianCity~\citep{DBLP:journals/corr/abs-2406-06526} \\
          } \\
        \cmidrule(lr){2-4}
        & GoogleEarth~\citep{DBLP:conf/cvpr/Xie0H024}
        & \makecell{%
            FID, KID, Depth Accuracy,\\
            Camera Pose Error
          }
        & \makecell{%
            CityDreamer~\citep{DBLP:conf/cvpr/Xie0H024} \\
            GaussianCity~\citep{DBLP:journals/corr/abs-2406-06526} \\
            CityDreamer4D~\citep{DBLP:journals/corr/abs-2501-08983} \\
          } \\
    \bottomrule
  \end{tabularx}

%% file: sections/applications.tex
\section{Tasks and Applications}
\label{sec:task-applications}

\rev{The rapid progress in 3D scene generation has enabled a wide range of applications.
In this section, we organize these applications into two categories: downstream tasks and application domains.
The former includes 3D scene editing, human--scene interaction, and embodied navigation, where generated scenes are directly used for creation, interaction, or agent-centric decision making.
The latter includes robotics and autonomous driving, where 3D scene generation provides realistic and diverse environments for simulation, training, and evaluation.}

\subsection{\rev{Downstream Tasks}}
\label{sec:downstream-tasks}

\subsubsection{3D Scene Editing}
\label{sec:scene-editing}

In 3D scene editing, 3D scene generation is used to modify a scene's appearance and structure, ranging from local object edits to full environment customization.
This task typically involves both texture editing, which alters surface appearance, and layout editing, which rearranges objects in a physically and semantically plausible manner.
\rev{These problems are mainly supported by procedural and neural 3D-based generation methods, which provide interactive simulation environments and enable motion-conditioned scene generation and human--scene contact modeling, respectively.}

Texturing and stylization aim to create aesthetic and stylish appearances based on user specifications. 
While recent advances achieve impressive results on scanned meshes~\citep{DBLP:conf/cvpr/Hollein0N22,DBLP:conf/mm/SongCXKTYY23,DBLP:conf/eccv/ChenHVSY24} or synthetic indoor datasets~\citep{DBLP:conf/cvpr/ChenL0TN24,DBLP:conf/cvpr/abs-2412-16778,DBLP:conf/cvpr/HwangK023}, they are constrained by incomplete geometry from reconstructions or extensive manual modeling.
To address these limitations, recent methods leverage 3D scene generation to synthesize complete and semantically consistent scenes, directly supporting texture generation tasks.
Methods such as Ctrl-Room~\citep{DBLP:conf/3dim/abs-2310-03602}, ControlRoom3D~\citep{DBLP:conf/cvpr/SchultTHWWMLWWH24}, RoomTex~\citep{DBLP:conf/eccv/WangLXWWDZX24}, and DreamSpace~\citep{DBLP:conf/vr/YangDMHLCM24} employ holistic generation techniques to create panoramic room textures, followed by detailed refinement.
Beyond direct generation, 3D scene generation also benefits the evaluation of texturing methods.
InstanceTex~\citep{DBLP:conf/siggrapha/YangGCCLC0H24} generates texture across both existing datasets and new scenes generated by EchoScene~\citep{DBLP:conf/eccv/ZhaiOCLDNTB24}, improving the diversity and robustness of benchmark evaluations.

3D scene layout editing focuses on arranging objects within a scene to produce semantically meaningful and physically plausible configurations.
Several methods, such as LEGO-Net~\citep{DBLP:conf/cvpr/WeiDPSP0G23}, CabiNet~\citep{DBLP:conf/icra/MuraliMEFF23}, and DeBaRA~\citep{DBLP:conf/nips/MaillardSDO24}, address the rearrangement of existing scenes. 
These approaches use object-level attributes, such as class labels, positions, and orientations, to produce more organized and regular arrangements.
Some methods enable more interactive and dynamic layout editing. For example, SceneExpander~\citep{DBLP:conf/mm/ZhangHYZLLZ24} and SceneDirector~\citep{DBLP:journals/tvcg/ZhangTLRFZ24} support real-time, intuitive edits (e.g., changing room shapes or moving objects) while maintaining spatial coherence.
Recent compositional generative NeRF approaches further extend layout control to implicit representations. DisCoScene~\citep{DBLP:conf/cvpr/XuCSPSSYSLZT23}, Neural Assets~\citep{DBLP:conf/nips/WuRKHGASAK24}, and Lift3D~\citep{DBLP:conf/cvpr/LiLWMC23} enable object-level editing via control signals such as spatial positions or latent features, allowing flexible scene manipulation.

\subsubsection{Human-Scene Interaction}
\label{sec:hsi}

In human--scene interaction, 3D scene generation is used to model how humans interact with and influence their environments.
A central goal is to synthesize believable interactions between virtual characters and surrounding scenes, including motion, contact, and behavior.
\rev{These goals are supported by procedural and neural 3D-based generation methods, which provide interactive simulation environments and motion-conditioned scene generation, respectively.}

Recent advances in HSI have made notable progress in generating realistic and physically plausible human motions within 3D environments~\citep{DBLP:conf/cvpr/HuangWLJLZLZ23,DBLP:conf/siggraph/HassanGWBFP23,DBLP:conf/cvpr/PanYDWHDKW25}, as well as creating scenes that align with specific motion sequences~\citep{DBLP:conf/siggrapha/YeWLPLX022,DBLP:conf/cvpr/YiHTHTB23,DBLP:conf/nips/AnVNHNVN23}.
%
%
To generate human motion conditioned on scene environments, some approaches~\citep{DBLP:conf/cvpr/0003RLYLD22,DBLP:conf/eccv/ZhaoWZBT22,DBLP:conf/cvpr/HuangWLJLZLZ23} directly learn from datasets containing scanned indoor scenes and captured human motion~\citep{DBLP:conf/iccv/HassanCTB19,DBLP:conf/nips/WangCLZLH22,DBLP:conf/eccv/MaYHGJPPGBKBFLLENN24}.
However, these datasets are often limited in scalability and restricted to static scenes, prohibiting the modeling of dynamic human-object interactions. 
Some works~\citep{DBLP:conf/siggraph/HassanGWBFP23,DBLP:conf/iccv/0004ZWB023,DBLP:conf/3dim/PanWHZWTW24,DBLP:conf/cvpr/PanYDWHDKW25,DBLP:journals/corr/abs-2411-19921} use simulated environments with reinforcement learning to produce physically plausible motion. 
However, their high setup cost often leads to simplified scenes, creating a sim-to-real gap when applied to more complex real-world environments.

Recent efforts like GenZI~\citep{DBLP:conf/cvpr/LiD24a} have initially addressed this issue by lifting the generated human in 2D images into 3D, enabling zero-shot generalization to novel scenes.
Although GenZI still depends on pre-designed synthetic scenes for evaluation, it highlights the potential of combining scene generation with motion generation to scale HSI data more effectively.
Integrating high-quality 3D scene generation is essential for advancing scalable and realistic HSI research, particularly by jointly considering human affordances, motion feasibility, and scene semantics.

\subsubsection{\rev{Embodied Navigation}}
\label{sec:navigation}

\rev{In embodied navigation, 3D scene generation is used to create visually rich, functionally plausible, and diverse environments for training and evaluation. 
Scalability and diversity are particularly important, as agents must generalize across a wide range of layouts, objects, and interactions~\citep{DBLP:conf/acl/GuSWTW22}. 
These requirements are best supported by procedural methods, which offer scalable, diverse, and controllable interactive environments.}

\rev{Existing simulation environments for embodied navigation are typically built from reconstructed real-world data~\citep{DBLP:conf/nips/RamakrishnanGWM21,DBLP:conf/nips/SzotCUWZTMMCMGV21} or manually designed scenes~\citep{DBLP:journals/corr/abs-1712-05474,DBLP:conf/corl/0002ZWGSMWLLSAH22}, both of which have limitations: reconstructions often suffer from incomplete geometry, limited diversity, and noisy annotations, while manual design is labor-intensive and hard to scale. 
In contrast, 3D scene generation offers a scalable alternative for creating diverse and physically plausible environments for embodied navigation.}

\rev{For indoor settings, ProcTHOR~\citep{DBLP:conf/nips/DeitkeVHWESHKKM22} uses procedural generation to create scenes with realistic layouts and physical constraints, supporting large-scale embodied navigation benchmarks.
Holodeck~\citep{DBLP:conf/cvpr/YangSWVHH0HKLCY24} leverages LLMs to automatically generate 3D environments that match user-specified prompts, enabling more flexible and customizable scene creation.
InfiniteWorld~\citep{DBLP:journals/corr/abs-2412-05789} further enriches generated environments by expanding assets with diverse textures and appearances.
PhyScene~\citep{DBLP:conf/cvpr/YangJZH24} incorporates physics and interactivity constraints into a conditional diffusion model to synthesize more plausible and interactive indoor scenes.
Architect~\citep{DBLP:conf/nips/WangQLCCWWXG24} employs iterative image-based inpainting to populate scenes with both large furniture and small objects, increasing layout complexity and visual diversity.}

\rev{Beyond indoor environments, procedural generation has also enabled city-scale embodied navigation and exploration.
MetaUrban~\citep{DBLP:conf/iclr/abs-2407-08725}, GRUtopia~\citep{DBLP:journals/corr/abs-2407-10943}, and URBAN-SIM~\citep{DBLP:conf/cvpr/abs-2505-00690} construct diverse large-scale urban environments for embodied agents.
EmbodiedCity~\citep{DBLP:journals/corr/abs-2410-09604} further provides a high-quality 3D environment based on a real city, supporting various agents, continuous decision making, and systematic benchmark tasks for embodied navigation and related embodied-agent settings.}

\input{tables/app-driven-paradigms}

\subsection{\rev{Application Domains}}
\label{sec:applications}

\subsubsection{Robotics}
\label{sec:robotics}

In robotics, 3D scene generation serves two main roles: constructing realistic simulated environments for learning, and modeling future visual outcomes for prediction and planning. 
The former remains dominant, as most large-scale pipelines still rely on teleoperation data and simulator-based or procedural environments for data collection, policy training, and evaluation. 
\rev{These systems are primarily supported by procedural and neural 3D-based methods, which enable controllable simulation and structured scene modeling, while recent video-based approaches are emerging for world modeling, prediction, and scalable data generation.}
Notably, when retrieved assets include watertight geometry, collision proxies, part-level articulation, and physical metadata, methods that predict layouts and retrieve such assets from curated libraries can produce scenes that are simulation-ready by construction.
This largely explains their dominance in robotics and embodied AI simulation pipelines, despite their diversity being bounded by the asset library.

Simulated environments~\citep{DBLP:journals/corr/abs-2511-04831} have become a central tool for developing robotic capabilities across various tasks, including complex manipulation and locomotion.
However, recent approaches in robot learning~\citep{DBLP:journals/ral/MeesHRB22, DBLP:conf/nips/LiuZGFLZS23, DBLP:conf/iccv/GongHZG0WAZTZJH23, DBLP:conf/corl/LiHGMPWFLSKL0F024, DBLP:journals/corr/abs-2406-02523} require tremendous human effort to construct these environments and the corresponding demonstrations, restricting the scalability of robot learning even in simulated worlds.
RoboGen~\citep{DBLP:conf/icml/WangXCWWFEHG24} and RoboVerse~\citep{DBLP:journals/corr/abs-2504-18904} automate task, scene, and supervision generation via a propose–generate–learn cycle, while Eurekaverse~\citep{DBLP:journals/corr/abs-2411-01775} scales skill acquisition by using LLMs to generate progressively harder terrains, forming an adaptive curriculum.

Beyond simulator construction, 3D scene generation is increasingly used for both data generation and world modeling in robotics.
While teleoperation and PCG-based simulators remain the dominant sources of robot training data, recent works such as RoboGen~\citep{DBLP:conf/icml/WangXCWWFEHG24}, RoboVerse~\citep{DBLP:journals/corr/abs-2504-18904}, and Eurekaverse~\citep{DBLP:journals/corr/abs-2411-01775} improve scalability by automatically generating tasks, scenes, supervision, and curricula.
Recent efforts~\citep{lightwheel2026leisaac} also enhance simulator realism by incorporating neural scene representations such as NeRFs and 3D Gaussian Splatting into robotic pipelines.

At the same time, robotics-focused video generation models~\citep{DBLP:conf/nips/DuY0DN0SA23,DBLP:conf/iclr/WuJCCXLLLK24,DBLP:journals/corr/abs-2410-06158,DBLP:conf/iclr/abs-2410-23277,DBLP:conf/icml/abs-2412-14803} have emerged as a promising direction for predicting future observations, planning manipulation behaviors, and generating additional training data.
This trend is consistent with recent generalist robot learning efforts such as GEN-0~\citep{generalist2025gen0}, which highlight data scaling as a central bottleneck.
Instead of directly synthesizing video frames, some methods~\citep{DBLP:conf/corl/ZeYWMGY0LW23,DBLP:conf/iros/DasguptaGTP24,DBLP:conf/eccv/LuZWLLT24} use NeRFs and dynamic 3D Gaussians to model spatial and semantic scene complexity for more accurate motion estimation and planning.

\subsubsection{Autonomous Driving}
\label{sec:autonomus-driving}

In autonomous driving, 3D scene generation serves two roles: constructing controllable simulation environments for training and evaluation, and modeling dynamic visual futures for prediction and data generation.
\rev{These applications are traditionally supported by procedural methods, while recent video-based approaches are increasingly used for world modeling, prediction, and scalable data generation.}

Autonomous driving has traditionally been dominated by simulator-based environments, with early systems such as CARLA~\citep{DBLP:conf/corl/DosovitskiyRCLK17}, AirSim~\citep{DBLP:conf/fsr/ShahDLK17}, and LGSVL/SVL~\citep{DBLP:conf/itsc/RongSTLLMBUGMAK20} providing scalable and controllable virtual worlds for training and evaluation.
Recently, 3D scene generation has increasingly been used as a world-modeling tool in autonomous driving, supporting future prediction, risk anticipation, and planning. 
Existing approaches can be broadly divided into two directions: some focus on future video prediction~\citep{DBLP:journals/corr/abs-2309-17080,DBLP:conf/cvpr/00010FLC024, DBLP:conf/eccv/WangZHCZL24, DBLP:conf/nips/GaoY0CQ0Z024, DBLP:conf/iclr/0001CXHLY024}, while others generate explicit 3D occupancy representations of the environment~\citep{DBLP:conf/eccv/ZhengCHZDL24, DBLP:conf/aaai/YangMMDCQFL25, DBLP:conf/cvpr/MaCHXLXG0024, DBLP:journals/corr/abs-2409-03272, DBLP:journals/corr/abs-2405-20337}.
\rev{Compared with pure future-frame prediction, these explicit 3D or 4D representations expose the spatial structure of evolving environments more directly, making them better suited for spatial reasoning in planning and simulation, such as reasoning about free space, object layout, and physical interactions~\citep{DBLP:conf/eccv/LuHYZZ24, DBLP:journals/corr/abs-2410-00337, DBLP:journals/corr/abs-2412-03934, DBLP:journals/corr/abs-2411-11252, DBLP:conf/cvpr/LiGLZDCZTZW0Z0Z25}.}

Autonomous driving requires large, diverse datasets, yet real-world collections like nuScenes~\citep{DBLP:conf/cvpr/CaesarBLVLXKPBB20}, KITTI~\citep{DBLP:conf/cvpr/GeigerLU12}, and Waymo~\citep{DBLP:conf/cvpr/SunKDCPTGZCCVHN20} are costly and miss critical corner cases.
Controllable video-based generation~\citep{DBLP:journals/corr/abs-2503-00045, DBLP:journals/corr/abs-2501-00601, DBLP:journals/corr/abs-2409-04003, DBLP:journals/corr/abs-2409-01595, DBLP:conf/cvpr/abs-2412-13188} addresses this by synthesizing diverse, controllable scenarios, especially rare safety-critical events.

\subsection{Paradigm Choice from an Application Perspective}
\label{sec:app-driven-paradigms}

In practice, paradigm choice is driven more by application requirements than by paradigm-level merits alone. Table~\ref{tab:app-driven-paradigms} provides this demand-side view by mapping representative application scenarios to their key requirements, the corresponding must-have dimensions defined in Section~\ref{sec:paradigm-tradeoffs}, and the best-aligned paradigms.

Different applications prioritize different subsets of the seven dimensions. 
Robotics and embodied interaction emphasize physical plausibility, controllability, and diversity (P, C, D); autonomous driving prioritizes controllability, diversity, realism, and view consistency (C, D, R, V) for scalable corner-case generation and trajectory prediction; XR and games prioritize efficiency, realism, and view consistency (E, R, V); while content creation focuses on controllability, efficiency, and diversity (C, E, D) for editable content and rapid iteration. 
Accordingly, procedural and retrieval-based neural 3D methods remain strong for structured, simulation-oriented tasks, while video-based world models are becoming increasingly attractive for prediction and scalable data generation.

Together with Figure~\ref{fig:representative-methods}, these mappings illustrate that the importance of each dimension is application dependent. For example, physical plausibility is essential for robotics but secondary for content creation, whereas efficiency is indispensable for XR but less critical for offline data generation.

%% file: tables/app-driven-paradigms.tex
\begin{table*}[t]
  \caption{\textbf{Paradigm selection across representative applications.} Representative applications are mapped to their key requirements, must-have dimensions, and best-suited generation paradigms.
  The listed paradigms indicate dominant trends rather than exclusive choices.
  Specifically, 
    \textcolor[rgb]{0.96078431, 0.20784314, 0.27058824}{\bf R}, 
    \textcolor[rgb]{1, 0.54117647, 0}{\bf D}, 
    \textcolor[rgb]{0.98039216, 0.84313725, 0.08235294}{\bf V}, 
    \textcolor[rgb]{0, 0.72941176, 0.44705882}{\bf S}, 
    \textcolor[rgb]{0, 0.76078431, 0.87058824}{\bf E}, 
    \textcolor[rgb]{0, 0.25882353, 0.55294118}{\bf C}, and 
    \textcolor[rgb]{0.37647059, 0.14509804, 0.47058824}{\bf P} denote 
    \textcolor[rgb]{0.96078431, 0.20784314, 0.27058824}{Realism}, 
    \textcolor[rgb]{1, 0.54117647, 0}{Diversity}, 
    \textcolor[rgb]{0.98039216, 0.84313725, 0.08235294}{View Consistency}, 
    \textcolor[rgb]{0, 0.72941176, 0.44705882}{Semantic Consistency}, 
    \textcolor[rgb]{0, 0.76078431, 0.87058824}{Efficiency}, 
    \textcolor[rgb]{0, 0.25882353, 0.55294118}{Controllability}, and 
    \textcolor[rgb]{0.37647059, 0.14509804, 0.47058824}{Physical Plausibility}, respectively.}
  \label{tab:app-driven-paradigms}
  \resizebox{\linewidth}{!}{
  \begin{tabular}{l p{5cm} c p{5cm}}
    \toprule
    \textbf{Application} &
    \textbf{Key Requirements} &
    \textbf{Must-have Dims.} &
    \textbf{Best-suited Paradigms} \\
    \midrule
    
    Robotics &
    Sim-ready object-level geometry; contact and articulation; scalable task diversity &
    \textcolor[rgb]{0.37647059,0.14509804,0.47058824}{\textbf{P}},
    \textcolor[rgb]{0,0.25882353,0.55294118}{\textbf{C}},
    \textcolor[rgb]{1,0.54117647,0}{\textbf{D}} &
    Procedural; Neural 3D (retrieval-based); Video-based methods for scalable data generation (emerging) \\
    
    Autonomous Driving &
    Data scale; rare corner cases; controllable trajectories; multi-view output &
    \textcolor[rgb]{0,0.25882353,0.55294118}{\textbf{C}},
    \textcolor[rgb]{1,0.54117647,0}{\textbf{D}},
    \textcolor[rgb]{0.96078431,0.20784314,0.27058824}{\textbf{R}},
    \textcolor[rgb]{0.98039216,0.84313725,0.08235294}{\textbf{V}} &
    Video-based (one-stage); Neural 3D (occupancy) \\
    
    XR / Games &
    Real-time photorealistic rendering; low latency; persistent scenes &
    \textcolor[rgb]{0,0.76078431,0.87058824}{\textbf{E}},
    \textcolor[rgb]{0.96078431,0.20784314,0.27058824}{\textbf{R}},
    \textcolor[rgb]{0.98039216,0.84313725,0.08235294}{\textbf{V}} &
    Neural 3D (3DGS); Image-based + 3DGS; Procedural \\
    
    Content Creation &
    Controllability; editability; iteration speed &
    \textcolor[rgb]{0,0.25882353,0.55294118}{\textbf{C}},
    \textcolor[rgb]{0,0.76078431,0.87058824}{\textbf{E}},
    \textcolor[rgb]{1,0.54117647,0}{\textbf{D}} &
    Image-based; Neural 3D (compositional) \\
    
    HSI / Embodied Nav. &
    Interactivity; physical plausibility; layout diversity &
    \textcolor[rgb]{0.37647059,0.14509804,0.47058824}{\textbf{P}},
    \textcolor[rgb]{1,0.54117647,0}{\textbf{D}},
    \textcolor[rgb]{0,0.25882353,0.55294118}{\textbf{C}} &
    Procedural; Neural 3D \\
    \bottomrule
  \end{tabular}
  }
\end{table*}

%% file: sections/future-work.tex
\section{Challenges and Future Directions}
\label{sec:challenges-and-future}

\subsection{Challenges}
\label{sec:challenges}

\rev{Building upon the failure patterns identified in Section~\ref{sec:critical-analysis}, we distill several core challenges that underlie current limitations in 3D scene generation.}

\noindent
\rev{\textbf{Consistency.}
A central challenge in 3D scene generation is maintaining consistency across viewpoints and time. 
Image-based methods~\citep{DBLP:conf/cvpr/YuDHSRFCSS0H24, DBLP:journals/corr/abs-2406-09394} often suffer from geometric drift and semantic discontinuity due to their reliance on local view-conditioned generation without sufficient global context. 
Video-based methods~\citep{DBLP:conf/nips/GaoY0CQ0Z024, DBLP:journals/corr/abs-2411-13807} improve short-term coherence through temporal modeling, yet temporal inconsistency and unstable scene evolution remain common, especially over long sequences.
Even neural 3D-based methods~\citep{DBLP:conf/cvpr/Xie0H024, DBLP:journals/corr/abs-2406-06526} with explicit representations such as NeRF or 3D Gaussians may still exhibit visual inconsistency, as the generated scene representations are not always coherent across viewpoints. 
Overall, current methods still struggle to preserve consistent scene structure, semantics, and appearance throughout generation, limiting their reliability for stable multi-view and long-horizon synthesis. 
Potential directions include stronger global scene constraints, persistent scene states, and explicit consistency objectives across viewpoints and time.}

\noindent 
\rev{\textbf{Realism--Structure Trade-off.}
Another major challenge lies in balancing visual realism with structural reliability, where structure is closely related to simulatability in physically grounded applications. 
Image- and video-based methods~\citep{DBLP:conf/nips/TangZCWF23, DBLP:journals/corr/abs-2411-00769} often achieve realistic and diverse appearances by leveraging large-scale visual data, but their outputs typically lack explicit object boundaries, collision structures, and physical parameters required for simulation. 
In contrast, procedural and retrieval-based methods~\citep{DBLP:conf/cvpr/RaistrickLMMWZK23, DBLP:conf/cvpr/RaistrickMKY0HW24} provide stronger structural control and simulation-ready representations, but their appearance diversity is constrained by predefined rules and asset libraries. 
Neural 3D-based methods with implicit representations lie between these two extremes, offering geometric consistency for rendering while requiring additional processing for physical interaction. 
This trade-off indicates that current approaches still struggle to jointly achieve realistic appearance, reliable structure, and physical simulatability, particularly in complex multi-object scenes.}

\noindent 
\textbf{Physical Grounding.}
Physical grounding remains a major challenge in 3D scene generation, especially for embodied AI, where generated scenes are expected to support interaction, planning, and decision making. 
\rev{Physical grounding differs from representation simulatability: the former concerns whether generated scenes obey physical relations, while the latter concerns whether the representation is directly usable by a physics engine.}
Even neural 3D-based methods~\citep{DBLP:conf/cvpr/HuangWLJLZLZ23} with explicit geometry may still produce interpenetration or incorrect contact relationships, showing that geometric validity does not guarantee physical correctness. 
This limitation is more pronounced in image- and video-based methods ~\citep{DBLP:journals/corr/abs-2603-16669}, where the lack of explicit scene structure makes it difficult to preserve basic physical relations. 
\rev{While sim-ready representations enable simulator-based validation and enforcement of physical constraints, outputs that are not directly simulation-ready generally require additional conversion and post-processing for physics-based validation; without such processing, such methods rely more heavily on learned physical priors, which can lead to implausible interactions.}
Therefore, current methods still require stronger contact-aware and collision-aware constraints, as well as more explicit modeling of object interactions.

\noindent 
\rev{\textbf{Evaluation.}
Progress in 3D scene generation is further hindered by the lack of unified evaluation protocols. 
As reflected in our comparative analysis, different paradigms are often assessed on different datasets and with different metrics, making direct comparison difficult. 
Although recent benchmarks~\citep{DBLP:conf/cvpr/HuangHYZS0Z0JCW24, DBLP:journals/corr/abs-2411-13503} have begun to evaluate properties related to consistency and temporal quality, these assessments are still largely based on proxy dimensions and are not yet fully tailored to scene-level 3D generation. 
Consequently, important aspects such as structural correctness, long-horizon coherence, and physical validity are still not comprehensively captured. 
Moving forward, more comprehensive evaluation frameworks are needed to jointly assess realism, consistency, and physical grounding across diverse generation settings.}

\subsection{Future Directions}
\label{sec:future-directions}

Building upon the above challenges, we outline future directions that may improve realism, consistency, and physical grounding in 3D scene generation.

\noindent \textbf{Better Fidelity.}
High-fidelity 3D scene generation demands coherence in geometry, texture, lighting, and multi-view consistency. 
While current methods often trade off between geometric accuracy and visual richness, future models should focus on bridging this gap by jointly reason about structure and appearance.
Key goals include improved material and lighting modeling, consistent object identity across views, and capturing subtle cues like shadows and occlusions. 
Achieving scene-level fidelity also means aligning local details with global spatial and semantic coherence, enabling more realistic and useful 3D environments.

\noindent \textbf{Physics-aware Generation.}
Despite impressive visual progress, current methods often overlook the physical plausibility of generated scenes. 
To ensure that object placements and articulations conform to physical laws, future work should incorporate physics priors, constraints, or simulations into the generation process. 
Emerging approaches that integrate physics-based feedback, such as differentiable simulators~\citep{DBLP:conf/iclr/HuALSCRD20}, offer a promising path toward jointly optimizing structure, semantics, and physical behavior. 
These capabilities are especially important for embodied AI and robotics, where agents depend on physically consistent environments for effective planning and control.

\noindent \textbf{Interactive Scene Generation.}
Recent advances in 4D scene generation~\citep{DBLP:journals/corr/abs-2501-08983} have enabled dynamic environments with movable objects, but these scenes remain largely non-interactive and thus mostly support passive rather than reactive experiences.
\rev{An important future direction is interactive and immersive scene generation, where objects respond meaningfully to physical interactions, user commands, and contextual changes. 
This requires generative models to move beyond geometry and motion, incorporating reasoning about object affordances, causal relationships, and multi-agent dynamics. 
Such capabilities are particularly important for immersive applications such as XR, virtual simulation, and interactive content creation, where scenes must not only be behaviorally responsive, but also support low-latency updates and affordable computation. 
Meeting these requirements calls for generation frameworks that jointly consider interactivity, real-time performance, and computational efficiency.}

\noindent \textbf{Unified Perception-Generation.}
A promising frontier lies in unifying perception and generation under a shared model. 
Tasks such as segmentation, reconstruction, and scene synthesis benefit from common spatial and semantic priors. 
Moreover, generation tasks inherently require an understanding of the input modalities.
A unified architecture could leverage bidirectional capabilities: enhancing generative performance via perceptual grounding and improving scene understanding through generative modeling. 
Such models could serve as general-purpose backbones for embodied agents, supporting joint reasoning across vision, language, and 3D spatial representations.

%% file: sections/acknowledgement.tex
\section*{Funding Declaration}

This study is supported by the Ministry of Education, Singapore, under its MOE AcRF Tier 2 (MOET2EP20221-0012, MOE-T2EP20223-0002), and under the RIE2020 Industry Alignment Fund – Industry Collaboration Projects (IAF-ICP) Funding Initiative, as well as cash and in-kind contribution from the industry partner(s).











%% file: references.bib
@String(PAMI  = {IEEE TPAMI})

@String(CVPR  = {CVPR})

@String(ICCV  = {ICCV})

@String(ECCV  = {ECCV})

@String(WACV = {WACV})

@String(ICML  =	{ICML})

@String(NIPS  = {NeurIPS})

@String(SIGGRAPH = {SIGGRAPH})

@String(SIGGRAPHA = {SIGGRAPH Asia})

@String(TOG   = {ACM TOG})

@String(TVCG  = {IEEE TVCG})

@String(TMM   =	{IEEE TMM})

@String(ACMMM = {ACM MM})

@String(ICIP  = {ICIP})

@String(ICLR  = {ICLR})

@String(IJCAI = {IJCAI})

@String(AAAI = {AAAI})

@String(ThreeDV = {3DV})

@String(CGF = {Computer Graphics Forum})

@String(EMNLP = {EMNLP})

@String(RSS = {RSS})

@String(CORL = {CORL})

@article{DBLP:journals/pami/HuangWWRJ24,
  author       = {Zhangjin Huang and
                  Yuxin Wen and
                  Zihao Wang and
                  Jinjuan Ren and
                  Kui Jia},
  title        = {Surface Reconstruction From Point Clouds: {A} Survey and a Benchmark},
  journal      = PAMI,
  volume       = {46},
  number       = {12},
  pages        = {9727--9748},
  year         = {2024}
}

@article{DBLP:journals/pami/HanLB21,
  author       = {Xian{-}Feng Han and
                  Hamid Laga and
                  Mohammed Bennamoun},
  title        = {Image-Based 3{D} Object Reconstruction: State-of-the-Art and Trends
                  in the Deep Learning Era},
  journal      = PAMI,
  volume       = {43},
  number       = {5},
  pages        = {1578--1604},
  year         = {2021}
}

@article{DBLP:journals/csur/XingFCDHXWJ25,
  author       = {Zhen Xing and
                  Qijun Feng and
                  Haoran Chen and
                  Qi Dai and
                  Han Hu and
                  Hang Xu and
                  Zuxuan Wu and
                  Yu{-}Gang Jiang},
  title        = {A Survey on Video Diffusion Models},
  journal      = {{ACM} Computing Surveys},
  volume       = {57},
  number       = {2},
  pages        = {41:1--41:42},
  year         = {2025}
}

@article{DBLP:journals/corr/abs-2402-17177,
  author       = {Yixin Liu and
                  Kai Zhang and
                  Yuan Li and
                  Zhiling Yan and
                  Chujie Gao and
                  Ruoxi Chen and
                  Zhengqing Yuan and
                  Yue Huang and
                  Hanchi Sun and
                  Jianfeng Gao and
                  Lifang He and
                  Lichao Sun},
  title        = {Sora: {A} Review on Background, Technology, Limitations, and Opportunities of Large Vision Models},
  journal      = {arXiv},
  volume       = {2402.17177},
  year         = {2024}
}

@article{DBLP:journals/corr/abs-2210-15663,
  author       = {Zifan Shi and
                  Sida Peng and
                  Yinghao Xu and
                  Yiyi Liao and
                  Yujun Shen},
  title        = {Deep Generative Models on 3{D} Representations: {A} Survey},
  journal      = {arXiv},
  volume       = {2210.15663},
  year         = {2022}
}

@article{DBLP:journals/corr/abs-2401-17807,
  author       = {Xiaoyu Li and
                  Qi Zhang and
                  Di Kang and
                  Weihao Cheng and
                  Yiming Gao and
                  Jingbo Zhang and
                  Zhihao Liang and
                  Jing Liao and
                  Yan{-}Pei Cao and
                  Ying Shan},
  title        = {Advances in 3{D} Generation: {A} Survey},
  journal      = {arXiv},
  volume       = {2401.17807},
  year         = {2024}
}

@article{DBLP:journals/corr/abs-2410-04738,
  author       = {Zhen Wang and
                  Dongyuan Li and
                  Renhe Jiang},
  title        = {Diffusion Models in 3{D} Vision: {A} Survey},
  journal      = {arXiv},
  volume       = {2410.04738},
  year         = {2024}
}

@article{DBLP:journals/corr/abs-2402-01166,
  author       = {Jian Liu and
                  Xiaoshui Huang and
                  Tianyu Huang and
                  Lu Chen and
                  Yuenan Hou and
                  Shixiang Tang and
                  Ziwei Liu and
                  Wanli Ouyang and
                  Wangmeng Zuo and
                  Junjun Jiang and
                  Xianming Liu},
  title        = {A Comprehensive Survey on 3{D} Content Generation},
  journal      = {arXiv},
  volume       = {2402.01166},
  year         = {2024}
}

@article{DBLP:journals/corr/abs-2305-06131,
  author       = {Chenghao Li and
                  Chaoning Zhang and
                  Atish Waghwase and
                  Lik{-}Hang Lee and
                  Fran{\c{c}}ois Rameau and
                  Yang Yang and
                  Sung{-}Ho Bae and
                  Choong Seon Hong},
  title        = {Generative {AI} meets 3{D}: {A} Survey on Text-to-3{D} in {AIGC} Era},
  journal      = {arXiv},
  volume       = {2305.06131},
  year         = {2023}
}

@inproceedings{DBLP:conf/nips/HoJA20,
  author       = {Jonathan Ho and
                  Ajay Jain and
                  Pieter Abbeel},
  title        = {Denoising Diffusion Probabilistic Models},
  booktitle    = NIPS,
  year         = {2020}
}

@article{DBLP:journals/cgf/PatilPLFSZ24,
  author       = {Akshay Gadi Patil and
                  Supriya Gadi Patil and
                  Manyi Li and
                  Matthew Fisher and
                  Manolis Savva and
                  Hao Zhang},
  title        = {Advances in Data-Driven Analysis and Synthesis of 3{D} Indoor Scenes},
  journal      = CGF,
  volume       = {43},
  number       = {1},
  year         = {2024}
}

@article{DBLP:journals/jcst/ZhangZLH19,
  author       = {Song{-}Hai Zhang and
                  Shao{-}Kui Zhang and
                  Yuan Liang and
                  Peter Hall},
  title        = {A Survey of 3{D} Indoor Scene Synthesis},
  journal      = {Journal of Computer Science and Technology},
  volume       = {34},
  number       = {3},
  pages        = {594--608},
  year         = {2019}
}

@inproceedings{DBLP:conf/pais/GhorabL22,
  author       = {Mostafa Anouar Ghorab and
                  Abdelaziz Lakhfif},
  title        = {Text to 3{D}, 2{D} scene generation systems, Frameworks and approaches:
                  a survey},
  booktitle    = {Pattern Analysis and Intelligent Systems},
  year         = {2022}
}

@article{DBLP:journals/cgf/SmelikTBB14,
  author       = {Ruben Micha{\"{e}}l Smelik and
                  Tim Tutenel and
                  Rafael Bidarra and
                  Bedrich Benes},
  title        = {A Survey on Procedural Modelling for Virtual Worlds},
  journal      = {Computer Graphics Forum},
  volume       = {33},
  number       = {6},
  pages        = {31--50},
  year         = {2014}
}

@article{Ayyildiz2024ASO,
  title        = {A Survey of Learning Techniques for Virtual Scene Generation},
  author       = {Dilara Vefa Ayyildiz and 
                  Ala’ J. Alnaser and 
                  Shahram Taj and 
                  Mahta Zakaria and 
                  Luis Gabriel Jaimes},
  journal      = {SAE International Journal of Connected and Automated Vehicles},
  year         = {2024}
}

@article{DBLP:journals/csur/DingZSZZFYSLSXL26,
  author       = {Jingtao Ding and
                  Yunke Zhang and
                  Yu Shang and
                  Yuheng Zhang and
                  Zefang Zong and
                  Jie Feng and
                  Yuan Yuan and
                  Hongyuan Su and
                  Nian Li and
                  Nicholas Sukiennik and
                  Fengli Xu and
                  Yong Li},
  title        = {Understanding World or Predicting Future? {A} Comprehensive Survey of World Models},
  journal      = {{ACM} Computing Surveys},
  volume       = {58},
  number       = {3},
  pages        = {57:1--57:38},
  year         = {2026},
}

@article{DBLP:journals/corr/abs-2403-16993,
  author       = {Dejia Xu and
                  Hanwen Liang and
                  Neel P. Bhatt and
                  Hezhen Hu and
                  Hanxue Liang and
                  Konstantinos N. Plataniotis and
                  Zhangyang Wang},
  title        = {{Comp4D}: LLM-Guided Compositional 4{D} Scene Generation},
  journal      = {arXiv},
  volume       = {2403.16993},
  year         = {2024}
}

@inproceedings{DBLP:conf/icml/SingerSPAMKGVP023,
  author       = {Uriel Singer and
                  Shelly Sheynin and
                  Adam Polyak and
                  Oron Ashual and
                  Iurii Makarov and
                  Filippos Kokkinos and
                  Naman Goyal and
                  Andrea Vedaldi and
                  Devi Parikh and
                  Justin Johnson and
                  Yaniv Taigman},
  title        = {Text-To-4{D} Dynamic Scene Generation},
  booktitle    = ICML,
  year         = {2023}
}

@inproceedings{DBLP:conf/cvpr/ZhengLNLHM24,
  author       = {Yufeng Zheng and
                  Xueting Li and
                  Koki Nagano and
                  Sifei Liu and
                  Otmar Hilliges and
                  Shalini De Mello},
  title        = {A Unified Approach for Text-and Image-Guided 4{D} Scene Generation},
  booktitle    = CVPR,
  year         = {2024}
}

@inproceedings{DBLP:conf/eccv/MildenhallSTBRN20,
  author       = {Ben Mildenhall and
                  Pratul P. Srinivasan and
                  Matthew Tancik and
                  Jonathan T. Barron and
                  Ravi Ramamoorthi and
                  Ren Ng},
  title        = {{NeRF}: Representing Scenes as Neural Radiance Fields for View Synthesis},
  booktitle    = ECCV,
  year         = {2020}
}

@article{DBLP:journals/corr/abs-1810-00597,
  author       = {Danilo Jimenez Rezende and
                  Fabio Viola},
  title        = {Taming {VAEs}},
  journal      = {arXiv},
  volume       = {1810.00597},
  year         = {2018}
}

@article{DBLP:journals/tog/KerblKLD23,
  author       = {Bernhard Kerbl and
                  Georgios Kopanas and
                  Thomas Leimk{\"{u}}hler and
                  George Drettakis},
  title        = {3{D} {G}aussian Splatting for Real-Time Radiance Field Rendering},
  journal      = TOG,
  volume       = {42},
  number       = {4},
  pages        = {139:1--139:14},
  year         = {2023}
}

@inproceedings{DBLP:journals/corr/GoodfellowPMXWOCB14,
  author       = {Ian J. Goodfellow and
                  Jean Pouget{-}Abadie and
                  Mehdi Mirza and
                  Bing Xu and
                  David Warde{-}Farley and
                  Sherjil Ozair and
                  Aaron C. Courville and
                  Yoshua Bengio},
  title        = {Generative Adversarial Networks},
  booktitle    = {NIPS},
  year         = {2014}
}

@article{DBLP:journals/corr/abs-2311-15127,
  author       = {Andreas Blattmann and
                  Tim Dockhorn and
                  Sumith Kulal and
                  Daniel Mendelevitch and
                  Maciej Kilian and
                  Dominik Lorenz and
                  Yam Levi and
                  Zion English and
                  Vikram Voleti and
                  Adam Letts and
                  Varun Jampani and
                  Robin Rombach},
  title        = {Stable Video Diffusion: Scaling Latent Video Diffusion Models to Large
                  Datasets},
  journal      = {arXiv},
  volume       = {2311.15127},
  year         = {2023}
}

@article{DBLP:journals/tvcg/Max95a,
  author       = {Nelson L. Max},
  title        = {Optical Models for Direct Volume Rendering},
  journal      = TVCG,
  volume       = {1},
  number       = {2},
  pages        = {99--108},
  year         = {1995}
}

@inproceedings{DBLP:conf/iclr/0001CXHLY024,
  author       = {Ruiyuan Gao and
                  Kai Chen and
                  Enze Xie and
                  Lanqing Hong and
                  Zhenguo Li and
                  Dit{-}Yan Yeung and
                  Qiang Xu},
  title        = {Magic{D}rive: Street View Generation with Diverse 3{D} Geometry Control},
  booktitle    = ICLR,
  year         = {2024}
}

@inproceedings{DBLP:conf/siggraph/Deng0LGSW24,
  author       = {Boyang Deng and
                  Richard Tucker and
                  Zhengqi Li and
                  Leonidas J. Guibas and
                  Noah Snavely and
                  Gordon Wetzstein},
  title        = {Streetscapes: Large-scale Consistent Street View Generation Using
                  Autoregressive Video Diffusion},
  booktitle    = SIGGRAPH,
  year         = {2024}
}

@inproceedings{DBLP:conf/cvpr/SchonbergerF16,
  author       = {Johannes L. Sch{\"{o}}nberger and
                  Jan{-}Michael Frahm},
  title        = {Structure-from-Motion Revisited},
  booktitle    = CVPR,
  year         = {2016}
}

@inproceedings{DBLP:conf/siggraph/CurlessL96,
  author       = {Brian Curless and
                  Marc Levoy},
  title        = {A Volumetric Method for Building Complex Models from Range Images},
  booktitle    = SIGGRAPH,
  year         = {1996}
}

@inproceedings{DBLP:conf/cvpr/ParkFSNL19,
  author       = {Jeong Joon Park and
                  Peter R. Florence and
                  Julian Straub and
                  Richard A. Newcombe and
                  Steven Lovegrove},
  title        = {{DeepSDF}: Learning Continuous Signed Distance Functions for Shape Representation},
  booktitle    = CVPR,
  year         = {2019}
}

@inproceedings{DBLP:conf/iclr/abs-2406-13527,
  author       = {Renjie Li and
                  Panwang Pan and
                  Bangbang Yang and
                  Dejia Xu and
                  Shijie Zhou and
                  Xuanyang Zhang and
                  Zeming Li and
                  Achuta Kadambi and
                  Zhangyang Wang and
                  Zhiwen Fan},
  title        = {{4K4DGen}: Panoramic 4{D} Generation at 4{K} Resolution},
  booktitle    = ICLR, 
  year         = {2025}
}

@inproceedings{DBLP:journals/corr/abs-2411-00769,
  author       = {Haoxuan Che and
                  Xuanhua He and
                  Quande Liu and
                  Cheng Jin and
                  Hao Chen},
  title        = {Game{G}en-{X}: Interactive Open-world Game Video Generation},
  booktitle    = ICLR, 
  year         = {2025}
}

@inproceedings{DBLP:conf/cvpr/WangLMCZ24,
  author       = {Qian Wang and
                  Weiqi Li and
                  Chong Mou and
                  Xinhua Cheng and
                  Jian Zhang},
  title        = {360{DVD}: Controllable Panorama Video Generation with 360-Degree Video
                  Diffusion Model},
  booktitle    = CVPR, 
  year         = {2024}
}

@inproceedings{DBLP:conf/cvpr/Xie0H024,
  author       = {Haozhe Xie and
                  Zhaoxi Chen and
                  Fangzhou Hong and
                  Ziwei Liu},
  title        = {{CityDreamer}: Compositional Generative Model of Unbounded 3{D} Cities},
  booktitle    = CVPR,
  year         = {2024}
}

@inproceedings{DBLP:conf/iccv/LiuM0SJK21,
  author       = {Andrew Liu and
                  Ameesh Makadia and
                  Richard Tucker and
                  Noah Snavely and
                  Varun Jampani and
                  Angjoo Kanazawa},
  title        = {Infinite {N}ature: Perpetual View Generation of Natural Scenes from
                  a Single Image},
  booktitle    = ICCV,
  year         = {2021}
}

@inproceedings{DBLP:conf/eccv/LiWSK22,
  author       = {Zhengqi Li and
                  Qianqian Wang and
                  Noah Snavely and
                  Angjoo Kanazawa},
  title        = {{InfiniteNature-Zero}: Learning Perpetual View Generation of Natural
                  Scenes from Single Images},
  booktitle    = ECCV,
  year         = {2022}
}

@article{DBLP:journals/tog/ChenWL22,
  author       = {Zhaoxi Chen and
                  Guangcong Wang and
                  Ziwei Liu},
  title        = {{Text2Light}: Zero-Shot Text-Driven {HDR} Panorama Generation},
  journal      = TOG,
  volume       = {41},
  number       = {6},
  pages        = {195:1--195:16},
  year         = {2022}
}

@inproceedings{DBLP:conf/nips/TangZCWF23,
  author       = {Shitao Tang and
                  Fuyang Zhang and
                  Jiacheng Chen and
                  Peng Wang and
                  Yasutaka Furukawa},
  title        = {{MVDiffusion}: Enabling Holistic Multi-view Image Generation with Correspondence-Aware Diffusion},
  booktitle    = NIPS,
  year         = {2023}
}

@inproceedings{DBLP:journals/corr/KingmaW13,
  author       = {Diederik P. Kingma and
                  Max Welling},
  title        = {Auto-Encoding Variational Bayes},
  booktitle    = ICLR,
  year         = {2014}
}

@inproceedings{DBLP:conf/nips/GulrajaniAADC17,
  author       = {Ishaan Gulrajani and
                  Faruk Ahmed and
                  Mart{\'{\i}}n Arjovsky and
                  Vincent Dumoulin and
                  Aaron C. Courville},
  title        = {Improved Training of Wasserstein GANs},
  booktitle    = {NIPS},
  year         = {2017}
}

@inproceedings{DBLP:conf/nips/LucasTG019,
  author       = {James Lucas and
                  George Tucker and
                  Roger B. Grosse and
                  Mohammad Norouzi},
  title        = {{Don't Blame the ELBO}! {A} Linear {VAE} Perspective on Posterior Collapse},
  booktitle    = NIPS, 
  year         = {2019}
}

@article{DBLP:journals/tvcg/ZhangTLRFZ24,
  author       = {Shao{-}Kui Zhang and
                  Hou Tam and
                  Yike Li and
                  Ke{-}Xin Ren and
                  Hongbo Fu and
                  Song{-}Hai Zhang},
  title        = {Scene{D}irector: Interactive Scene Synthesis by Simultaneously Editing
                  Multiple Objects in Real-Time},
  journal      = TVCG,
  volume       = {30},
  number       = {8},
  pages        = {4558--4569},
  year         = {2024}
}

@inproceedings{DBLP:conf/nips/DhariwalN21,
  author       = {Prafulla Dhariwal and
                  Alexander Quinn Nichol},
  title        = {Diffusion Models Beat GANs on Image Synthesis},
  booktitle    = NIPS,
  year         = {2021}
}

@article{DBLP:journals/cgf/WuFLW18,
  author       = {Wenming Wu and
                  Lubin Fan and
                  Ligang Liu and
                  Peter Wonka},
  title        = {MIQP-based Layout Design for Building Interiors},
  journal      = CGF,
  year         = {2018}
}

@article{DBLP:journals/tvcg/JiangYZW20,
  author       = {Haiyong Jiang and
                  Dong{-}Ming Yan and
                  Xiaopeng Zhang and
                  Peter Wonka},
  title        = {Selection Expressions for Procedural Modeling},
  journal      = TVCG,
  volume       = {26},
  number       = {4},
  pages        = {1775--1788},
  year         = {2020}
}

@article{DBLP:journals/corr/abs-2407-17572,
  author       = {Shougao Zhang and
                  Mengqi Zhou and
                  Yuxi Wang and
                  Chuanchen Luo and
                  Rongyu Wang and
                  Yiwei Li and
                  Xucheng Yin and
                  Zhaoxiang Zhang and
                  Junran Peng},
  title        = {{CityX}: Controllable Procedural Content Generation for Unbounded 3{D}
                  Cities},
  journal      = {arXiv},
  volume       = {2407.17572},
  year         = {2024}
}

@article{DBLP:journals/tog/YangWVW13,
  author       = {Yong{-}Liang Yang and
                  Jun Wang and
                  Etienne Vouga and
                  Peter Wonka},
  title        = {Urban pattern: layout design by hierarchical domain splitting},
  journal      = TOG,
  volume       = {32},
  number       = {6},
  pages        = {181:1--181:12},
  year         = {2013}
}

@article{DBLP:journals/tog/TaltonLLDMK11,
  author       = {Jerry O. Talton and
                  Yu Lou and
                  Steve Lesser and
                  Jared Duke and
                  Radom{\'{\i}}r Mech and
                  Vladlen Koltun},
  title        = {Metropolis procedural modeling},
  journal      = TOG,
  volume       = {30},
  number       = {2},
  pages        = {11:1--11:14},
  year         = {2011}
}

@article{DBLP:journals/tog/YuYTTCO11,
  author       = {Lap{-}Fai Yu and
                  Sai Kit Yeung and
                  Chi{-}Keung Tang and
                  Demetri Terzopoulos and
                  Tony F. Chan and
                  Stanley J. Osher},
  title        = {Make it home: automatic optimization of furniture arrangement},
  journal      = TOG,
  volume       = {30},
  number       = {4},
  pages        = {86},
  year         = {2011}
}

@inproceedings{DBLP:conf/cvpr/RaistrickMKY0HW24,
  author       = {Alexander Raistrick and
                  Lingjie Mei and
                  Karhan Kayan and
                  David Yan and
                  Yiming Zuo and
                  Beining Han and
                  Hongyu Wen and
                  Meenal Parakh and
                  Stamatis Alexandropoulos and
                  Lahav Lipson and
                  Zeyu Ma and
                  Jia Deng},
  title        = {Infinigen {I}ndoors: Photorealistic Indoor Scenes using Procedural Generation},
  booktitle    = CVPR, 
  year         = {2024}
}

@inproceedings{DBLP:conf/aaai/abs-2403-15698,
  author       = {Mengqi Zhou and
                  Jun Hou and
                  Chuanchen Luo and
                  Yuxi Wang and
                  Zhaoxiang Zhang and
                  Junran Peng},
  title        = {{SceneX:} Procedural Controllable Large-scale Scene Generation via Large-language Models},
  booktitle    = AAAI,
  year         = {2025}
}

@article{DBLP:journals/corr/abs-2501-11260,
  title        = {A Survey of World Models for Autonomous Driving},
  author       = {Feng, Tuo and 
                  Wang, Wenguan and
                  Yang, Yi},
  journal      = {arXiv},
  volume       = {2501.11260},
  year         = {2025}
}

@inproceedings{DBLP:conf/siggraph/KajiyaH84,
  author       = {James T. Kajiya and
                  Brian Von Herzen},
  title        = {Ray tracing volume densities},
  booktitle    = SIGGRAPH,
  year         = {1984}
}

@article{DBLP:journals/corr/abs-2407-10943,
  author       = {Hanqing Wang and
                  Jiahe Chen and
                  Wensi Huang and
                  Qingwei Ben and
                  Tai Wang and
                  Boyu Mi and
                  Tao Huang and
                  Siheng Zhao and
                  Yilun Chen and
                  Sizhe Yang and
                  Peizhou Cao and
                  Wenye Yu and
                  Zichao Ye and
                  Jialun Li and
                  Junfeng Long and
                  Zirui Wang and
                  Huiling Wang and
                  Ying Zhao and
                  Zhongying Tu and
                  Yu Qiao and
                  Dahua Lin and
                  Jiangmiao Pang},
  title        = {{GRUtopia}: Dream General Robots in a City at Scale},
  journal      = {arXiv},
  volume       = {2407.10943},
  year         = {2024}
}

@article{DBLP:journals/corr/abs-2403-09227,
  author       = {Chengshu Li and
                  Ruohan Zhang and
                  Josiah Wong and
                  Cem Gokmen and
                  Sanjana Srivastava and
                  Roberto Mart{\'{\i}}n{-}Mart{\'{\i}}n and
                  Chen Wang and
                  Gabrael Levine and
                  Wensi Ai and
                  Benjamin Jose Martinez and
                  Hang Yin and
                  Michael Lingelbach and
                  Minjune Hwang and
                  Ayano Hiranaka and
                  Sujay Garlanka and
                  Arman Aydin and
                  Sharon Lee and
                  Jiankai Sun and
                  Mona Anvari and
                  Manasi Sharma and
                  Dhruva Bansal and
                  Samuel Hunter and
                  Kyu{-}Young Kim and
                  Alan Lou and
                  Caleb R. Matthews and
                  Ivan Villa{-}Renteria and
                  Jerry Huayang Tang and
                  Claire Tang and
                  Fei Xia and
                  Yunzhu Li and
                  Silvio Savarese and
                  Hyowon Gweon and
                  C. Karen Liu and
                  Jiajun Wu and
                  Li Fei{-}Fei},
  title        = {{BEHAVIOR-1K:} {A} Human-Centered, Embodied {AI} Benchmark with 1,
                  000 Everyday Activities and Realistic Simulation},
  journal      = {arXiv},
  volume       = {2403.09227},
  year         = {2024}
}

@inproceedings{DBLP:conf/nips/DeitkeVHWESHKKM22,
  author       = {Matt Deitke and
                  Eli VanderBilt and
                  Alvaro Herrasti and
                  Luca Weihs and
                  Kiana Ehsani and
                  Jordi Salvador and
                  Winson Han and
                  Eric Kolve and
                  Aniruddha Kembhavi and
                  Roozbeh Mottaghi},
  title        = {{ProcTHOR}: Large-Scale Embodied
                  {AI} Using Procedural Generation},
  booktitle    = NIPS,
  year         = {2022}
}

@inproceedings{DBLP:conf/iclr/abs-2407-08725,
  author       = {Wayne Wu and
                  Honglin He and
                  Yiran Wang and
                  Chenda Duan and
                  Jack He and
                  Zhizheng Liu and
                  Quanyi Li and
                  Bolei Zhou},
  title        = {{MetaUrban}: {A} Simulation Platform for Embodied {AI} in Urban Spaces},
  booktitle    = ICLR,
  year         = {2025}
}

@inproceedings{DBLP:conf/corl/DosovitskiyRCLK17,
  author       = {Alexey Dosovitskiy and
                  Germ{\'{a}}n Ros and
                  Felipe Codevilla and
                  Antonio M. L{\'{o}}pez and
                  Vladlen Koltun},
  title        = {{CARLA:} An Open Urban Driving Simulator},
  booktitle    = CORL,
  year         = {2017}
}

@inproceedings{DBLP:conf/rss/abs-2410-24164,
  author       = {Kevin Black and
                  Noah Brown and
                  Danny Driess and
                  Adnan Esmail and
                  Michael Equi and
                  Chelsea Finn and
                  Niccolo Fusai and
                  Lachy Groom and
                  Karol Hausman and
                  Brian Ichter and
                  Szymon Jakubczak and
                  Tim Jones and
                  Liyiming Ke and
                  Sergey Levine and
                  Adrian Li{-}Bell and
                  Mohith Mothukuri and
                  Suraj Nair and
                  Karl Pertsch and
                  Lucy Xiaoyang Shi and
                  James Tanner and
                  Quan Vuong and
                  Anna Walling and
                  Haohuan Wang and
                  Ury Zhilinsky},
  title        = {{\(\pi\)}\({}_{\mbox{0}}\): {A} Vision-Language-Action Flow Model for General Robot Control},
  booktitle    = RSS,
  year         = {2025}
}

@article{DBLP:journals/pami/ChenWCJGL24,
  author       = {Li Chen and
                  Penghao Wu and
                  Kashyap Chitta and
                  Bernhard Jaeger and
                  Andreas Geiger and
                  Hongyang Li},
  title        = {End-to-End Autonomous Driving: Challenges and Frontiers},
  journal      = PAMI,
  volume       = {46},
  number       = {12},
  pages        = {10164--10183},
  year         = {2024}
}

@article{DBLP:journals/tomccap/HendrikxMVI13,
  author       = {Mark Hendrikx and
                  Sebastiaan A. Meijer and
                  Joeri Van Der Velden and
                  Alexandru Iosup},
  title        = {Procedural content generation for games: {A} survey},
  journal      = {{ACM} Transactions on Multimedia Computing, Communications, and Applications},
  volume       = {9},
  number       = {1},
  pages        = {1:1--1:22},
  year         = {2013}
}

@inproceedings{DBLP:journals/corr/abs-2412-15200,
  author       = {Wang Zhao and
                  Yan{-}Pei Cao and
                  Jiale Xu and
                  Yuejiang Dong and
                  Ying Shan},
  title        = {{DI-PCG:} Diffusion-based Efficient Inverse Procedural Content Generation for High-quality 3{D} Asset Creation},
  booktitle    = CVPR,
  year         = {2025}
}

@article{DBLP:journals/corr/abs-2501-10928,
  title        = {Generative Physical {AI} in Vision: A Survey},
  author       = {Liu, Daochang and 
                  Zhang, Junyu and 
                  Dinh, Anh-Dung and 
                  Park, Eunbyung and 
                  Zhang, Shichao and 
                  Xu, Chang},
  journal      = {arXiv},
  volume       = {2501.10928},
  year         = {2025}
}

@article{DBLP:journals/corr/abs-2501-03575,
  title        = {Cosmos world foundation model platform for physical {AI}},
  author       = {Agarwal, Niket and 
                  Ali, Arslan and 
                  Bala, Maciej and 
                  Balaji, Yogesh and 
                  Barker, Erik and 
                  Cai, Tiffany and 
                  Chattopadhyay, Prithvijit and 
                  Chen, Yongxin and 
                  Cui, Yin and 
                  Ding, Yifan and 
                  others},
  journal      = {arXiv},
  volume       = {2501.03575},
  year={2025}
}

@article{DBLP:journals/corr/abs-2405-03520,
  author       = {Zheng Zhu and
                  Xiaofeng Wang and
                  Wangbo Zhao and
                  Chen Min and
                  Nianchen Deng and
                  Min Dou and
                  Yuqi Wang and
                  Botian Shi and
                  Kai Wang and
                  Chi Zhang and
                  Yang You and
                  Zhaoxiang Zhang and
                  Dawei Zhao and
                  Liang Xiao and
                  Jian Zhao and
                  Jiwen Lu and
                  Guan Huang},
  title        = {Is Sora a World Simulator? {A} Comprehensive Survey on General World
                  Models and Beyond},
  journal      = {arXiv},
  volume       = {2405.03520},
  year         = {2024}
}

@book{lavalle2023virtual,
  title={Virtual reality},
  author={LaValle, Steven M},
  year={2023},
  publisher={Cambridge university press}
}

@book{mendiburu20123d,
  title={3{D} movie making: stereoscopic digital cinema from script to screen},
  author={Mendiburu, Bernard},
  year={2012},
  publisher={Routledge}
}

@article{DBLP:journals/fthci/LeeBZWXLKBH24,
  author       = {Lik{-}Hang Lee and
                  Tristan Braud and
                  Peng Yuan Zhou and
                  Lin Wang and
                  Dianlei Xu and
                  Zijun Lin and
                  Abhishek Kumar and
                  Carlos Bermejo and
                  Pan Hui},
  title        = {All One Needs to Know about Metaverse: {A} Complete Survey on Technological
                  Singularity, Virtual Ecosystem, and Research Agenda},
  journal      = {Foundations and Trends in Human-Computer Interaction},
  volume       = {18},
  number       = {2-3},
  pages        = {100--337},
  year         = {2024}
}

@inproceedings{DBLP:conf/cvpr/RaistrickLMMWZK23,
  author       = {Alexander Raistrick and
                  Lahav Lipson and
                  Zeyu Ma and
                  Lingjie Mei and
                  Mingzhe Wang and
                  Yiming Zuo and
                  Karhan Kayan and
                  Hongyu Wen and
                  Beining Han and
                  Yihan Wang and
                  Alejandro Newell and
                  Hei Law and
                  Ankit Goyal and
                  Kaiyu Yang and
                  Jia Deng},
  title        = {Infinite Photorealistic Worlds Using Procedural Generation},
  booktitle    = CVPR,
  year         = {2023}
}

@article{DBLP:journals/air/SolimanADH24,
  author       = {Mona M. Soliman and
                  Eman Ahmed and
                  Ashraf Darwish and
                  Aboul Ella Hassanien},
  title        = {Artificial intelligence powered Metaverse: analysis, challenges and
                  future perspectives},
  journal      = {Artificial Intelligence Review},
  volume       = {57},
  number       = {2},
  pages        = {36},
  year         = {2024}
}

@article{DBLP:journals/air/Anantrasirichai22,
  author       = {Nantheera Anantrasirichai and
                  David Bull},
  title        = {Artificial intelligence in the creative industries: a review},
  journal      = {Artificial Intelligence Review},
  volume       = {55},
  number       = {1},
  pages        = {589--656},
  year         = {2022}
}

@book{yannakakis2018artificial,
  title={Artificial intelligence and games},
  author={Yannakakis, Georgios N and Togelius, Julian},
  volume={2},
  year={2018},
  publisher={Springer}
}

@book{short2017procedural,
  title={Procedural generation in game design},
  author={Short, Tanya and Adams, Tarn},
  year={2017},
  publisher={CRC Press}
}

@inproceedings{DBLP:conf/iccv/ChangCLMNT21,
  author       = {Kai{-}Hung Chang and
                  Chin{-}Yi Cheng and
                  Jieliang Luo and
                  Shingo Murata and
                  Mehdi Nourbakhsh and
                  Yoshito Tsuji},
  title        = {{Building-GAN}: Graph-Conditioned Architectural Volumetric Design Generation},
  booktitle    = ICCV,
  year         = {2021}
}

@inproceedings{DBLP:conf/cvpr/LuLCOPQ20,
  author       = {Xiaohu Lu and
                  Zuoyue Li and
                  Zhaopeng Cui and
                  Martin R. Oswald and
                  Marc Pollefeys and
                  Rongjun Qin},
  title        = {Geometry-Aware Satellite-to-Ground Image Synthesis for Urban Areas},
  booktitle    = CVPR,
  year         = {2020}
}

@inproceedings{DBLP:conf/iccv/LiLCQPO21,
  author       = {Zuoyue Li and
                  Zhenqiang Li and
                  Zhaopeng Cui and
                  Rongjun Qin and
                  Marc Pollefeys and
                  Martin R. Oswald},
  title        = {Sat2Vid: Street-view Panoramic Video Synthesis from a Single Satellite
                  Image},
  booktitle    = ICCV,
  year         = {2021}
}

@article{DBLP:journals/pami/ShiCYL22,
  author       = {Yujiao Shi and
                  Dylan Campbell and
                  Xin Yu and
                  Hongdong Li},
  title        = {Geometry-Guided Street-View Panorama Synthesis From Satellite Imagery},
  journal      = PAMI,
  volume       = {44},
  number       = {12},
  pages        = {10009--10022},
  year         = {2022}
}

@article{DBLP:journals/tmm/WuTJZQSY23,
  author       = {Songsong Wu and
                  Hao Tang and
                  Xiao{-}Yuan Jing and
                  Haifeng Zhao and
                  Jianjun Qian and
                  Nicu Sebe and
                  Yan Yan},
  title        = {Cross-View Panorama Image Synthesis},
  journal      = TMM,
  volume       = {25},
  pages        = {3546--3559},
  year         = {2023}
}

@inproceedings{DBLP:conf/iccv/QianXX023,
  author       = {Ming Qian and
                  Jincheng Xiong and
                  Gui{-}Song Xia and
                  Nan Xue},
  title        = {{Sat2Density}: Faithful Density Learning from Satellite-Ground Image
                  Pairs},
  booktitle    = ICCV,
  year         = {2023}
}

@inproceedings{DBLP:conf/eccv/XuQ24,
  author       = {Ningli Xu and
                  Rongjun Qin},
  title        = {Geospecific View Generation Geometry-Context Aware High-Resolution
                  Ground View Inference from Satellite Views},
  booktitle    = ECCV,
  year         = {2024}
}

@inproceedings{DBLP:conf/wacv/SumantriP20,
  author       = {Julius Surya Sumantri and
                  In Kyu Park},
  title        = {360 Panorama Synthesis from a Sparse Set of Images with Unknown Field
                  of View},
  booktitle    = WACV,
  year         = {2020}
}

@inproceedings{DBLP:conf/cvpr/SomanathK21,
  author       = {Gowri Somanath and
                  Daniel Kurz},
  title        = {{HDR} Environment Map Estimation for Real-Time Augmented Reality},
  booktitle    = CVPR,
  year         = {2021}
}

@inproceedings{DBLP:conf/aaai/HaraMH21,
  author       = {Takayuki Hara and
                  Yusuke Mukuta and
                  Tatsuya Harada},
  title        = {Spherical Image Generation from a Single Image by Considering Scene
                  Symmetry},
  booktitle    = AAAI,
  year         = {2021}
}

@inproceedings{DBLP:conf/cvpr/AkimotoMA22,
  author       = {Naofumi Akimoto and
                  Yuhi Matsuo and
                  Yoshimitsu Aoki},
  title        = {Diverse Plausible 360-Degree Image Outpainting for Efficient {3DCG}
                  Background Creation},
  booktitle    = CVPR,
  year         = {2022}
}

@inproceedings{DBLP:conf/eccv/OhCCPWY22,
  author       = {Changgyoon Oh and
                  Wonjune Cho and
                  Yujeong Chae and
                  Daehee Park and
                  Lin Wang and
                  Kuk{-}Jin Yoon},
  title        = {{BIPS:} Bi-modal Indoor Panorama Synthesis via Residual Depth-Aided
                  Adversarial Learning},
  booktitle    = ECCV,
  year         = {2022}
}

@inproceedings{DBLP:conf/3dim/DastjerdiHEKL22,
  author       = {Mohammad Reza Karimi Dastjerdi and
                  Yannick Hold{-}Geoffroy and
                  Jonathan Eisenmann and
                  Siavash Khodadadeh and
                  Jean{-}Fran{\c{c}}ois Lalonde},
  title        = {Guided Co-Modulated {GAN} for 360{\textdegree} Field of View Extrapolation},
  booktitle    = ThreeDV,
  year         = {2022}
}

@inproceedings{DBLP:journals/corr/abs-2501-17162,
      title={Cube{D}iff: Repurposing Diffusion-Based Image Models for Panorama Generation}, 
      author={Nikolai Kalischek and Michael Oechsle and Fabian Manhardt and Philipp Henzler and Konrad Schindler and Federico Tombari},
      booktitle=ICLR,
      year={2025}
}

@article{DBLP:journals/corr/abs-2305-10853,
  author       = {Gabriela Ben Melech Stan and
                  Diana Wofk and
                  Scottie Fox and
                  Alex Redden and
                  Will Saxton and
                  Jean Yu and
                  Estelle Aflalo and
                  Shao{-}Yen Tseng and
                  Fabio Nonato and
                  Matthias M{\"{u}}ller and
                  Vasudev Lal},
  title        = {{LDM3D:} Latent Diffusion Model for 3{D}},
  journal      = {arXiv},
  volume       = {2305.10853},
  year         = {2023}
}

@inproceedings{DBLP:conf/nips/LiB23,
  author       = {Jialu Li and
                  Mohit Bansal},
  title        = {{PanoGen}: Text-Conditioned Panoramic Environment Generation for Vision-and-Language Navigation},
  booktitle    = NIPS,
  year         = {2023}
}

@article{DBLP:journals/corr/abs-2311-13141,
  author       = {Mengyang Feng and
                  Jinlin Liu and
                  Miaomiao Cui and
                  Xuansong Xie},
  title        = {Diffusion360: Seamless 360 Degree Panoramic Image Generation based
                  on Diffusion Models},
  journal      = {arXiv},
  volume       = {2311.13141},
  year         = {2023}
}

@inproceedings{DBLP:conf/iclr/WuZC24,
  author       = {Tianhao Wu and
                  Chuanxia Zheng and
                  Tat{-}Jen Cham},
  title        = {{PanoDiffusion}: 360-degree Panorama Outpainting via Diffusion},
  booktitle    = ICLR,
  year         = {2024}
}

@article{DBLP:journals/tvcg/AiCLCMZKHW24,
  author       = {Hao Ai and
                  Zidong Cao and
                  Haonan Lu and
                  Chen Chen and
                  Jian Ma and
                  Pengyuan Zhou and
                  Tae{-}Kyun Kim and
                  Pan Hui and
                  Lin Wang},
  title        = {Dream360: Diverse and Immersive Outdoor Virtual Scene Creation via
                  Transformer-Based 360{\textdegree} Image Outpainting},
  journal      = TVCG,
  volume       = {30},
  number       = {5},
  pages        = {2734--2744},
  year         = {2024}
}

@inproceedings{DBLP:conf/cvpr/ZhangWGH0O024,
  author       = {Cheng Zhang and
                  Qianyi Wu and
                  Camilo Cruz Gambardella and
                  Xiaoshui Huang and
                  Dinh Phung and
                  Wanli Ouyang and
                  Jianfei Cai},
  title        = {Taming Stable Diffusion for Text to 360{\textdegree} Panorama Image
                  Generation},
  booktitle    = CVPR,
  year         = {2024}
}

@inproceedings{DBLP:conf/nips/YeJ00HZOHZ024,
  author       = {Weicai Ye and
                  Chenhao Ji and
                  Zheng Chen and
                  Junyao Gao and
                  Xiaoshui Huang and
                  Song{-}Hai Zhang and
                  Wanli Ouyang and
                  Tong He and
                  Cairong Zhao and
                  Guofeng Zhang},
  title        = {{DiffPano}: Scalable and Consistent Text to Panorama Generation with
                  Spherical Epipolar-Aware Diffusion},
  booktitle    = NIPS,
  year         = {2024}
}

@inproceedings{DBLP:conf/cvpr/SchultTHWWMLWWH24,
  author       = {Jonas Schult and
                  Sam S. Tsai and
                  Lukas H{\"{o}}llein and
                  Bichen Wu and
                  Jialiang Wang and
                  Chih{-}Yao Ma and
                  Kunpeng Li and
                  Xiaofang Wang and
                  Felix Wimbauer and
                  Zijian He and
                  Peizhao Zhang and
                  Bastian Leibe and
                  Peter Vajda and
                  Ji Hou},
  title        = {{ControlRoom3D}: Room Generation Using Semantic Proxy Rooms},
  booktitle    = CVPR,
  year         = {2024}
}

@inproceedings{DBLP:conf/ijcai/MaZJ24,
  author       = {Yikun Ma and
                  Dandan Zhan and
                  Zhi Jin},
  title        = {{FastScene}: Text-Driven Fast Indoor 3{D} Scene Generation via Panoramic
                  {G}aussian Splatting},
  booktitle    = IJCAI,
  year         = {2024}
}

@inproceedings{DBLP:conf/eccv/ZhouFXCCBYWK24,
  author       = {Shijie Zhou and
                  Zhiwen Fan and
                  Dejia Xu and
                  Haoran Chang and
                  Pradyumna Chari and
                  Tejas Bharadwaj and
                  Suya You and
                  Zhangyang Wang and
                  Achuta Kadambi},
  title        = {Dream{S}cene360: Unconstrained Text-to-3{D} Scene Generation with Panoramic
                  {G}aussian Splatting},
  booktitle    = ECCV,
  year         = {2024}
}

@article{DBLP:journals/corr/abs-2407-15187,
  author       = {Haiyang Zhou and
                  Xinhua Cheng and
                  Wangbo Yu and
                  Yonghong Tian and
                  Li Yuan},
  title        = {{HoloDreamer}: Holistic 3{D} Panoramic World Generation from Text Descriptions},
  journal      = {arXiv},
  volume       = {2407.15187},
  year         = {2024}
}

@article{DBLP:journals/corr/abs-2408-13711,
  author       = {Wenrui Li and
                  Yapeng Mi and
                  Fucheng Cai and
                  Zhe Yang and
                  Wangmeng Zuo and
                  Xingtao Wang and
                  Xiaopeng Fan},
  title        = {{SceneDreamer360}: Text-Driven 3{D}-Consistent Scene Generation with Panoramic {G}aussian Splatting},
  journal      = {arXiv},
  volume       = {2408.13711},
  year         = {2024}
}

@article{DBLP:journals/corr/abs-2407-10923,
  author       = {Penglei Gao and
                  Kai Yao and
                  Tiandi Ye and
                  Steven Wang and
                  Yuan Yao and
                  Xiaofeng Wang},
  title        = {OPa-Ma: Text Guided Mamba for 360-degree Image Out-painting},
  journal      = {arXiv},
  volume       = {2407.10923},
  year         = {2024}
}

@inproceedings{DBLP:conf/wacv/WangXFX24,
  author       = {Hai Wang and
                  Xiaoyu Xiang and
                  Yuchen Fan and
                  Jing{-}Hao Xue},
  title        = {Customizing 360-Degree Panoramas through Text-to-Image Diffusion Models},
  booktitle    = WACV,
  year         = {2024}
}

@inproceedings{DBLP:conf/mm/WangCLX023,
  author       = {Jionghao Wang and
                  Ziyu Chen and
                  Jun Ling and
                  Rong Xie and
                  Li Song},
  title        = {360-Degree Panorama Generation from Few Unregistered NFoV Images},
  booktitle    = ACMMM,
  year         = {2023}
}

@article{DBLP:journals/pami/HaraMH23,
  author       = {Takayuki Hara and
                  Yusuke Mukuta and
                  Tatsuya Harada},
  title        = {Spherical Image Generation From a Few Normal-Field-of-View Images
                  by Considering Scene Symmetry},
  journal      = PAMI,
  volume       = {45},
  number       = {5},
  pages        = {6339--6353},
  year         = {2023}
}

@inproceedings{DBLP:conf/icip/AkimotoKHA19,
  author       = {Naofumi Akimoto and
                  Seito Kasai and
                  Masaki Hayashi and
                  Yoshimitsu Aoki},
  title        = {360-Degree Image Completion by Two-Stage Conditional Gans},
  booktitle    = ICIP,
  year         = {2019}
}

@inproceedings{DBLP:journals/corr/abs-2408-13252,
  author       = {Shuai Yang and
                  Jing Tan and
                  Mengchen Zhang and
                  Tong Wu and
                  Yixuan Li and
                  Gordon Wetzstein and
                  Ziwei Liu and
                  Dahua Lin},
  title        = {{LayerPano3D}: Layered 3{D} Panorama for Hyper-Immersive Scene Generation},
  booktitle    = SIGGRAPH,
  year         = {2025}
}

@inproceedings{DBLP:conf/cvpr/EsserRO21,
  author       = {Patrick Esser and
                  Robin Rombach and
                  Bj{\"{o}}rn Ommer},
  title        = {Taming Transformers for High-Resolution Image Synthesis},
  booktitle    = CVPR,
  year         = {2021}
}

@inproceedings{DBLP:conf/iclr/ZhaoCSDLCX21,
  author       = {Shengyu Zhao and
                  Jonathan Cui and
                  Yilun Sheng and
                  Yue Dong and
                  Xiao Liang and
                  Eric I{-}Chao Chang and
                  Yan Xu},
  title        = {Large Scale Image Completion via Co-Modulated Generative Adversarial
                  Networks},
  booktitle    = ICLR,
  year         = {2021}
}

@inproceedings{DBLP:conf/cvpr/RombachBLEO22,
  author       = {Robin Rombach and
                  Andreas Blattmann and
                  Dominik Lorenz and
                  Patrick Esser and
                  Bj{\"{o}}rn Ommer},
  title        = {High-Resolution Image Synthesis with Latent Diffusion Models},
  booktitle    = CVPR,
  year         = {2022}
}

@inproceedings{DBLP:conf/icml/RadfordKHRGASAM21,
  author       = {Alec Radford and
                  Jong Wook Kim and
                  Chris Hallacy and
                  Aditya Ramesh and
                  Gabriel Goh and
                  Sandhini Agarwal and
                  Girish Sastry and
                  Amanda Askell and
                  Pamela Mishkin and
                  Jack Clark and
                  Gretchen Krueger and
                  Ilya Sutskever},
  title        = {Learning Transferable Visual Models From Natural Language Supervision},
  booktitle    = ICML,
  year         = {2021}
}

@inproceedings{DBLP:conf/icml/Bar-TalYLD23,
  author       = {Omer Bar{-}Tal and
                  Lior Yariv and
                  Yaron Lipman and
                  Tali Dekel},
  title        = {{MultiDiffusion}: Fusing Diffusion Paths for Controlled Image Generation},
  booktitle    = ICML,
  year         = {2023}
}

@inproceedings{DBLP:conf/aaai/LuHW0024,
  author       = {Zhuqiang Lu and
                  Kun Hu and
                  Chaoyue Wang and
                  Lei Bai and
                  Zhiyong Wang},
  title        = {Autoregressive Omni-Aware Outpainting for Open-Vocabulary 360-Degree Image Generation},
  booktitle    = AAAI, 
  year         = {2024}
}

@inproceedings{DBLP:conf/cvpr/WilesGS020,
  author       = {Olivia Wiles and
                  Georgia Gkioxari and
                  Richard Szeliski and
                  Justin Johnson},
  title        = {{SynSin}: End-to-End View Synthesis From a Single Image},
  booktitle    = CVPR,
  year         = {2020}
}

@inproceedings{DBLP:conf/cvpr/TuckerS20,
  author       = {Richard Tucker and
                  Noah Snavely},
  title        = {Single-View View Synthesis With Multiplane Images},
  booktitle    = CVPR,
  year         = {2020}
}

@inproceedings{DBLP:conf/cvpr/ShihSKH20,
  author       = {Meng{-}Li Shih and
                  Shih{-}Yang Su and
                  Johannes Kopf and
                  Jia{-}Bin Huang},
  title        = {{3D} Photography Using Context-Aware Layered Depth Inpainting},
  booktitle    = CVPR,
  year         = {2020}
}

@inproceedings{DBLP:conf/nips/HabtegebrialJGS20,
  author       = {Tewodros Amberbir Habtegebrial and
                  Varun Jampani and
                  Orazio Gallo and
                  Didier Stricker},
  title        = {Generative View Synthesis: From Single-view Semantics to Novel-view
                  Images},
  booktitle    = NIPS,
  year         = {2020}
}

@inproceedings{DBLP:conf/iccv/RockwellF021,
  author       = {Chris Rockwell and
                  David F. Fouhey and
                  Justin Johnson},
  title        = {Pixel{S}ynth: Generating a 3{D}-Consistent Experience from a Single Image},
  booktitle    = ICCV,
  year         = {2021}
}

@inproceedings{DBLP:conf/iccv/HuRBP21,
  author       = {Ronghang Hu and
                  Nikhila Ravi and
                  Alexander C. Berg and
                  Deepak Pathak},
  title        = {Worldsheet: Wrapping the World in a 3{D} Sheet for View Synthesis from
                  a Single Image},
  booktitle    = ICCV,
  year         = {2021}
}

@inproceedings{DBLP:conf/iccv/RombachEO21,
  author       = {Robin Rombach and
                  Patrick Esser and
                  Bj{\"{o}}rn Ommer},
  title        = {Geometry-Free View Synthesis: Transformers and no 3{D} Priors},
  booktitle    = ICCV,
  year         = {2021}
}

@inproceedings{DBLP:conf/aaai/KohABTWLYBA23,
  author       = {Jing Yu Koh and
                  Harsh Agrawal and
                  Dhruv Batra and
                  Richard Tucker and
                  Austin Waters and
                  Honglak Lee and
                  Yinfei Yang and
                  Jason Baldridge and
                  Peter Anderson},
  title        = {Simple and Effective Synthesis of Indoor 3{D} Scenes},
  booktitle    = AAAI,
  year         = {2023}
}

@inproceedings{DBLP:conf/cvpr/TsengL0A0023,
  author       = {Hung{-}Yu Tseng and
                  Qinbo Li and
                  Changil Kim and
                  Suhib Alsisan and
                  Jia{-}Bin Huang and
                  Johannes Kopf},
  title        = {Consistent View Synthesis with Pose-Guided Diffusion Models},
  booktitle    = CVPR,
  year         = {2023}
}

@inproceedings{DBLP:conf/cvpr/Li0S0XL23,
  author       = {Xingyi Li and
                  Zhiguo Cao and
                  Huiqiang Sun and
                  Jianming Zhang and
                  Ke Xian and
                  Guosheng Lin},
  title        = {3{D} Cinemagraphy from a Single Image},
  booktitle    = CVPR,
  year         = {2023}
}

@inproceedings{DBLP:conf/cvpr/RenW22,
  author       = {Xuanchi Ren and
                  Xiaolong Wang},
  title        = {Look Outside the Room: Synthesizing {A} Consistent Long-Term 3{D} Scene
                  Video from {A} Single Image},
  booktitle    = CVPR,
  year         = {2022}
}

@inproceedings{DBLP:conf/iccv/CaiCPSOGW23,
  author       = {Shengqu Cai and
                  Eric Ryan Chan and
                  Songyou Peng and
                  Mohamad Shahbazi and
                  Anton Obukhov and
                  Luc Van Gool and
                  Gordon Wetzstein},
  title        = {Diff{D}reamer: Towards Consistent Unsupervised Single-view Scene Extrapolation with Conditional Diffusion Models},
  booktitle    = ICCV,
  year         = {2023}
}

@inproceedings{DBLP:conf/nips/FridmanAKD23,
  author       = {Rafail Fridman and
                  Amit Abecasis and
                  Yoni Kasten and
                  Tali Dekel},
  title        = {Scene{S}cape: Text-Driven Consistent Scene Generation},
  booktitle    = NIPS,
  year         = {2023}
}

@inproceedings{DBLP:conf/mm/ShenLSPX0L23,
  author       = {Liao Shen and
                  Xingyi Li and
                  Huiqiang Sun and
                  Juewen Peng and
                  Ke Xian and
                  Zhiguo Cao and
                  Guosheng Lin},
  title        = {Make-It-4{D}: Synthesizing a Consistent Long-Term Dynamic Scene Video
                  from a Single Image},
  booktitle    = ACMMM,
  year         = {2023}
}

@inproceedings{DBLP:conf/cvpr/YuDHSRFCSS0H24,
  author       = {Hong{-}Xing Yu and
                  Haoyi Duan and
                  Junhwa Hur and
                  Kyle Sargent and
                  Michael Rubinstein and
                  William T. Freeman and
                  Forrester Cole and
                  Deqing Sun and
                  Noah Snavely and
                  Jiajun Wu and
                  Charles Herrmann},
  title        = {Wonder{J}ourney: Going from Anywhere to Everywhere},
  booktitle    = CVPR,
  year         = {2024}
}

@inproceedings{DBLP:journals/corr/abs-2406-09394,
  author       = {Hong{-}Xing Yu and
                  Haoyi Duan and
                  Charles Herrmann and
                  William T. Freeman and
                  Jiajun Wu},
  title        = {Wonder{W}orld: Interactive 3{D} Scene Generation from a Single Image},
  booktitle    = CVPR,
  year         = {2025}
}

@inproceedings{DBLP:conf/iccv/HolleinCO0N23,
  author       = {Lukas H{\"{o}}llein and
                  Ang Cao and
                  Andrew Owens and
                  Justin Johnson and
                  Matthias Nie{\ss}ner},
  title        = {Text2{R}oom: Extracting Textured 3{D} Meshes from 2{D} Text-to-Image Models},
  booktitle    = ICCV,
  year         = {2023}
}

@inproceedings{DBLP:conf/mm/LiWCPWXWCL24,
  author       = {Xingyi Li and
                  Yizheng Wu and
                  Jun Cen and
                  Juewen Peng and
                  Kewei Wang and
                  Ke Xian and
                  Zhe Wang and
                  Zhiguo Cao and
                  Guosheng Lin},
  title        = {{iControl3D}: An Interactive System for Controllable 3{D} Scene Generation},
  booktitle    = ACMMM,
  year         = {2024}
}

@article{DBLP:journals/tvcg/ZhangLWWL24,
  author       = {Jing{-}Bo Zhang and
                  Xiaoyu Li and
                  Ziyu Wan and
                  C. Y. Wang and
                  Jing Liao},
  title        = {{Text2NeRF}: Text-Driven 3{D} Scene Generation With Neural Radiance Fields},
  journal      = TVCG,
  year         = {2024}
}

@article{DBLP:journals/corr/abs-2311-13384,
  author       = {Jaeyoung Chung and
                  Suyoung Lee and
                  Hyeongjin Nam and
                  Jaerin Lee and
                  Kyoung Mu Lee},
  title        = {Lucid{D}reamer: Domain-free Generation of 3{D} {G}aussian Splatting Scenes},
  journal      = TVCG,
  volume       = {31},
  number       = {12},
  pages        = {10640--10651},
  year         = {2025}
}

@inproceedings{DBLP:conf/cvpr/ZhangZ0MHBXZ24,
  author       = {Songchun Zhang and
                  Yibo Zhang and
                  Quan Zheng and
                  Rui Ma and
                  Wei Hua and
                  Hujun Bao and
                  Weiwei Xu and
                  Changqing Zou},
  title        = {{3D-SceneDreamer}: Text-Driven 3{D}-Consistent Scene Generation},
  booktitle    = CVPR,
  year         = {2024}
}

@inproceedings{DBLP:journals/corr/abs-2408-05477,
  author       = {Yiying Yang and
                  Fukun Yin and
                  Jiayuan Fan and
                  Xin Chen and
                  Wanzhang Li and
                  Gang Yu},
  title        = {Scene123: One Prompt to 3{D} Scene Generation via Video-Assisted and
                  Consistency-Enhanced {MAE}},
  booktitle    = ACMMM,
  year         = {2025}
}

@inproceedings{DBLP:journals/corr/abs-2404-07199,
  author       = {Jaidev Shriram and
                  Alex Trevithick and
                  Lingjie Liu and
                  Ravi Ramamoorthi},
  title        = {{RealmDreamer:} Text-Driven 3{D} Scene Generation with Inpainting and
                  Depth Diffusion},
  booktitle    = ThreeDV,
  year         = {2025}
}

@inproceedings{DBLP:conf/eccv/LiuLCLXP24,
  author       = {Aoming Liu and
                  Zhong Li and
                  Zhang Chen and
                  Nannan Li and
                  Yi Xu and
                  Bryan A. Plummer},
  title        = {{PanoFree:} Tuning-Free Holistic Multi-view Image Generation with Cross-View Self-guidance},
  booktitle    = ECCV,
  year         = {2024}
}

@article{DBLP:journals/tog/NiklausMYL19,
  author       = {Simon Niklaus and
                  Long Mai and
                  Jimei Yang and
                  Feng Liu},
  title        = {3{D} Ken Burns effect from a single image},
  journal      = TOG,
  volume       = {38},
  number       = {6},
  pages        = {184:1--184:15},
  year         = {2019}
}

@article{DBLP:journals/pami/RanftlLHSK22,
  author       = {Ren{\'{e}} Ranftl and
                  Katrin Lasinger and
                  David Hafner and
                  Konrad Schindler and
                  Vladlen Koltun},
  title        = {Towards Robust Monocular Depth Estimation: Mixing Datasets for Zero-Shot Cross-Dataset Transfer},
  journal      = PAMI,
  volume       = {44},
  number       = {3},
  pages        = {1623--1637},
  year         = {2022}
}

@article{DBLP:journals/corr/abs-2302-12288,
  author       = {Shariq Farooq Bhat and
                  Reiner Birkl and
                  Diana Wofk and
                  Peter Wonka and
                  Matthias M{\"{u}}ller},
  title        = {{ZoeDepth:} Zero-shot Transfer by Combining Relative and Metric Depth},
  journal      = {arXiv},
  volume       = {2302.12288},
  year         = {2023}
}

@inproceedings{DBLP:conf/cvpr/MiangolehDMPA21,
  author       = {S. Mahdi H. Miangoleh and
                  Sebastian Dille and
                  Long Mai and
                  Sylvain Paris and
                  Yagiz Aksoy},
  title        = {Boosting Monocular Depth Estimation Models to High-Resolution via
                  Content-Adaptive Multi-Resolution Merging},
  booktitle    = CVPR,
  year         = {2021}
}

@inproceedings{DBLP:conf/iccv/ZhangRA23,
  author       = {Lvmin Zhang and
                  Anyi Rao and
                  Maneesh Agrawala},
  title        = {Adding Conditional Control to Text-to-Image Diffusion Models},
  booktitle    = ICCV,
  year         = {2023}
}

@inproceedings{DBLP:conf/eccv/LuHYZZ24,
  author       = {Jiachen Lu and
                  Ze Huang and
                  Zeyu Yang and
                  Jiahui Zhang and
                  Li Zhang},
  title        = {{WoVoGen:} World Volume-Aware Diffusion for Controllable Multi-camera
                  Driving Scene Generation},
  booktitle    = ECCV,
  year         = {2024}
}

@inproceedings{DBLP:conf/eccv/WangZHCZL24,
  author       = {Xiaofeng Wang and
                  Zheng Zhu and
                  Guan Huang and
                  Xinze Chen and
                  Jiwen Lu},
  title        = {Drive{D}reamer: Towards Real-world-driven World Models for Autonomous Driving},
  booktitle    = ECCV,
  year         = {2024}
}

@inproceedings{DBLP:conf/eccv/LiZY24,
  author       = {Xiaofan Li and
                  Yifu Zhang and
                  Xiaoqing Ye},
  title        = {{DrivingDiffusion:} Layout-Guided Multi-view Driving Scenarios Video Generation with Latent Diffusion Model},
  booktitle    = ECCV,
  year         = {2024}
}

@inproceedings{DBLP:conf/cvpr/WenZLJWLZWS024,
  author       = {Yuqing Wen and
                  Yucheng Zhao and
                  Yingfei Liu and
                  Fan Jia and
                  Yanhui Wang and
                  Chong Luo and
                  Chi Zhang and
                  Tiancai Wang and
                  Xiaoyan Sun and
                  Xiangyu Zhang},
  title        = {Panacea: Panoramic and Controllable Video Generation for Autonomous
                  Driving},
  booktitle    = CVPR,
  year         = {2024}
}

@inproceedings{DBLP:conf/cvpr/00010FLC024,
  author       = {Yuqi Wang and
                  Jiawei He and
                  Lue Fan and
                  Hongxin Li and
                  Yuntao Chen and
                  Zhaoxiang Zhang},
  title        = {Driving Into the Future: Multiview Visual Forecasting and Planning
                  with World Model for Autonomous Driving},
  booktitle    = CVPR,
  year         = {2024}
}

@inproceedings{DBLP:conf/nips/GaoY0CQ0Z024,
  author       = {Shenyuan Gao and
                  Jiazhi Yang and
                  Li Chen and
                  Kashyap Chitta and
                  Yihang Qiu and
                  Andreas Geiger and
                  Jun Zhang and
                  Hongyang Li},
  title        = {Vista: {A} Generalizable Driving World Model with High Fidelity and
                  Versatile Controllability},
  booktitle    = NIPS,
  year         = {2024}
}

@article{DBLP:journals/corr/abs-2410-00337,
  author       = {Leheng Li and
                  Weichao Qiu and
                  Yingjie Cai and
                  Xu Yan and
                  Qing Lian and
                  Bingbing Liu and
                  Ying{-}Cong Chen},
  title        = {Synthe{O}cc: Synthesize Geometric-Controlled Street View Images through
                  3{D} Semantic MPIs},
  journal      = {arXiv},
  volume       = {2410.00337},
  year         = {2024}
}

@inproceedings{DBLP:journals/corr/abs-2411-11252,
  author       = {Tianyi Yan and
                  Dongming Wu and
                  Wencheng Han and
                  Junpeng Jiang and
                  Xia Zhou and
                  Kun Zhan and
                  Cheng{-}zhong Xu and
                  Jianbing Shen},
  title        = {Driving{S}phere: Building a High-fidelity 4{D} World for Closed-loop Simulation},
  booktitle    = CVPR,
  year         = {2025}
}

@article{DBLP:journals/corr/abs-2501-00601,
  author       = {Jiageng Mao and
                  Boyi Li and
                  Boris Ivanovic and
                  Yuxiao Chen and
                  Yan Wang and
                  Yurong You and
                  Chaowei Xiao and
                  Danfei Xu and
                  Marco Pavone and
                  Yue Wang},
  title        = {Dream{D}rive: Generative 4{D} Scene Modeling from Street View Images},
  journal      = {arXiv},
  volume       = {2501.00601},
  year         = {2025}
}

@inproceedings{DBLP:journals/corr/abs-2411-13807,
  author       = {Ruiyuan Gao and Kai Chen and Bo Xiao and Lanqing Hong and Zhenguo Li and Qiang Xu},
  title        = {Magic{D}rive-{V2}: High-Resolution Long Video Generation for Autonomous Driving with Adaptive Control},
  booktitle    = ICCV,
  year         = {2025}
}

@inproceedings{DBLP:journals/corr/abs-2412-03934,
  author       = {Yifan Lu and
                  Xuanchi Ren and
                  Jiawei Yang and
                  Tianchang Shen and
                  Zhangjie Wu and
                  Jun Gao and
                  Yue Wang and
                  Siheng Chen and
                  Mike Chen and
                  Sanja Fidler and
                  Jiahui Huang},
  title        = {Infini{C}ube: Unbounded and Controllable Dynamic 3{D} Driving Scene Generation with World-Guided Video Models},
  booktitle    = ICCV,
  year         = {2025}
}

@inproceedings{DBLP:conf/cvpr/abs-2412-13188,
  author       = {Yunzhi Yan and
                  Zhen Xu and
                  Haotong Lin and
                  Haian Jin and
                  Haoyu Guo and
                  Yida Wang and
                  Kun Zhan and
                  Xianpeng Lang and
                  Hujun Bao and
                  Xiaowei Zhou and
                  Sida Peng},
  title        = {Street{C}rafter: Street View Synthesis with Controllable Video Diffusion
                  Models},
  booktitle    = CVPR,
  year         = {2025}
}

@article{DBLP:journals/corr/abs-2409-04003,
  author       = {Jianbiao Mei and
                  Yukai Ma and
                  Xuemeng Yang and
                  Licheng Wen and
                  Tiantian Wei and
                  Min Dou and
                  Botian Shi and
                  Yong Liu},
  title        = {Dream{F}orge: Motion-Aware Autoregressive Video Generation for Multi-View Driving Scenes},
  journal      = {arXiv},
  volume       = {2409.04003},
  year         = {2024}
}

@inproceedings{DBLP:conf/cvpr/abs-2409-05463,
  author       = {Wei Wu and
                  Xi Guo and
                  Weixuan Tang and
                  Tingxuan Huang and
                  Chiyu Wang and
                  Dongyue Chen and
                  Chenjing Ding},
  title        = {Drive{S}cape: Towards High-Resolution Controllable Multi-View Driving
                  Video Generation},
  booktitle    = CVPR,
  year         = {2025}
}

@article{DBLP:journals/corr/abs-2408-00415,
  author       = {Xuemeng Yang and
                  Licheng Wen and
                  Yukai Ma and
                  Jianbiao Mei and
                  Xin Li and
                  Tiantian Wei and
                  Wenjie Lei and
                  Daocheng Fu and
                  Pinlong Cai and
                  Min Dou and
                  Botian Shi and
                  Liang He and
                  Yong Liu and
                  Yu Qiao},
  title        = {Drive{A}rena: {A} Closed-loop Generative Simulation Platform for Autonomous Driving},
  journal      = {arXiv},
  volume       = {2408.00415},
  year         = {2024}
}

@article{DBLP:journals/corr/abs-2412-03552,
  author       = {Jing Tan and
                  Shuai Yang and
                  Tong Wu and
                  Jingwen He and
                  Yuwei Guo and
                  Ziwei Liu and
                  Dahua Lin},
  title        = {Imagine360: Immersive 360 Video Generation from Perspective Anchor},
  journal      = {arXiv},
  volume       = {2412.03552},
  year         = {2024}
}

@inproceedings{DBLP:conf/nips/YuWZMSCJT024,
  author       = {Heng Yu and
                  Chaoyang Wang and
                  Peiye Zhuang and
                  Willi Menapace and
                  Aliaksandr Siarohin and
                  Junli Cao and
                  L{\'{a}}szl{\'{o}} A. Jeni and
                  Sergey Tulyakov and
                  Hsin{-}Ying Lee},
  title        = {4{R}eal: Towards Photorealistic 4{D} Scene Generation via Video Diffusion Models},
  booktitle    = NIPS,
  year         = {2024}
}

@inproceedings{DBLP:journals/corr/abs-2411-04928,
  author       = {Wenqiang Sun and
                  Shuo Chen and
                  Fangfu Liu and
                  Zilong Chen and
                  Yueqi Duan and
                  Jun Zhang and
                  Yikai Wang},
  title        = {Dimension{X}: Create Any 3{D} and 4{D} Scenes from a Single Image with Controllable Video Diffusion},
  booktitle    = ICCV,
  year         = {2025}
}

@article{DBLP:journals/corr/abs-2409-02048,
  author       = {Wangbo Yu and
                  Jinbo Xing and
                  Li Yuan and
                  Wenbo Hu and
                  Xiaoyu Li and
                  Zhipeng Huang and
                  Xiangjun Gao and
                  Tien{-}Tsin Wong and
                  Ying Shan and
                  Yonghong Tian},
  title        = {View{C}rafter: Taming Video Diffusion Models for High-fidelity Novel View Synthesis},
  journal      = PAMI,
  year         = {2025}
}

@inproceedings{DBLP:journals/corr/abs-2501-05763,
  author       = {Shangjin Zhai and
                  Zhichao Ye and
                  Jialin Liu and
                  Weijian Xie and
                  Jiaqi Hu and
                  Zhen Peng and
                  Hua Xue and
                  Danpeng Chen and
                  Xiaomeng Wang and
                  Lei Yang and
                  Nan Wang and
                  Haomin Liu and
                  Guofeng Zhang},
  title        = {Star{G}en: {A} Spatiotemporal Autoregression Framework with Video Diffusion Model for Scalable and Controllable Scene Generation},
  booktitle    = CVPR,
  year         = {2025}
}

@article{DBLP:journals/corr/abs-2411-11844,
  author       = {Taiming Lu and
                  Tianmin Shu and
                  Alan L. Yuille and
                  Daniel Khashabi and
                  Jieneng Chen},
  title        = {Generative World Explorer},
  journal      = ICLR,
  year         = {2025}
}

@article{DBLP:journals/corr/abs-2411-02319,
  author       = {Yuyang Zhao and
                  Chung{-}Ching Lin and
                  Kevin Lin and
                  Zhiwen Yan and
                  Linjie Li and
                  Zhengyuan Yang and
                  Jianfeng Wang and
                  Gim Hee Lee and
                  Lijuan Wang},
  title        = {{GenXD:} Generating Any 3{D} and 4{D} Scenes},
  journal      = ICLR,
  year         = {2025}
}

@article{DBLP:journals/corr/abs-2405-20334,
  author       = {Yao{-}Chih Lee and
                  Yi{-}Ting Chen and
                  Andrew Wang and
                  Ting{-}Hsuan Liao and
                  Brandon Y. Feng and
                  Jia{-}Bin Huang},
  title        = {Vivid{D}ream: Generating 3{D} Scene with Ambient Dynamics},
  journal      = ICLR,
  year         = {2025}
}

@article{DBLP:journals/cgf/KermaniLTZ16,
  author       = {Z. Sadeghipour Kermani and
                  Z. Liao and
                  P. Tan and
                  Hao Zhang},
  title        = {Learning 3{D} Scene Synthesis from Annotated {RGB-D} Images},
  journal      = CGF,
  volume       = {35},
  number       = {5},
  pages        = {197--206},
  year         = {2016}
}

@article{DBLP:journals/tog/WangLWSCR19,
  author       = {Kai Wang and
                  Yu{-}An Lin and
                  Ben Weissmann and
                  Manolis Savva and
                  Angel X. Chang and
                  Daniel Ritchie},
  title        = {{PlanIT}: planning and instantiating indoor scenes with relation graph
                  and spatial prior networks},
  journal      = TOG,
  volume       = {38},
  number       = {4},
  pages        = {132:1--132:15},
  year         = {2019}
}

@article{DBLP:journals/tog/LiPXCKSTCCZ19,
  author       = {Manyi Li and
                  Akshay Gadi Patil and
                  Kai Xu and
                  Siddhartha Chaudhuri and
                  Owais Khan and
                  Ariel Shamir and
                  Changhe Tu and
                  Baoquan Chen and
                  Daniel Cohen{-}Or and
                  Hao (Richard) Zhang},
  title        = {{GRAINS:} Generative Recursive Autoencoders for INdoor Scenes},
  journal      = TOG,
  volume       = {38},
  number       = {2},
  pages        = {12:1--12:16},
  year         = {2019}
}

@inproceedings{DBLP:conf/iccv/KarPLCYRA0F19,
  author       = {Amlan Kar and
                  Aayush Prakash and
                  Ming{-}Yu Liu and
                  Eric Cameracci and
                  Justin Yuan and
                  Matt Rusiniak and
                  David Acuna and
                  Antonio Torralba and
                  Sanja Fidler},
  title        = {Meta-Sim: Learning to Generate Synthetic Datasets},
  booktitle    = ICCV,
  year         = {2019}
}

@inproceedings{DBLP:conf/eccv/DevaranjanKF20,
  author       = {Jeevan Devaranjan and
                  Amlan Kar and
                  Sanja Fidler},
  title        = {Meta-Sim2: Unsupervised Learning of Scene Structure for Synthetic
                  Data Generation},
  booktitle    = ECCV,
  year         = {2020}
}

@inproceedings{DBLP:conf/cvpr/LuoZ0T20,
  author       = {Andrew Luo and
                  Zhoutong Zhang and
                  Jiajun Wu and
                  Joshua B. Tenenbaum},
  title        = {End-to-End Optimization of Scene Layout},
  booktitle    = CVPR,
  year         = {2020}
}

@inproceedings{DBLP:conf/iccv/DhamoMNT21,
  author       = {Helisa Dhamo and
                  Fabian Manhardt and
                  Nassir Navab and
                  Federico Tombari},
  title        = {Graph-to-3{D}: End-to-End Generation and Manipulation of 3{D} Scenes Using
                  Scene Graphs},
  booktitle    = ICCV,
  year         = {2021}
}

@article{DBLP:journals/pami/GaoSMLGY23,
  author       = {Lin Gao and
                  Jia{-}Mu Sun and
                  Kaichun Mo and
                  Yu{-}Kun Lai and
                  Leonidas J. Guibas and
                  Jie Yang},
  title        = {{SceneHGN}: Hierarchical Graph Networks for 3{D} Indoor Scene Generation
                  With Fine-Grained Geometry},
  journal      = PAMI,
  volume       = {45},
  number       = {7},
  pages        = {8902--8919},
  year         = {2023}
}

@inproceedings{DBLP:conf/nips/ZhaiOWDTNB23,
  author       = {Guangyao Zhai and
                  Evin Pinar {\"{O}}rnek and
                  Shun{-}Cheng Wu and
                  Yan Di and
                  Federico Tombari and
                  Nassir Navab and
                  Benjamin Busam},
  title        = {Common{S}cenes: Generating Commonsense 3{D} Indoor Scenes with Scene Graph Diffusion},
  booktitle    = NIPS,
  year         = {2023}
}

@inproceedings{DBLP:conf/eccv/ZhaiOCLDNTB24,
  author       = {Guangyao Zhai and
                  Evin Pinar {\"{O}}rnek and
                  Dave Zhenyu Chen and
                  Ruotong Liao and
                  Yan Di and
                  Nassir Navab and
                  Federico Tombari and
                  Benjamin Busam},
  title        = {Echo{S}cene: Indoor Scene Generation via Information Echo Over Scene
                  Graph Diffusion},
  booktitle    = ECCV,
  year         = {2024}
}

@inproceedings{DBLP:conf/eccv/WuFWXDMM24,
  author       = {Zijie Wu and
                  Mingtao Feng and
                  Yaonan Wang and
                  He Xie and
                  Weisheng Dong and
                  Bo Miao and
                  Ajmal Mian},
  title        = {External Knowledge Enhanced 3{D} Scene Generation from Sketch},
  booktitle    = ECCV,
  year         = {2024}
}

@inproceedings{DBLP:conf/iclr/LinM24,
  author       = {Chenguo Lin and
                  Yadong Mu},
  title        = {Instruct{S}cene: Instruction-Driven 3{D} Indoor Scene Synthesis with Semantic Graph Prior},
  booktitle    = ICLR,
  year         = {2024}
}

@inproceedings{DBLP:conf/iclr/abs-2408-14837,
  author       = {Dani Valevski and
                  Yaniv Leviathan and
                  Moab Arar and
                  Shlomi Fruchter},
  title        = {Diffusion Models Are Real-Time Game Engines},
  booktitle    = ICLR,
  year         = {2025}
}

@inproceedings{DBLP:conf/nips/AlonsoJMKSPF24,
  author       = {Eloi Alonso and
                  Adam Jelley and
                  Vincent Micheli and
                  Anssi Kanervisto and
                  Amos J. Storkey and
                  Tim Pearce and
                  Fran{\c{c}}ois Fleuret},
  title        = {Diffusion for World Modeling: Visual Details Matter in Atari},
  booktitle    = NIPS,
  year         = {2024}
}

@misc{oasis2024,
  author    = {Decart and Julian Quevedo and Quinn McIntyre and Spruce Campbell and Xinlei Chen and Robert Wachen},
  title     = {Oasis: A Universe in a Transformer},
  year      = {2024},
  url       = {https://oasis-model.github.io/}
}

@article{DBLP:journals/corr/abs-2412-04471,
  author       = {Vinayak Gupta and
                  Yunze Man and
                  Yu{-}Xiong Wang},
  title        = {{PaintScene4D}: Consistent 4{D} Scene Generation from Text Prompts},
  journal      = {arXiv},
  volume       = {2412.04471},
  year         = {2024}
}

@inproceedings{DBLP:journals/corr/abs-2503-00045,
  author       = {Bin Xie and Yingfei Liu and Tiancai Wang and Jiale Cao and Xiangyu Zhang},
  title        = {Glad: A Streaming Scene Generator for Autonomous Driving},
  booktitle    = ICLR,
  year         = {2025}
}

@inproceedings{DBLP:conf/cvpr/WuYFX0000W24,
  author       = {Guanjun Wu and
                  Taoran Yi and
                  Jiemin Fang and
                  Lingxi Xie and
                  Xiaopeng Zhang and
                  Wei Wei and
                  Wenyu Liu and
                  Qi Tian and
                  Xinggang Wang},
  title        = {4{D} {G}aussian Splatting for Real-Time Dynamic Scene Rendering},
  booktitle    = CVPR,
  year         = {2024}
}

@inproceedings{DBLP:conf/eccv/FengDXNABZ24,
  author       = {Haiwen Feng and
                  Zheng Ding and
                  Zhihao Xia and
                  Simon Niklaus and
                  Victoria Fern{\'{a}}ndez Abrevaya and
                  Michael J. Black and
                  Xuaner Zhang},
  title        = {Explorative Inbetweening of Time and Space},
  booktitle    = ECCV,
  year         = {2024}
}

@inproceedings{DBLP:conf/cvpr/YangGZJ0024,
  author       = {Ziyi Yang and
                  Xinyu Gao and
                  Wen Zhou and
                  Shaohui Jiao and
                  Yuqing Zhang and
                  Xiaogang Jin},
  title        = {Deformable 3{D} {G}aussians for High-Fidelity Monocular Dynamic Scene
                  Reconstruction},
  booktitle    = CVPR,
  year         = {2024}
}

@inproceedings{DBLP:conf/emnlp/ChangSM14,
  author       = {Angel X. Chang and
                  Manolis Savva and
                  Christopher D. Manning},
  title        = {Learning Spatial Knowledge for Text to 3{D} Scene Generation},
  booktitle    = {EMNLP},
  year         = {2014}
}

@article{DBLP:journals/tog/FuCWWZF17,
  author       = {Qiang Fu and
                  Xiaowu Chen and
                  Xiaotian Wang and
                  Sijia Wen and
                  Bin Zhou and
                  Hongbo Fu},
  title        = {Adaptive synthesis of indoor scenes via activity-associated object
                  relation graphs},
  journal      = TOG,
  volume       = {36},
  number       = {6},
  pages        = {201:1--201:13},
  year         = {2017}
}

@article{DBLP:journals/tog/MaPFLPHYTGZ18,
  author       = {Rui Ma and
                  Akshay Gadi Patil and
                  Matthew Fisher and
                  Manyi Li and
                  S{\"{o}}ren Pirk and
                  Binh{-}Son Hua and
                  Sai{-}Kit Yeung and
                  Xin Tong and
                  Leonidas J. Guibas and
                  Hao Zhang},
  title        = {Language-driven synthesis of 3{D} scenes from scene databases},
  journal      = TOG,
  volume       = {37},
  number       = {6},
  pages        = {212},
  year         = {2018}
}

@inproceedings{DBLP:conf/iclr/jincjhkkk25,
  author       = {In-Hwan Jin and Haesoo Choo and Seong-Hun Jeong and Heemoon Park and Junghwan Kim and Oh-joon Kwon and Kyeongbo Kong},
  title        = {Optimizing 4{D} {G}aussians for Dynamic Scene Video from Single Landscape Images},
  booktitle    = ICLR,
  year         = {2025}
}

@inproceedings{DBLP:conf/eccv/FuWLS24,
  author       = {Rao Fu and
                  Zehao Wen and
                  Zichen Liu and
                  Srinath Sridhar},
  title        = {Any{H}ome: Open-Vocabulary Generation of Structured and Textured 3{D}
                  Homes},
  booktitle    = ECCV,
  year         = {2024}
}

@article{DBLP:journals/tog/WangSCR18,
  author       = {Kai Wang and
                  Manolis Savva and
                  Angel X. Chang and
                  Daniel Ritchie},
  title        = {Deep convolutional priors for indoor scene synthesis},
  journal      = TOG,
  volume       = {37},
  number       = {4},
  pages        = {70},
  year         = {2018}
}

@inproceedings{DBLP:conf/cvpr/Ritchie0L19,
  author       = {Daniel Ritchie and
                  Kai Wang and
                  Yu{-}An Lin},
  title        = {Fast and Flexible Indoor Scene Synthesis via Deep Convolutional Generative
                  Models},
  booktitle    = CVPR,
  year         = {2019}
}

@article{DBLP:journals/tog/ZhangYMLHVH20,
  author       = {Zaiwei Zhang and
                  Zhenpei Yang and
                  Chongyang Ma and
                  Linjie Luo and
                  Alexander Huth and
                  Etienne Vouga and
                  Qixing Huang},
  title        = {Deep Generative Modeling for Scene Synthesis via Hybrid Representations},
  journal      = TOG,
  volume       = {39},
  number       = {2},
  pages        = {17:1--17:21},
  year         = {2020}
}

@inproceedings{DBLP:conf/iccv/0005ZYHMZBH21,
  author       = {Haitao Yang and
                  Zaiwei Zhang and
                  Siming Yan and
                  Haibin Huang and
                  Chongyang Ma and
                  Yi Zheng and
                  Chandrajit Bajaj and
                  Qixing Huang},
  title        = {Scene Synthesis via Uncertainty-Driven Attribute Synchronization},
  booktitle    = ICCV,
  year         = {2021}
}

@inproceedings{DBLP:conf/nips/PaschalidouKSKG21,
  author       = {Despoina Paschalidou and
                  Amlan Kar and
                  Maria Shugrina and
                  Karsten Kreis and
                  Andreas Geiger and
                  Sanja Fidler},
  title        = {{ATISS:} Autoregressive Transformers for Indoor Scene Synthesis},
  booktitle    = NIPS,
  year         = {2021}
}

@inproceedings{DBLP:conf/3dim/WangYN21,
  author       = {Xinpeng Wang and
                  Chandan Yeshwanth and
                  Matthias Nie{\ss}ner},
  title        = {Scene{F}ormer: Indoor Scene Generation with Transformers},
  booktitle    = ThreeDV,
  year         = {2021}
}

@inproceedings{DBLP:conf/nips/FengZFJAHBWW23,
  author       = {Weixi Feng and
                  Wanrong Zhu and
                  Tsu{-}Jui Fu and
                  Varun Jampani and
                  Arjun R. Akula and
                  Xuehai He and
                  Sugato Basu and
                  Xin Eric Wang and
                  William Yang Wang},
  title        = {Layout{GPT}: Compositional Visual Planning and Generation with Large
                  Language Models},
  booktitle    = NIPS,
  year         = {2023}
}

@inproceedings{DBLP:conf/cvpr/YangJZH24,
  author       = {Yandan Yang and
                  Baoxiong Jia and
                  Peiyuan Zhi and
                  Siyuan Huang},
  title        = {Phy{S}cene: Physically Interactable 3{D} Scene Synthesis for Embodied
                  {AI}},
  booktitle    = CVPR,
  year         = {2024}
}

@inproceedings{DBLP:conf/eccv/OcalTKG24,
  author       = {Basak Melis {\"{O}}cal and
                  Maxim Tatarchenko and
                  Sezer Karaoglu and
                  Theo Gevers},
  title        = {Scene{T}eller: Language-to-3{D} Scene Generation},
  booktitle    = ECCV,
  year         = {2024}
}

@inproceedings{DBLP:conf/siggrapha/YanLWCSSLCDMLJ24,
  author       = {Han Yan and
                  Yang Li and
                  Zhennan Wu and
                  Shenzhou Chen and
                  Weixuan Sun and
                  Taizhang Shang and
                  Weizhe Liu and
                  Tian Chen and
                  Xiaqiang Dai and
                  Chao Ma and
                  Hongdong Li and
                  Pan Ji},
  title        = {Frankenstein: Generating Semantic-Compositional 3{D} Scenes in One Tri-Plane},
  booktitle    = SIGGRAPHA,
  year         = {2024}
}

@inproceedings{DBLP:conf/cvpr/TangNMDTN24,
  author       = {Jiapeng Tang and
                  Yinyu Nie and
                  Lev Markhasin and
                  Angela Dai and
                  Justus Thies and
                  Matthias Nie{\ss}ner},
  title        = {Diffu{S}cene: Denoising Diffusion Models for Generative Indoor Scene
                  Synthesis},
  booktitle    = CVPR,
  year         = {2024}
}

@inproceedings{DBLP:conf/3dim/ZhaoZLDG24,
  author       = {Yiqun Zhao and
                  Zibo Zhao and
                  Jing Li and
                  Sixun Dong and
                  Shenghua Gao},
  title        = {Room{D}esigner: Encoding Anchor-latents for Style-consistent and Shape-compatible Indoor Scene Generation},
  booktitle    = ThreeDV,
  year         = {2024}
}

@inproceedings{DBLP:conf/nips/MaillardSDO24,
  author       = {L{\'{e}}opold Maillard and
                  Nicolas Sereyjol{-}Garros and
                  Tom Durand and
                  Maks Ovsjanikov},
  title        = {{DeBaRA}: Denoising-Based 3{D} Room Arrangement Generation},
  booktitle    = NIPS,
  year         = {2024}
}

@inproceedings{DBLP:conf/mm/YeZLP24,
  author       = {Zhaoda Ye and
                  Xinhan Zheng and
                  Yang Liu and
                  Yuxin Peng},
  title        = {Rel{S}cene: {A} Benchmark and baseline for Spatial Relations in text-driven 3{D} Scene Generation},
  booktitle    = ACMMM,
  year         = {2024}
}

@inproceedings{DBLP:conf/cvpr/abs-2412-01801,
  author       = {Alexey Bokhovkin and
                  Quan Meng and
                  Shubham Tulsiani and
                  Angela Dai},
  title        = {Scene{F}actor: Factored Latent 3{D} Diffusion for Controllable 3{D} Scene Generation},
  booktitle    = CVPR,
  year         = {2025}
}

@article{DBLP:journals/corr/abs-2501-02519,
  author       = {Minglin Chen and
                  Longguang Wang and
                  Sheng Ao and
                  Ye Zhang and
                  Kai Xu and
                  Yulan Guo},
  title        = {Layout2{S}cene: 3{D} Semantic Layout Guided Scene Generation via Geometry
                  and Appearance Diffusion Priors},
  journal      = {arXiv},
  volume       = {2501.02519},
  year         = {2025}
}

@inproceedings{DBLP:conf/3dim/abs-2310-03602,
  author       = {Chuan Fang and
                  Xiaotao Hu and
                  Kunming Luo and
                  Ping Tan},
  title        = {{Ctrl-Room}: Controllable Text-to-3{D} Room Meshes Generation with Layout Constraints},
  booktitle    = ThreeDV,
  year         = {2025}
}

@inproceedings{DBLP:conf/3dim/PoW24,
  author       = {Ryan Po and
                  Gordon Wetzstein},
  title        = {Compositional 3{D} Scene Generation using Locally Conditioned Diffusion},
  booktitle    = ThreeDV,
  year         = {2024}
}

@inproceedings{DBLP:conf/iccvw/Cohen-BarRMGC23,
  author       = {Dana Cohen{-}Bar and
                  Elad Richardson and
                  Gal Metzer and
                  Raja Giryes and
                  Daniel Cohen{-}Or},
  title        = {Set-the-{S}cene: Global-Local Training for Generating Controllable NeRF
                  Scenes},
  booktitle    = ICCV,
  year         = {2023}
}

@inproceedings{DBLP:journals/corr/abs-2312-08885,
  author       = {Qihang Zhang and
                  Chaoyang Wang and
                  Aliaksandr Siarohin and
                  Peiye Zhuang and
                  Yinghao Xu and
                  Ceyuan Yang and
                  Dahua Lin and
                  Bolei Zhou and
                  Sergey Tulyakov and
                  Hsin{-}Ying Lee},
  title        = {Towards Text-guided 3D Scene Composition},
  booktitle    = CVPR,
  year         = {2024}
}

@inproceedings{DBLP:conf/eccv/LiSZWLWLZ24,
  author       = {Haoran Li and
                  Haolin Shi and
                  Wenli Zhang and
                  Wenjun Wu and
                  Yong Liao and
                  Lin Wang and
                  Lik{-}Hang Lee and
                  Peng Yuan Zhou},
  title        = {Dream{S}cene: 3{D} {G}aussian-Based Text-to-3{D} Scene Generation via Formation Pattern Sampling},
  booktitle    = ECCV,
  year         = {2024}
}

@article{DBLP:journals/corr/abs-2410-15391,
  author       = {Junwei Zhou and
                  Xueting Li and
                  Lu Qi and
                  Ming{-}Hsuan Yang},
  title        = {Layout-your-3{D}: Controllable and Precise 3{D} Generation with 2{D} Blueprint},
  journal      = {arXiv},
  volume       = {2410.15391},
  year         = {2024}
}

@inproceedings{DBLP:conf/icml/EpsteinPMEH24,
  author       = {Dave Epstein and
                  Ben Poole and
                  Ben Mildenhall and
                  Alexei A. Efros and
                  Aleksander Holynski},
  title        = {Disentangled 3{D} Scene Generation with Layout Learning},
  booktitle    = ICML,
  year         = {2024}
}

@inproceedings{DBLP:conf/icml/ZhouRXHLWS024,
  author       = {Xiaoyu Zhou and
                  Xingjian Ran and
                  Yajiao Xiong and
                  Jinlin He and
                  Zhiwei Lin and
                  Yongtao Wang and
                  Deqing Sun and
                  Ming{-}Hsuan Yang},
  title        = {{GALA3D:} Towards Text-to-3{D} Complex Scene Generation via Layout-guided Generative {G}aussian Splatting},
  booktitle    = ICML,
  year         = {2024}
}

@article{DBLP:journals/pami/ChenWL23,
  author       = {Zhaoxi Chen and
                  Guangcong Wang and
                  Ziwei Liu},
  title        = {Scene{D}reamer: Unbounded 3{D} Scene Generation From 2{D} Image Collections},
  journal      = PAMI,
  volume       = {45},
  number       = {12},
  pages        = {15562--15576},
  year         = {2023}
}

@inproceedings{DBLP:conf/iccv/LinLMCS0T23,
  author       = {Chieh Hubert Lin and
                  Hsin{-}Ying Lee and
                  Willi Menapace and
                  Menglei Chai and
                  Aliaksandr Siarohin and
                  Ming{-}Hsuan Yang and
                  Sergey Tulyakov},
  title        = {Infini{C}ity: Infinite-Scale City Synthesis},
  booktitle    = ICCV,
  year         = {2023}
}

@inproceedings{DBLP:conf/iccv/YangYGX0L23,
  author       = {Yuanbo Yang and
                  Yifei Yang and
                  Hanlei Guo and
                  Rong Xiong and
                  Yue Wang and
                  Yiyi Liao},
  title        = {{UrbanGIRAFFE}: Representing Urban Scenes as Compositional Generative
                  Neural Feature Fields},
  booktitle    = ICCV,
  year         = {2023}
}

@inproceedings{DBLP:conf/iccv/BahmaniPPYWGT23,
  author       = {Sherwin Bahmani and
                  Jeong Joon Park and
                  Despoina Paschalidou and
                  Xingguang Yan and
                  Gordon Wetzstein and
                  Leonidas J. Guibas and
                  Andrea Tagliasacchi},
  title        = {{CC3D:} Layout-Conditioned Generation of Compositional 3{D} Scenes},
  booktitle    = ICCV,
  year         = {2023}
}

@inproceedings{DBLP:conf/cvpr/ZhangXS0ZY24,
  author       = {Qihang Zhang and
                  Yinghao Xu and
                  Yujun Shen and
                  Bo Dai and
                  Bolei Zhou and
                  Ceyuan Yang},
  title        = {Berf{S}cene: Bev-conditioned Equivariant Radiance Fields for Infinite
                  3{D} Scene Generation},
  booktitle    = CVPR,
  year         = {2024}
}

@inproceedings{DBLP:journals/corr/abs-2406-06526,
  author       = {Haozhe Xie and
                  Zhaoxi Chen and
                  Fangzhou Hong and
                  Ziwei Liu},
  title        = {Generative {G}aussian Splatting for Unbounded 3{D} City Generation},
  booktitle    = CVPR,
  year         = {2025}
}

@article{DBLP:journals/corr/abs-2404-06780,
  author       = {Fan Lu and
                  Kwan{-}Yee Lin and
                  Yan Xu and
                  Hongsheng Li and
                  Guang Chen and
                  Changjun Jiang},
  title        = {Urban {A}rchitect: Steerable 3{D} Urban Scene Generation with Layout Prior},
  journal      = {arXiv},
  volume       = {2404.06780},
  year         = {2024}
}

@article{DBLP:journals/corr/abs-2501-08983,
  author       = {Haozhe Xie and
                  Zhaoxi Chen and
                  Fangzhou Hong and
                  Ziwei Liu},
  title        = {Compositional Generative Model of Unbounded 4{D} Cities},
  journal      = PAMI,
  volume       = {48},
  number       = {1},
  pages        = {312-328},
  year         = {2026}
}

@inproceedings{DBLP:conf/icml/KosiorekSZMSMR21,
  author       = {Adam R. Kosiorek and
                  Heiko Strathmann and
                  Daniel Zoran and
                  Pol Moreno and
                  Rosalia Schneider and
                  Sona Mokr{\'{a}} and
                  Danilo Jimenez Rezende},
  title        = {{NeRF-VAE}: {A} Geometry Aware 3{D} Scene Generative Model},
  booktitle    = ICML,
  year         = {2021}
}

@inproceedings{DBLP:conf/cvpr/Niemeyer021,
  author       = {Michael Niemeyer and
                  Andreas Geiger},
  title        = {{GIRAFFE:} Representing Scenes As Compositional Generative Neural
                  Feature Fields},
  booktitle    = CVPR,
  year         = {2021}
}

@inproceedings{DBLP:conf/iccv/DeVries0STS21,
  author       = {Terrance DeVries and
                  Miguel {\'{A}}ngel Bautista and
                  Nitish Srivastava and
                  Graham W. Taylor and
                  Joshua M. Susskind},
  title        = {Unconstrained Scene Generation with Locally Conditioned Radiance Fields},
  booktitle    = ICCV,
  year         = {2021}
}

@inproceedings{DBLP:conf/iccv/HaoMB021,
  author       = {Zekun Hao and
                  Arun Mallya and
                  Serge J. Belongie and
                  Ming{-}Yu Liu},
  title        = {{GANcraft}: Unsupervised 3{D} Neural Rendering of Minecraft Worlds},
  booktitle    = ICCV,
  year         = {2021}
}

@inproceedings{DBLP:conf/nips/0001GATTCDZGUDS22,
  author       = {Miguel {\'{A}}ngel Bautista and
                  Pengsheng Guo and
                  Samira Abnar and
                  Walter Talbott and
                  Alexander Toshev and
                  Zhuoyuan Chen and
                  Laurent Dinh and
                  Shuangfei Zhai and
                  Hanlin Goh and
                  Daniel Ulbricht and
                  Afshin Dehghan and
                  Joshua M. Susskind},
  title        = {{GAUDI:} {A} Neural Architect for Immersive 3{D} Scene Generation},
  booktitle    = NIPS,
  year         = {2022}
}

@inproceedings{DBLP:conf/nips/ShenMW22,
  author       = {Yuan Shen and
                  Wei{-}Chiu Ma and
                  Shenlong Wang},
  title        = {{SGAM:} Building a Virtual 3{D} World through Simultaneous Generation
                  and Mapping},
  booktitle    = NIPS,
  year         = {2022}
}

@inproceedings{DBLP:conf/cvpr/Chai0LIS23,
  author       = {Lucy Chai and
                  Richard Tucker and
                  Zhengqi Li and
                  Phillip Isola and
                  Noah Snavely},
  title        = {Persistent {N}ature: {A} Generative Model of Unbounded 3{D} Worlds},
  booktitle    = CVPR,
  year         = {2023}
}

@inproceedings{DBLP:conf/cvpr/0001BYKSLR0F23,
  author       = {Seung Wook Kim and
                  Bradley Brown and
                  Kangxue Yin and
                  Karsten Kreis and
                  Katja Schwarz and
                  Daiqing Li and
                  Robin Rombach and
                  Antonio Torralba and
                  Sanja Fidler},
  title        = {Neural{F}ield-{LDM}: Scene Generation with Hierarchical Latent Diffusion
                  Models},
  booktitle    = CVPR,
  year         = {2023}
}

@inproceedings{DBLP:conf/cvpr/JuHL00024,
  author       = {Xiaoliang Ju and
                  Zhaoyang Huang and
                  Yijiin Li and
                  Guofeng Zhang and
                  Yu Qiao and
                  Hongsheng Li},
  title        = {{DiffInDScene}: Diffusion-Based High-Quality 3{D} Indoor Scene Generation},
  booktitle    = CVPR,
  year         = {2024}
}

@article{DBLP:journals/tog/WuLYSSWCLSLJ24,
  author       = {Zhennan Wu and
                  Yang Li and
                  Han Yan and
                  Taizhang Shang and
                  Weixuan Sun and
                  Senbo Wang and
                  Ruikai Cui and
                  Weizhe Liu and
                  Hiroyuki Sato and
                  Hongdong Li and
                  Pan Ji},
  title        = {Block{F}usion: Expandable 3{D} Scene Generation using Latent Tri-plane
                  Extrapolation},
  journal      = TOG,
  volume       = {43},
  number       = {4},
  pages        = {43:1--43:17},
  year         = {2024}
}

@inproceedings{DBLP:conf/nips/LiLXQCZ0J24,
  author       = {Xinyang Li and
                  Zhangyu Lai and
                  Linning Xu and
                  Yansong Qu and
                  Liujuan Cao and
                  Shengchuan Zhang and
                  Bo Dai and
                  Rongrong Ji},
  title        = {Director3{D}: Real-world Camera Trajectory and 3{D} Scene Generation from Text},
  booktitle    = NIPS,
  year         = {2024}
}

@inproceedings{DBLP:conf/nips/WallingfordBKRD24,
  author       = {Matthew Wallingford and
                  Anand Bhattad and
                  Aditya Kusupati and
                  Vivek Ramanujan and
                  Matt Deitke and
                  Aniruddha Kembhavi and
                  Roozbeh Mottaghi and
                  Wei{-}Chiu Ma and
                  Ali Farhadi},
  title        = {{From an Image to a Scene}: Learning to Imagine the World from a Million
                  360{\textdegree} Videos},
  booktitle    = NIPS,
  year         = {2024}
}

@inproceedings{DBLP:journals/corr/abs-2412-21117,
  author       = {Yuanbo Yang and
                  Jiahao Shao and
                  Xinyang Li and
                  Yujun Shen and
                  Andreas Geiger and
                  Yiyi Liao},
  title        = {Prometheus: 3{D}-Aware Latent Diffusion Models for Feed-Forward Text-to-3{D} Scene Generation},
  booktitle    = CVPR,
  year         = {2025}
}

@inproceedings{DBLP:conf/nips/YangMCW24,
  author       = {Xiuyu Yang and
                  Yunze Man and
                  Junkun Chen and
                  Yu{-}Xiong Wang},
  title        = {Scene{C}raft: Layout-Guided 3{D} Scene Generation},
  booktitle    = NIPS,
  year         = {2024}
}

@article{DBLP:journals/corr/abs-2303-13843,
  author       = {Yiqi Lin and
                  Haotian Bai and
                  Sijia Li and
                  Haonan Lu and
                  Xiaodong Lin and
                  Hui Xiong and
                  Lin Wang},
  title        = {{CompoNeRF}: Text-guided Multi-object Compositional NeRF with Editable
                  3{D} Scene Layout},
  journal      = {arXiv},
  volume       = {2303.13843},
  year         = {2023}
}

@article{DBLP:journals/corr/abs-2309-17080,
  author       = {Anthony Hu and
                  Lloyd Russell and
                  Hudson Yeo and
                  Zak Murez and
                  George Fedoseev and
                  Alex Kendall and
                  Jamie Shotton and
                  Gianluca Corrado},
  title        = {{GAIA-1:} {A} Generative World Model for Autonomous Driving},
  journal      = {arXiv},
  volume       = {2309.17080},
  year         = {2023}
}

@inproceedings{DBLP:conf/naacl/DevlinCLT19,
  author       = {Jacob Devlin and
                  Ming{-}Wei Chang and
                  Kenton Lee and
                  Kristina Toutanova},
  title        = {{BERT:} Pre-training of Deep Bidirectional Transformers for Language
                  Understanding},
  booktitle    = {NAACL-HLT},
  year         = {2019}
}

@article{DBLP:journals/tog/MullerESK22,
  author       = {Thomas M{\"{u}}ller and
                  Alex Evans and
                  Christoph Schied and
                  Alexander Keller},
  title        = {Instant neural graphics primitives with a multiresolution hash encoding},
  journal      = TOG,
  volume       = {41},
  number       = {4},
  pages        = {102:1--102:15},
  year         = {2022}
}

@inproceedings{DBLP:conf/iccv/YuFDB23,
  author       = {Jason J. Yu and
                  Fereshteh Forghani and
                  Konstantinos G. Derpanis and
                  Marcus A. Brubaker},
  title        = {Long-Term Photometric Consistent Novel View Synthesis with Diffusion
                  Models},
  booktitle    = ICCV,
  year         = {2023}
}

@article{DBLP:journals/pami/WangWCWLL24,
  author       = {Guangcong Wang and
                  Peng Wang and
                  Zhaoxi Chen and
                  Wenping Wang and
                  Chen Change Loy and
                  Ziwei Liu},
  title        = {{PERF:} Panoramic Neural Radiance Field From a Single Panorama},
  journal      = PAMI,
  volume       = {46},
  number       = {10},
  pages        = {6905--6918},
  year         = {2024}
}

@inproceedings{DBLP:conf/iccv/KohLYBA21,
  author       = {Jing Yu Koh and
                  Honglak Lee and
                  Yinfei Yang and
                  Jason Baldridge and
                  Peter Anderson},
  title        = {Pathdreamer: {A} World Model for Indoor Navigation},
  booktitle    = ICCV,
  year         = {2021}
}

@inproceedings{DBLP:conf/nips/LeeKKS23,
  author       = {Yuseung Lee and
                  Kunho Kim and
                  Hyunjin Kim and
                  Minhyuk Sung},
  title        = {Sync{D}iffusion: Coherent Montage via Synchronized Joint Diffusions},
  booktitle    = NIPS,
  year         = {2023}
}

@inproceedings{DBLP:conf/cvpr/ZhangSHC023,
  author       = {Qinsheng Zhang and
                  Jiaming Song and
                  Xun Huang and
                  Yongxin Chen and
                  Ming{-}Yu Liu},
  title        = {Diff{C}ollage: Parallel Generation of Large Content with Diffusion Models},
  booktitle    = CVPR,
  year         = {2023}
}

@inproceedings{DBLP:conf/cvpr/abs-2503-07152,
  author       = {Yuheng Liu and Xinke Li and Yuning Zhang and Lu Qi and Xin Li and Wenping Wang and Chongshou Li and Xueting Li and Ming-Hsuan Yang},
  title        = {Controllable 3{D} Outdoor Scene Generation via Scene Graphs},
  booktitle    = CVPR,
  year         = {2025}
}

@inproceedings{DBLP:conf/iclr/LinLCT022,
  author       = {Chieh Hubert Lin and
                  Hsin{-}Ying Lee and
                  Yen{-}Chi Cheng and
                  Sergey Tulyakov and
                  Ming{-}Hsuan Yang},
  title        = {Infinity{GAN}: Towards Infinite-Pixel Image Synthesis},
  booktitle    = ICLR,
  year         = {2022}
}

@inproceedings{DBLP:conf/iclr/PooleJBM23,
  author       = {Ben Poole and
                  Ajay Jain and
                  Jonathan T. Barron and
                  Ben Mildenhall},
  title        = {Dream{F}usion: Text-to-3{D} using 2{D} Diffusion},
  booktitle    = ICLR,
  year         = {2023}
}

@inproceedings{DBLP:conf/cvpr/KimP0F21,
  author       = {Seung Wook Kim and
                  Jonah Philion and
                  Antonio Torralba and
                  Sanja Fidler},
  title        = {Drive{GAN}: Towards a Controllable High-Quality Neural Simulation},
  booktitle    = CVPR,
  year         = {2021}
}

@inproceedings{DBLP:conf/cvpr/GaoLC0S24,
  author       = {Gege Gao and
                  Weiyang Liu and
                  Anpei Chen and
                  Andreas Geiger and
                  Bernhard Sch{\"{o}}lkopf},
  title        = {Graph{D}reamer: Compositional 3{D} Scene Synthesis from Scene Graphs},
  booktitle    = CVPR,
  year         = {2024}
}

@inproceedings{DBLP:conf/siggrapha/LiL0M024,
  author       = {Xiao{-}Lei Li and
                  Haodong Li and
                  Hao{-}Xiang Chen and
                  Tai{-}Jiang Mu and
                  Shi{-}Min Hu},
  title        = {{DIS}cene: Object Decoupling and Interaction Modeling for Complex Scene
                  Generation},
  booktitle    = SIGGRAPHA,
  year         = {2024}
}

@inproceedings{DBLP:conf/cvpr/NieD0N23,
  author       = {Yinyu Nie and
                  Angela Dai and
                  Xiaoguang Han and
                  Matthias Nie{\ss}ner},
  title        = {Learning 3{D} Scene Priors with 2{D} Supervision},
  booktitle    = CVPR,
  year         = {2023}
}

@inproceedings{DBLP:conf/cvpr/abs-2412-03558,
  author       = {Zehuan Huang and
                  Yuan{-}Chen Guo and
                  Xingqiao An and
                  Yunhan Yang and
                  Yangguang Li and
                  Zi{-}Xin Zou and
                  Ding Liang and
                  Xihui Liu and
                  Yan{-}Pei Cao and
                  Lu Sheng},
  title        = {{MIDI:} Multi-Instance Diffusion for Single Image to 3{D} Scene Generation},
  booktitle    = CVPR,
  year         = {2024}
}

@article{DBLP:journals/corr/abs-2409-01595,
  author       = {Junpeng Jiang and
                  Gangyi Hong and
                  Lijun Zhou and
                  Enhui Ma and
                  Hengtong Hu and
                  Xia Zhou and
                  Jie Xiang and
                  Fan Liu and
                  Kaicheng Yu and
                  Haiyang Sun and
                  Kun Zhan and
                  Peng Jia and
                  Miao Zhang},
  title        = {{DiVE}: DiT-based Video Generation with Enhanced Control},
  journal      = {arXiv},
  volume       = {2409.01595},
  year         = {2024}
}

@inproceedings{DBLP:conf/aaai/ZhaoWZCHB025,
  author       = {Guosheng Zhao and
                  Xiaofeng Wang and
                  Zheng Zhu and
                  Xinze Chen and
                  Guan Huang and
                  Xiaoyi Bao and
                  Xingang Wang},
  title        = {Drive{D}reamer-2: {LLM}-Enhanced World Models for Diverse Driving Video Generation},
  booktitle    = AAAI,
  year         = {2025},
}

@article{DBLP:journals/corr/abs-2412-19505,
  author       = {Xiaotao Hu and
                  Wei Yin and
                  Mingkai Jia and
                  Junyuan Deng and
                  Xiaoyang Guo and
                  Qian Zhang and
                  Xiaoxiao Long and
                  Ping Tan},
  title        = {Driving{W}orld: Constructing World Model for Autonomous Driving via
                  Video {GPT}},
  journal      = {arXiv},
  volume       = {2412.19505},
  year         = {2024}
}

@article{DBLP:journals/corr/abs-2412-01407,
  author       = {Zehuan Wu and
                  Jingcheng Ni and
                  Xiaodong Wang and
                  Yuxin Guo and
                  Rui Chen and
                  Lewei Lu and
                  Jifeng Dai and
                  Yuwen Xiong},
  title        = {Holo{D}rive: Holistic {2D-3D} Multi-Modal Street Scene Generation for
                  Autonomous Driving},
  journal      = {arXiv},
  volume       = {2412.01407},
  year         = {2024}
}

@article{DBLP:journals/corr/abs-2407-05679,
  author       = {Yumeng Zhang and
                  Shi Gong and
                  Kaixin Xiong and
                  Xiaoqing Ye and
                  Xiao Tan and
                  Fan Wang and
                  Jizhou Huang and
                  Hua Wu and
                  Haifeng Wang},
  title        = {{BEVW}orld: {A} Multimodal World Model for Autonomous Driving via Unified
                  {BEV} Latent Space},
  journal      = {arXiv},
  volume       = {2407.05679},
  year         = {2024}
}

@article{DBLP:journals/corr/abs-2406-01349,
  author       = {Enhui Ma and
                  Lijun Zhou and
                  Tao Tang and
                  Zhan Zhang and
                  Dong Han and
                  Junpeng Jiang and
                  Kun Zhan and
                  Peng Jia and
                  Xianpeng Lang and
                  Haiyang Sun and
                  Di Lin and
                  Kaicheng Yu},
  title        = {Unleashing Generalization of End-to-End Autonomous Driving with Controllable
                  Long Video Generation},
  journal      = {arXiv},
  volume       = {2406.01349},
  year         = {2024}
}

@inproceedings{DBLP:conf/cvpr/NiGLCLW25,
  author       = {Jingcheng Ni and 
                  Yuxin Guo and 
                  Yichen Liu and 
                  Rui Chen and 
                  Lewei Lu and 
                  Zehuan Wu},
  title        = {{MaskGWM}: A Generalizable Driving World Model with Video Mask Reconstruction},
  booktitle    = CVPR,
  year         = {2025}
}

@inproceedings{DBLP:conf/cvpr/abs-2503-13265,
  author       = {Yanhao Wu and Haoyang Zhang and Tianwei Lin and Lichao Huang and Shujie Luo and Rui Wu and Congpei Qiu and Wei Ke and Tong Zhang},
  title        = {Generating Multimodal Driving Scenes via Next-Scene Prediction},
  booktitle    = CVPR,
  year         = {2025}
}

@inproceedings{DBLP:conf/cvpr/abs-2409-08215,
  author       = {Quan Meng and
                  Lei Li and
                  Matthias Nie{\ss}ner and
                  Angela Dai},
  title        = {{LT3SD:} Latent Trees for 3{D} Scene Diffusion},
  booktitle    = CVPR,
  year         = {2025}
}

@article{DBLP:journals/corr/abs-2503-14501,
  author       = {Qiaowei Miao and
                  Kehan Li and
                  Jinsheng Quan and
                  Zhiyuan Min and
                  Shaojie Ma and
                  Yichao Xu and
                  Yi Yang and
                  Yawei Luo},
  title        = {Advances in 4{D} Generation: A Survey},
  journal      = {arXiv},
  volume       = {2503.14501},
  year         = {2025}
}

@inproceedings{DBLP:conf/nips/Ouyang0JAWMZASR22,
  author       = {Long Ouyang and
                  Jeffrey Wu and
                  Xu Jiang and
                  Diogo Almeida and
                  Carroll L. Wainwright and
                  Pamela Mishkin and
                  Chong Zhang and
                  Sandhini Agarwal and
                  Katarina Slama and
                  Alex Ray and
                  John Schulman and
                  Jacob Hilton and
                  Fraser Kelton and
                  Luke Miller and
                  Maddie Simens and
                  Amanda Askell and
                  Peter Welinder and
                  Paul F. Christiano and
                  Jan Leike and
                  Ryan Lowe},
  title        = {Training language models to follow instructions with human feedback},
  booktitle    = NIPS,
  year         = {2022}
}

@article{DBLP:journals/corr/abs-2303-08774,
  author       = {OpenAI},
  title        = {{GPT-4} Technical Report},
  journal      = {arXiv},
  volume       = {2303.08774},
  year         = {2023}
}

@inproceedings{DBLP:conf/cvpr/XuCSPSSYSLZT23,
  author       = {Yinghao Xu and
                  Menglei Chai and
                  Zifan Shi and
                  Sida Peng and
                  Ivan Skorokhodov and
                  Aliaksandr Siarohin and
                  Ceyuan Yang and
                  Yujun Shen and
                  Hsin{-}Ying Lee and
                  Bolei Zhou and
                  Sergey Tulyakov},
  title        = {{DisCoScene}: Spatially Disentangled Generative Radiance Fields for
                  Controllable 3{D}-aware Scene Synthesis},
  booktitle    = CVPR,
  year         = {2023}
}

@inproceedings{DBLP:conf/cvpr/LeeLJISY24,
  author       = {Jumin Lee and
                  Sebin Lee and
                  Changho Jo and
                  Woobin Im and
                  Juhyeong Seon and
                  Sung{-}Eui Yoon},
  title        = {Sem{C}ity: Semantic Scene Generation with Triplane Diffusion},
  booktitle    = CVPR,
  year         = {2024}
}

@inproceedings{DBLP:conf/cvpr/RenHZMFW24,
  author       = {Xuanchi Ren and
                  Jiahui Huang and
                  Xiaohui Zeng and
                  Ken Museth and
                  Sanja Fidler and
                  Francis Williams},
  title        = {X{C}ube: Large-Scale 3{D} Generative Modeling using Sparse Voxel Hierarchies},
  booktitle    = CVPR,
  year         = {2024}
}

@inproceedings{DBLP:conf/eccv/LiuLLQLY24,
  author       = {Yuheng Liu and
                  Xinke Li and
                  Xueting Li and
                  Lu Qi and
                  Chongshou Li and
                  Ming{-}Hsuan Yang},
  title        = {Pyramid Diffusion for Fine 3{D} Large Scene Generation},
  booktitle    = ECCV,
  year         = {2024}
}

@article{DBLP:journals/corr/abs-2301-00527,
  author       = {Jumin Lee and
                  Woobin Im and
                  Sebin Lee and
                  Sung{-}Eui Yoon},
  title        = {Diffusion Probabilistic Models for Scene-Scale 3{D} Categorical Data},
  journal      = {arXiv},
  volume       = {2301.00527},
  year         = {2023}
}

@inproceedings{DBLP:conf/iclr/abs-2410-18084,
  author       = {Hengwei Bian and Lingdong Kong and Haozhe Xie and Liang Pan and Yu Qiao and Ziwei Liu},
  title        = {{DynamicCity}: Large-Scale 4{D} Occupancy Generation from Dynamic Scenes},
  booktitle    = ICLR,
  year         = {2025}
}

@inproceedings{DBLP:conf/graphicsinterface/XuSF02,
  author       = {Ken Xu and
                  James Stewart and
                  Eugene Fiume},
  title        = {Constraint-Based Automatic Placement for Scene Composition},
  booktitle    = {Graphics Interface},
  year         = {2002}
}

@article{DBLP:journals/tog/MerrellSLAK11,
  author       = {Paul Merrell and
                  Eric Schkufza and
                  Zeyang Li and
                  Maneesh Agrawala and
                  Vladlen Koltun},
  title        = {Interactive furniture layout using interior design guidelines},
  journal      = TOG,
  volume       = {30},
  number       = {4},
  pages        = {87},
  year         = {2011}
}

@article{DBLP:journals/tvcg/YuYT16,
  author       = {Lap{-}Fai Yu and
                  Sai Kit Yeung and
                  Demetri Terzopoulos},
  title        = {The {C}lutterpalette: An Interactive Tool for Detailing Indoor Scenes},
  journal      = TVCG,
  volume       = {22},
  number       = {2},
  pages        = {1138--1148},
  year         = {2016}
}

@article{DBLP:journals/corr/abs-1712-05474,
  author       = {Eric Kolve and
                  Roozbeh Mottaghi and
                  Daniel Gordon and
                  Yuke Zhu and
                  Abhinav Gupta and
                  Ali Farhadi},
  title        = {{AI2-THOR:} An Interactive 3{D} Environment for Visual {AI}},
  journal      = {arXiv},
  volume       = {1712.05474},
  year         = {2017}
}

@inproceedings{DBLP:conf/cvpr/QiZHJZ18,
  author       = {Siyuan Qi and
                  Yixin Zhu and
                  Siyuan Huang and
                  Chenfanfu Jiang and
                  Song{-}Chun Zhu},
  title        = {Human-{C}entric Indoor Scene Synthesis Using Stochastic Grammar},
  booktitle    = CVPR,
  year         = {2018}
}

@inproceedings{DBLP:conf/iccv/FuC0ZWLZSJZ021,
  author       = {Huan Fu and
                  Bowen Cai and
                  Lin Gao and
                  Lingxiao Zhang and
                  Jiaming Wang and
                  Cao Li and
                  Qixun Zeng and
                  Chengyue Sun and
                  Rongfei Jia and
                  Binqiang Zhao and
                  Hao Zhang},
  title        = {{3D-FRONT}: 3{D} Furnished Rooms with layOuts and semaNTics},
  booktitle    = ICCV,
  year         = {2021}
}

@article{DBLP:journals/corr/abs-2111-05527,
  author       = {Yizhou Zhao and
                  Kaixiang Lin and
                  Zhiwei Jia and
                  Qiaozi Gao and
                  Govind Thattai and
                  Jesse Thomason and
                  Gaurav S. Sukhatme},
  title        = {{LUMINOUS:} Indoor Scene Generation for Embodied {AI} Challenges},
  journal      = {arXiv},
  volume       = {2111.05527},
  year         = {2021}
}

@inproceedings{DBLP:conf/mm/ZhangLHYZ21,
  author       = {Shao{-}Kui Zhang and
                  Yi{-}Xiao Li and
                  Yu He and
                  Yong{-}Liang Yang and
                  Song{-}Hai Zhang},
  title        = {Mage{A}dd: Real-Time Interaction Simulation for Scene Synthesis},
  booktitle    = ACMMM,
  year         = {2021}
}

@inproceedings{DBLP:conf/siggrapha/YeWLPLX022,
  author       = {Sifan Ye and
                  Yixing Wang and
                  Jiaman Li and
                  Dennis Park and
                  C. Karen Liu and
                  Huazhe Xu and
                  Jiajun Wu},
  title        = {Scene Synthesis from Human Motion},
  booktitle    = SIGGRAPHA,
  year         = {2022}
}

@inproceedings{DBLP:conf/siggrapha/LeimerG0M22,
  author       = {Kurt Leimer and
                  Paul Guerrero and
                  Tomer Weiss and
                  Przemyslaw Musialski},
  title        = {Layout{E}nhancer: Generating Good Indoor Layouts from Imperfect Data},
  booktitle    = SIGGRAPHA,
  year         = {2022}
}

@inproceedings{DBLP:conf/cvpr/YiHTHTB23,
  author       = {Hongwei Yi and
                  Chun{-}Hao P. Huang and
                  Shashank Tripathi and
                  Lea Hering and
                  Justus Thies and
                  Michael J. Black},
  title        = {{MIME:} Human-Aware 3{D} Scene Generation},
  booktitle    = CVPR,
  year         = {2023}
}

@inproceedings{DBLP:conf/nips/AnVNHNVN23,
  author       = {Vuong Dinh An and
                  Minh Nhat Vu and
                  Toan Nguyen and
                  Baoru Huang and
                  Dzung Nguyen and
                  Thieu Vo and
                  Anh Nguyen},
  title        = {Language-driven Scene Synthesis using Multi-conditional Diffusion
                  Model},
  booktitle    = NIPS,
  year         = {2023}
}

@inproceedings{DBLP:conf/cvpr/YangSWVHH0HKLCY24,
  author       = {Yue Yang and
                  Fan{-}Yun Sun and
                  Luca Weihs and
                  Eli VanderBilt and
                  Alvaro Herrasti and
                  Winson Han and
                  Jiajun Wu and
                  Nick Haber and
                  Ranjay Krishna and
                  Lingjie Liu and
                  Chris Callison{-}Burch and
                  Mark Yatskar and
                  Aniruddha Kembhavi and
                  Christopher Clark},
  title        = {Holodeck: Language Guided Generation of 3{D} Embodied {AI} Environments},
  booktitle    = CVPR,
  year         = {2024}
}

@inproceedings{DBLP:conf/nips/WangQLCCWWXG24,
  author       = {Yian Wang and
                  Xiaowen Qiu and
                  Jiageng Liu and
                  Zhehuan Chen and
                  Jiting Cai and
                  Yufei Wang and
                  Tsun{-}Hsuan Johnson Wang and
                  Zhou Xian and
                  Chuang Gan},
  title        = {Architect: Generating Vivid and Interactive 3{D} Scenes with Hierarchical
                  2{D} Inpainting},
  booktitle    = NIPS,
  year         = {2024}
}

@inproceedings{DBLP:conf/mm/ZhangHYZLLZ24,
  author       = {Shao{-}Kui Zhang and
                  Junkai Huang and
                  Liang Yue and
                  Jia{-}Tong Zhang and
                  Jia{-}Hong Liu and
                  Yu{-}Kun Lai and
                  Song{-}Hai Zhang},
  title        = {Scene{E}xpander: Real-Time Scene Synthesis for Interactive Floor Plan
                  Editing},
  booktitle    = ACMMM,
  year         = {2024}
}

@inproceedings{DBLP:conf/siggraph/LiHZW24,
  author       = {Jianan Li and
                  Tao Huang and
                  Qingxu Zhu and
                  Tien{-}Tsin Wong},
  title        = {Physics-based Scene Layout Generation from Human Motion},
  booktitle    = SIGGRAPH,
  year         = {2024}
}

@inproceedings{DBLP:conf/3dim/abs-2310-12945,
  author       = {Chunyi Sun and
                  Junlin Han and
                  Weijian Deng and
                  Xinlong Wang and
                  Zishan Qin and
                  Stephen Gould},
  title        = {{3D-GPT}: Procedural 3{D} Modeling with Large Language Models},
  booktitle    = ThreeDV,
  year         = {2025}
}

@inproceedings{DBLP:conf/mm/LiuZZZ24,
  author       = {Jia{-}Hong Liu and
                  Shao{-}Kui Zhang and
                  Chuyue Zhang and
                  Song{-}Hai Zhang},
  title        = {Controllable Procedural Generation of Landscapes},
  booktitle    = ACMMM,
  year         = {2024}
}

@article{DBLP:journals/corr/abs-2502-15601,
  author       = {Xinhang Liu and
                  Chi{-}Keung Tang and
                  Yu{-}Wing Tai},
  title        = {World{C}raft: Photo-Realistic 3{D} World Creation and Customization via
                  {LLM} Agents},
  journal      = {arXiv},
  volume       = {2502.15601},
  year         = {2025}
}

@inproceedings{DBLP:conf/siggraph/ParishM01,
  author       = {Yoav I. H. Parish and
                  Pascal M{\"{u}}ller},
  title        = {Procedural modeling of cities},
  booktitle    = SIGGRAPH,
  year         = {2001}
}

@article{DBLP:journals/cgf/NishidaGA16,
  author       = {Gen Nishida and
                  Ignacio Garcia{-}Dorado and
                  Daniel G. Aliaga},
  title        = {Example-Driven Procedural Urban Roads},
  journal      = {Comput. Graph. Forum},
  volume       = {35},
  number       = {6},
  pages        = {5--17},
  year         = {2016}
}

@article{DBLP:journals/corr/abs-2406-04983,
  author       = {Jie Deng and
                  Wenhao Chai and
                  Junsheng Huang and
                  Zhonghan Zhao and
                  Qixuan Huang and
                  Mingyan Gao and
                  Jianshu Guo and
                  Shengyu Hao and
                  Wenhao Hu and
                  Jenq{-}Neng Hwang and
                  Xi Li and
                  Gaoang Wang},
  title        = {City{C}raft: {A} Real Crafter for 3{D} City Generation},
  journal      = {arXiv},
  volume       = {2406.04983},
  year         = {2024}
}

@inproceedings{DBLP:conf/cvpr/ChanLCNPMGGTKKW22,
  author       = {Eric R. Chan and
                  Connor Z. Lin and
                  Matthew A. Chan and
                  Koki Nagano and
                  Boxiao Pan and
                  Shalini De Mello and
                  Orazio Gallo and
                  Leonidas J. Guibas and
                  Jonathan Tremblay and
                  Sameh Khamis and
                  Tero Karras and
                  Gordon Wetzstein},
  title        = {Efficient Geometry-aware 3{D} Generative Adversarial Networks},
  booktitle    = CVPR,
  year         = {2022}
}

@article{doi:10.1126/science.156.3775.636,
  author = {Benoit Mandelbrot},
  title = {How Long Is the Coast of Britain? Statistical Self-Similarity and Fractional Dimension},
  journal = {Science},
  volume = {156},
  number = {3775},
  pages = {636-638},
  year = {1967}
}

@article{mandelbrot1983fractal,
  title={The fractal geometry of nature},
  author={Benoit Mandelbrot},
  journal={New York},
  year={1983}
}

@article{lindenmayer1968mathematical,
  title={Mathematical models for cellular interactions in development I. Filaments with one-sided inputs},
  author={Aristid Lindenmayer},
  journal={Journal of theoretical biology},
  volume={18},
  number={3},
  pages={280--299},
  year={1968}
}

@article{mandelbrot1968fractional,
  title={Fractional Brownian motions, fractional noises and applications},
  author={Mandelbrot Benoit B and Van Ness John W},
  journal={SIAM review},
  volume={10},
  number={4},
  pages={422--437},
  year={1968}
}

@inproceedings{DBLP:conf/siggraph/KelleyMN88,
  author       = {Alex D. Kelley and
                  Michael C. Malin and
                  Gregory M. Nielson},
  title        = {Terrain simulation using a model of stream erosion},
  booktitle    = SIGGRAPH,
  year         = {1988}
}

@inproceedings{DBLP:conf/siggraph/MusgraveKM89,
  author       = {F. Kenton Musgrave and
                  Craig E. Kolb and
                  Robert S. Mace},
  title        = {The synthesis and rendering of eroded fractal terrains},
  booktitle    = SIGGRAPH,
  year         = {1989}
}

@inproceedings{DBLP:conf/siggraph/DeussenHLMPP98,
  author       = {Oliver Deussen and
                  Pat Hanrahan and
                  Bernd Lintermann and
                  Radom{\'{\i}}r Mech and
                  Matt Pharr and
                  Przemyslaw Prusinkiewicz},
  title        = {Realistic Modeling and Rendering of Plant Ecosystems},
  booktitle    = SIGGRAPH,
  year         = {1998}
}

@inproceedings{DBLP:conf/graphite/BelhadjA05,
  author       = {Far{\`{e}}s Belhadj and
                  Pierre Audibert},
  title        = {Modeling landscapes with ridges and rivers: bottom up approach},
  booktitle    = {GRAPHITE},
  year         = {2005}
}
